\documentclass[journal]{IEEEtran}
\usepackage{amsmath,amsfonts}
\usepackage{algorithmic}
\usepackage{algorithm}
\usepackage{array}
\usepackage[caption=false,font=normalsize,labelfont=sf,textfont=sf]{subfig}
\usepackage{textcomp}
\usepackage{stfloats}
\usepackage{url}
\usepackage{verbatim}
\usepackage{graphicx}
\usepackage{cite}
\usepackage{amsthm}  
\usepackage{amssymb}
\usepackage{color} 

\newtheorem{theorem}{\textbf{Theorem}}[section]
 
\newtheorem{problem}[theorem]{\textbf{Problem}}
\newtheorem{definition}[theorem]{\textbf{Definition}}
\newtheorem{proposition}[theorem]{\textbf{Proposition}}
\newtheorem{corollary}[theorem]{\textbf{Corollary}}

\newcommand{\norm}[1]{|\!|#1|\!|}

\newcommand{\NN}{{\mathcal{N}\negthickspace\mathcal{N}}\negthinspace}
\newcommand{\pr}{\mathrm{Pr}}
\newcommand{\Int}{\mathrm{Int}}
\newcommand{\abs}{\mathrm{abs}}
\newcommand{\ct}{\mathrm{ct}}

\begin{document}

\title{
Neurosymbolic Motion and Task Planning for Linear Temporal Logic Tasks
}

\author{Xiaowu Sun,~\IEEEmembership{Graduate Student Member,~IEEE} and Yasser Shoukry,~\IEEEmembership{Senior Member,~IEEE}
\thanks{The authors are with the Department of Electrical Engineering and Computer Science, University of California, Irvine, CA 92697, USA (e-mail: \{xiaowus,yshoukry\}@uci.edu).}}




\maketitle

\begin{abstract}
This paper presents a neurosymbolic framework to solve motion planning problems for mobile robots involving temporal goals. The temporal goals are described using temporal logic formulas such as Linear Temporal Logic (LTL) to capture complex tasks. The proposed framework trains Neural Network (NN)-based planners that enjoy strong correctness guarantees when applying to unseen tasks, i.e., the exact task (including workspace, LTL formula, and dynamic constraints of a robot) is unknown during the training of NNs. Our approach to achieving theoretical guarantees and computational efficiency is based on two insights. First, we incorporate a symbolic model into the training of NNs such that the resulting NN-based planner inherits the interpretability and correctness guarantees of the symbolic model. Moreover, the symbolic model serves as a discrete ``memory'', which is necessary for satisfying temporal logic formulas. Second, we train a library of neural networks offline and combine a subset of the trained NNs into a single NN-based planner at runtime when a task is revealed. In particular, we develop a novel constrained NN training procedure, named formal NN training, to enforce that each neural network in the library represents a ``symbol'' in the symbolic model. As a result, our neurosymbolic framework enjoys the scalability and flexibility benefits of machine learning and inherits the provable guarantees from control-theoretic and formal-methods techniques. We demonstrate the effectiveness of our framework in both simulations and on an actual robotic vehicle, and show that our framework can generalize to unknown tasks where state-of-the-art meta-reinforcement learning techniques fail. 
\end{abstract}


\section{Introduction}
\label{sec:introduction}


\IEEEPARstart{D}{eveloping} 
intelligent machines with a considerable level of cognition dates to the early 1950s. With the current rise of machine learning (ML) techniques, robotic platforms are witnessing a breakthrough in their cognition. Nevertheless, regardless of how many environments they were trained (or programmed) to consider, such intelligent machines will always face new environments which the human designer failed to examine during the training phase. To circumvent the lack of autonomous systems to adapt to new environments, several researchers asked whether we could build autonomous agents that can learn how to learn. In other words, while conventional machine learning focuses on designing agents that can perform one task, the so-called meta-learning aims instead to solve the problem of designing agents that can generalize to different tasks that were not considered during the design or the training of these agents. For example, in the context of meta-Reinforcement Learning (meta-RL), given data collected from a multitude of tasks (e.g., changes in the environments, goals, and robot dynamics), meta-RL aims to combine all such experiences and use them to design agents that can quickly adapt to unseen tasks. While the current successes of meta-RL are undeniable, significant drawbacks of meta-RL in its current form are (i) \emph{the lack of formal guarantees on its ability to generalize to unseen tasks}, (ii) \emph{the lack of formal guarantees with regards to its safety} and (iii) \emph{the lack of interpretability due to the use of black-box deep learning techniques}.

In this paper, we focus on the problem of designing Neural Network (NN)-based task and motion planners that are guaranteed to generalize to unseen tasks, enjoy strong safety guarantees, and are interpretable. We consider agents who need to accomplish temporal goals captured by temporal logic formulas such as Linear Temporal Logic (LTL)~\cite{bltl,scltl}. The use of LTL in task and motion planning has been widely studied (e.g., ~\cite{kress2011correct,kress2018synthesis,bhatia2011motion,garrett2018ffrob,guo2013motion,bhatia2010sampling,fainekos2005hybrid,fainekos2006translating,kress2007s,shoukry2017linear,tabuada2009verification,belta2017formal}) due to the ability of LTL formulas to capture complex goals such as ``eventually visit region A followed by a visit to region B or region C while always avoiding hitting obstacle D.'' On the one hand,  motion and task planning using symbolic techniques enjoy the guarantees of satisfying task specifications in temporal logic. Nevertheless, these algorithms need an explicit model of the dynamic constraints of the robot and suffer from computational complexity whenever such dynamic constraints are highly nonlinear and complex. On the other hand, machine learning approaches are capable of training NN planners without the explicit knowledge of the dynamic constraints and scale favorably to highly nonlinear and complex dynamics. Nevertheless, these data-driven approaches suffer from the lack of safety and generalization guarantees. Therefore, in this work, we aim to design a novel \emph{neurosymbolic} framework for motion and task planning by combining the benefits of symbolic control and machine learning techniques.

At the heart of the proposed framework is using a symbolic model to guide the training of NNs and restricting the behavior of NNs to ``symbols'' in the symbolic model. Specifically, our framework consists of offline (or training) and online (or runtime) phases. During the offline phase, we assume access to a ``nominal'' simulator that approximates the dynamic constraints of a robot. We assume no knowledge of the exact task (e.g., workspace, LTL formula, and exact dynamic constraints of a robot). We use this information to train a ``library'' of NNs through a novel NN training procedure, named formal NN training, which enforces each trained NN to represent a continuous piece-wise affine (CPWA) function from a chosen family of CPWA functions. 
The exact task becomes available only during the online (or runtime) phase. Given the dynamic constraints of a robot, we compute a finite-state Markov decision process (MDP) as our symbolic model. Thanks to the formal NN training procedure, the symbolic model can be constructed so that each of the trained NNs in the library represents a transition in the MDP (and hence a symbol in this MDP). By analyzing this symbolic model, our framework selects NNs from the library and combines them into a single NN-based planner to perform the task and motion planning. 


In summary, the main contributions of this article are:
\begin{itemize}
    \item[1)] We propose a \emph{neurosymbolic} framework that integrates machine learning and symbolic techniques in training NN-based planners for an agent to accomplish \emph{unseen} tasks. Thanks to the use of a symbolic model, the resulting NN-based planners are guaranteed to satisfy the temporal goals described in linear temporal logic formulas, which cannot be satisfied by existing NN training algorithms.
    
    \item[2)] We develop a formal training algorithm that restricts the trained NNs to specific local behavior. The training procedure combines classical gradient descent training of NNs with a novel \emph{NN weight projection operator} that modifies the NN weights as little as possible to ensure the trained NN belongs to a chosen family of CPWA functions. We provide theoretical guarantees on the proposed \emph{NN weight projection operator} in terms of correctness and upper bounds on the error between the NN before and after the projection.
    
    \item[3)] We provide a theoretical analysis of the overall \emph{neurosymbolic} framework. We show theoretical guarantees that govern the correctness of the resulting NN-based planners when generalizing to \emph{unseen} tasks, including unknown workspaces, unknown temporal logic formulas, and uncertain dynamic constraints.
    
    \item[4)] We pursue the high performance of the proposed framework in fast adaptation to unseen tasks with efficient training. For example, we accelerate the training of NNs by employing ideas from transfer learning and constructing the symbolic model using a data-driven approach. We validate the effectiveness of the proposed framework on an actual robotic vehicle and demonstrate that our framework can generalize to unknown tasks where state-of-the-art meta-RL techniques are known to fail (e.g., when the tasks are chosen from across homotopy classes~\cite{cao2021ral}).
\end{itemize}

The remainder of the paper is organized as follows. After the problem formulation in Section~\ref{sec:formulation}, we present the formal NN training algorithm in Section~\ref{sec:formal_training}. In Section~\ref{sec:algorithm}, we introduce the neurosymbolic framework that uses the formal NN training algorithm to obtain a library of NNs and combines them into a single NN-based planner at runtime. In Section~\ref{sec:guarantees}, we provide theoretical guarantees of the proposed framework. In Section~\ref{sec:speedups}, we present some key elements for performance improvement while maintaining the same theoretical guarantees. Experimental results are given in Section~\ref{sec:results}, and all proofs can be found in the appendix.

\noindent \textbf{Comparison with the preliminary results:} A preliminary version of this article was presented in~\cite{sun2022cdc}. In~\cite{sun2022cdc}, we confined our goal to generating collision-free trajectories, whereas in this work we consider agents that need to satisfy general temporal logic formulas such as LTL. Also, we allow the temporal logic formulas and the exact robot dynamics to be unknown during the training of NNs. In this article, we present for the first time the formal NN training algorithm (see Section~\ref{sec:formal_training}). Moreover, we present a theoretical analysis of the proposed framework (see Section~\ref{sec:guarantees}). All the speedup techniques in Section~\ref{sec:speedups}, the implementation of our framework on an actual robotic vehicle, and the performance comparison with meta-RL algorithms are also new in this article. 

\noindent \textbf{Related work:} The literature on the safe design of ML-based motion and task planners can be classified according to three broad approaches, namely (i) incorporating safety in the training of ML-based planners, (ii) post-training verification of ML models, and (iii) online validation of safety and control intervention. Representative examples of the first approach include reward-shaping~\cite{stone2020reward,saunders2018trial}, Bayesian and robust regression~\cite{berkenkamp2016bayesian,liu2019robust,pauli2020training}, and policy optimization with constraints~\cite{abbeel2017icml,turchetta2016safe,wen2020safe}. Unfortunately, these approaches do not provide provable guarantees about the safety of the trained ML-based planners. 

To provide strong safety and reliability guarantees, several works in the literature focus on applying formal verification techniques (e.g., model checking) to verify pre-trained ML models against formal safety properties. Representative examples of this approach include the use of SMT-like solvers~\cite{dutta2018output,liu2019algorithms,sun2019formal,khedr2021peregrinn,ferlez2022fast,santa2022nnlander} and hybrid-system verification~\cite{fazlyab2019efficient,ivanov2019verisig,xiang2019reachable}. However, these techniques only assess a given ML-based planner's safety rather than design or train a safe agent.

Due to the lack of safety guarantees on the resulting ML-based planners, researchers proposed several techniques to \emph{restrict} the output of the ML models to a set of safe control actions. Such a set of safe actions can be obtained through Hamilton-Jacobi analysis~\cite{fisac2018general,fisac2021rss} and barrier certificates~\cite{abate2021hscc,pappas2021hscc,cheng2019end,matni2020cdc,taylor2020control,wang2018safe,cassandras2020adaptivecbf}. Unfortunately, methods of this type suffer from being computationally expensive, specific to certain controller structures, or requiring assumptions on the system model. Other techniques in this domain include synthesizing a safety layer (shield) based on model predictive control with the assumption of safe terminal sets~\cite{bastani2021rss,zeilinger2018cdc,zeilinger2021filter}, logically-constrained reinforcement learning~\cite{hasanbeig2019cdc,balakrishnan2019iros,alshiekh2018aaai}, and Lyapunov methods~\cite{berkenkamp2017safe,chow2018lyapunov,chow2019lyapunov} that focus on providing stability guarantees rather than safety or general temporal logic guarantees.

The idea of learning neurosymbolic models is studied in works~\cite{anderson2020neurosymbolic,verma2019imitation,bastani2018neurlps} that use NNs to guide the synthesis of control policies represented as short programs. The algorithms in~\cite{anderson2020neurosymbolic,verma2019imitation,bastani2018neurlps} train a NN controller, project it to the space of program languages, analyze the short programs, and lift the programs back to the space of NNs for further training. These works focus on tasks given during the training of NNs, and the final controller is a short program. Another line of related work is reported in~\cite{weiss2018extracting,carr2020verifiable}, which study the problem of extracting a finite-state controller from a recurrent neural network. Unlike the above works, we consider temporal logic specifications and unseen tasks, and our final planner is NNs in tandem with a finite-state MDP.

\section{Problem Formulation}
\label{sec:formulation}
\subsection{Notations}  
Let $\mathbb{R}$, $\mathbb{R}^+$, $\mathbb{N}$ be the set of real numbers, positive real numbers, and natural numbers, respectively. For a non-empty set $S$, let $2^S$ be the power set of $S$, $\mathbf{1}_S$ be the indicator function of $S$, and $\Int(S)$ be the interior of $S$. Furthermore, we use $S^n$ to denote the set of all finite sequences of length $n \in \mathbb{N}$ of elements in $S$. The product of two sets is defined as $S_1 \times S_2 := \{(s_1, s_2) | s_1 \in S_1, s_2 \in S_2\}$. Let $\norm{x}$ be the Euclidean norm of a vector $x \in \mathbb{R}^n$, $\norm{A}$ be the induced 2-norm of a matrix $A \in \mathbb{R}^{m \times n}$, and $\norm{A}_{\max} = \underset{i,j}{\max}|A_{ij}|$ be the max norm of a matrix $A$. Any Borel space $X$ is assumed to be endowed with a Borel $\sigma$-algebra denoted by $\mathcal{B}(X)$.

\subsection{Assumptions and Information Structure}
\label{subsec:assumption}
We consider a meta-RL setting that aims to train neural networks for controlling a robot to achieve tasks that were unseen during training. To be specific, we denote a task by a tuple $\mathcal{T} = (t, \varphi, \mathcal{W}, X_0)$, where $t$ captures the dynamic constraints of a robot (see Section~\ref{subsec:dynamcis}), 
$\varphi$ is a Linear Temporal Logic (LTL) formula that defines the mission for a robot to accomplish (see Section~\ref{subsec:spec}),
$\mathcal{W}$ is a workspace (or an environment) in which a robot operates, and $X_0$ contains the initials states of a robot. Furthermore, we use $J$ to denote a cost functional of controllers, and the cost of using a neural network $\NN$ is given by $J(\NN)$ (see Section~\ref{subsec:main}).


During training, we assume the availability of the cost functional $J$ and an approximation of the dynamical model $t$ (see Section~\ref{subsec:dynamcis} for details). The mission specification $\varphi$, the workspace $\mathcal{W}$, and the set of initial states $X_0$ are unknown during training and only become available at runtime. 
Despite the limited knowledge of tasks during training, we aim to design provably correct NNs for unseen tasks $\mathcal{T}$ while minimizing some given cost $J$.

\subsection{Dynamical Model}
\label{subsec:dynamcis}
We consider robotic systems that can be modeled as stochastic, discrete-time, nonlinear dynamical systems with a transition probability of the form:
\begin{equation}
    \label{eq:int_A}
    \pr(x^\prime \in A | x, u) = \int_A t(dx^\prime|x,u),
\end{equation}
where states of a robot $x \in X$ and control actions $u \in U$ are from continuous state and action spaces $X \subset \mathbb{R}^n$ and $U \subset \mathbb{R}^m$, respectively. In~\eqref{eq:int_A}, we use $t: \mathcal{B}(X) \times X \times U \rightarrow [0, 1]$ to denote a stochastic kernel that assigns to any state $x \in X$ and action $u \in U$ a probability measure $t(\cdot | x, u)$. Then, $\pr(x^\prime \in A | x, u)$ is the probability of reaching a subset $A \in \mathcal{B}(X)$ in one time step from state $x \in X$ under action $u \in U$. We assume that $t$ consists of a priori known nominal model $f$ and an unknown model-error $g$ capturing unmodeled dynamics. As a well-studied technique to learn unknown functions from data, we assume the model-error $g$ can be learned by a Gaussian Process (GP) regression model $\mathcal{GP}(\mu_g, \sigma^2_g)$, where  $\mu_g$ and $\sigma^2_g$ are the posterior mean and variance functions, respectively~\cite{GP}. Hence, we can re-write~\eqref{eq:int_A} as:
\begin{equation}
    \label{eq:dyn}
    \pr(x^\prime \in A | x, u) = f(x, u) + \int_A g(dx^\prime|x,u),
\end{equation}
which is an integral of normal distribution and hence can be easily computed.

We assume the nominal model $f$ is given during the NN training phase, while the model-error $g$ is evaluated at runtime, and hence the exact stochastic kernel $t$ only becomes known at runtime. This allows us to apply the trained NN to various robotic systems with different dynamics captured by the model error $g$.

\textbf{Remark:} We note that our algorithm does not require the knowledge of the function $f$ in a closed-form/symbolic representation. Access to a simulator would suffice. 



\subsection{Temporal Logic Specification and Workspace}
\label{subsec:spec}
A well-known weakness of RL and meta-RL algorithms is the difficulty in designing reward functions that capture the exact intent of designers~\cite{finn2016icml,hasanbeig2019cdc,balakrishnan2019iros}. Agent behavior that scores high according to a user-defined reward function may not be aligned with the user's intention, which is often referred to as ``specification gaming''~\cite{specGame}. 
To that end, we adopt the representation of an agent's mission in temporal logic specifications, which have been extensively demonstrated the capability to capture complex behaviors of robotic systems. 


In particular, we consider mission specifications defined in either bounded linear temporal logic (BLTL)~\cite{bltl} or syntactically co-safe linear temporal logic (scLTL)~\cite{scltl}. Let $AP$ be a finite set of atomic propositions that describe a robotic system's states with respect to a workspace $\mathcal{W}$. For example, these atomic propositions can describe the location of a robot with respect to the obstacles to avoid and the goal location to achieve. Given $AP$, any BLTL formula can be generated according to the following grammar: 
\begin{equation*}
    \varphi := \sigma\ |\ \neg \varphi\ |\ \varphi_1 \vee \varphi_2\ |\ \varphi_1\; \mathcal{U}_{[k_1, k_2]}\; \varphi_2
\end{equation*}
where $\sigma \in AP$ and time steps $k_1 < k_2$. Given the above grammar, we can define  $\varphi_1 \wedge \varphi_2 = \neg (\neg \varphi_1 \vee \neg \varphi_2)$, $false = \varphi \wedge \neg \varphi$, and $true = \neg false$. Furthermore, the bounded-time \emph{eventually} operator can be derived as $\lozenge_{[k_1, k_2]} \varphi = true\; \mathcal{U}_{[k_1, k_2]}\; \varphi$ and the bounded-time \emph{always} operator is given by $\square_{[k_1, k_2]} \varphi = \neg \lozenge_{[k_1, k_2]} \neg \varphi$. 

Given a set of atomic propositions $AP$, the corresponding alphabet is defined as $\mathbb{A} := 2^{AP}$, and a finite (infinite) word $\omega$ is a finite (infinite) sequence of letters from the alphabet $\mathbb{A}$, i.e., $\omega = \omega^{(0)} \omega^{(1)} \ldots \omega^{(H)} \in \mathbb{A}^{H+1}$. The satisfaction of a word $\omega$ to a specification $\varphi$ can be determined based on the semantics of BLTL~\cite{bltl}. Given a robotic system and an alphabet $\mathbb{A}$, let $L: X \rightarrow \mathbb{A}$ be a labeling function that assigns to each state $x \in X$ the subset of atomic propositions $L(x) \in \mathbb{A}$ that evaluate $true$ at $x$. Then, a robotic system's trajectory $\xi$ satisfies a specification $\varphi$, denoted by $\xi \models \varphi$, if the corresponding word satisfies $\varphi$, i.e., $L(\xi) \models \varphi$, where $\xi = x^{(0)} x^{(1)} \ldots x^{(H)} \in X^{H+1}$ and $L(\xi) = L(x^{(0)}) L(x^{(1)}) \ldots L(x^{(H)}) \in \mathbb{A}^{H+1}$. Similarly, we can consider scLTL specifications interpreted over infinite words based on the fact that any infinite word that satisfies a scLTL formula $\varphi$ contains a finite ``good'' prefix such that all infinite words that contain the prefix satisfy $\varphi$~\cite{scltl}. 


\noindent\emph{\textbf{Example 1} (Reach-avoid Specification)}: Consider a robot that navigates a workspace $\mathcal{W} = \{X_\text{goal}, O_1, \ldots, O_c\}$, where $X_\text{goal} \subset X$ is a set of goal states that the robot would like to reach and $O_1, \ldots, O_c \subset X$ are obstacles that the robot needs to avoid. The set of atomic propositions is given by $AP = \{x \in X_\text{goal}, x \in O_1, \ldots, x \in O_c\}$, where $x$ is the state of the robot. Then, a reach-avoid specification can be expressed as $\varphi = \varphi_\text{liveness} \wedge \varphi_\text{safety}$, where $\varphi_\text{liveness} = \lozenge_{[0, H]} (x \in X_\text{goal})$ requires the robot to reach the goal $X_\text{goal}$ in $H$ time steps and $\varphi_\text{safety} = \square_{[0, H]} \bigwedge_{i=1, \ldots, c} \neg (x \in O_i)$ specifies to avoid all the obstacles during the time horizon $H$. Let $\xi = x^{(0)} x^{(1)} \ldots x^{(H)}$ be a trajectory of the robot, then the reach-avoid specification $\varphi$ is interpreted as:
\begin{align*}    
    &\xi \models \varphi_{\text{liveness}} \Longleftrightarrow \exists k \in \{0,\ldots H\}, x^{(k)} \in X_\text{goal}, \\
    &\xi \models \varphi_{\text{safety}} \Longleftrightarrow \forall k \in \{0,\ldots H\}, \forall i \in \{1, \ldots, c\}, x^{(k)} \not\in O_i.
\end{align*}

\subsection{Neural Network}
\label{subsec:nn_controller}
To account for the stochastic behavior of a robot, we aim to design a state-feedback neural network $\NN: X \rightarrow U$ that can achieve temporal motion and task specifications $\varphi$.
An $F$-layer Rectified Linear Unit (ReLU) NN is specified by composing $F$ layer functions (or just layers). A layer $l$ with $\mathfrak{i}_l$ inputs and $\mathfrak{o}_l$ outputs is specified by a weight matrix $W^{(l)} \in \mathbb{R}^{\mathfrak{o}_l \times \mathfrak{i}_l}$ and a bias vector $b^{(l)} \in \mathbb{R}^{\mathfrak{o}_l}$ as follows:
\begin{equation}
    \label{eq:layer_fnc}
    L^{\theta^{(l)}}: z \mapsto \max\{ W^{(l)} z + b^{(l)}, 0 \}, 
\end{equation}
where the $\max$ function is taken element-wise, and $\theta^{(l)} \triangleq (W^{(l)}, b^{(l)})$ for brevity. Thus, an $F$-layer ReLU NN is specified by $F$ layer functions $\{L^{\theta^{(l)}} : l = 1, \dots, F\}$ whose input and output dimensions are composable: that is, they satisfy $\mathfrak{i}_{l} = \mathfrak{o}_{l-1}$, $l = 2, \dots, F$. Specifically:
\begin{equation}
	\NN^{\:\theta}(x) = (L^{\theta^{(F)}} \circ L^{\theta^{(F-1)}} \circ \dots \circ L^{\theta^{(1)}})(x),
\end{equation}
where we index a ReLU NN function by a list of parameters $\theta \triangleq (\theta^{(1)}, \dots, \theta^{(F)})$. As a common practice, we allow the output layer $ L^{\theta^{(F)}}$ to omit the $\max$ function. For simplicity of notation, we drop the superscript $\theta$ in $\NN^{\:\theta}$ whenever the dependence on $\theta$ is obvious.

\subsection{Main Problem}
\label{subsec:main}
We consider training a finite set (or a library) of ReLU NNs (during the offline phase) and designing a selection algorithm (during the online phase) that can select the correct NNs once the exact task $\mathcal{T} = (t, \varphi, \mathcal{W}, X_0)$ is revealed at runtime. Before formalizing the problem under consideration, we introduce the following notion of neural network composition.
\begin{definition}
    Given a set (or a library) of neural networks $\mathfrak{NN} = \{\NN_1, \NN_2, \ldots, \NN_d\}$ along with an activation map $\Gamma:X \to \{1, \ldots, d\}$, the composed neural network $\NN_{[\mathfrak{NN},\Gamma]}$ is defined as: $\NN_{[\mathfrak{NN},\Gamma]}(x) = \NN_{\Gamma(x)}(x)$.
\end{definition}

In other words, the activation map $\Gamma$ selects the NN that needs to be activated at each state $x \in X$. In addition to achieving the motion and task specifications, the neural network needs to minimize a given cost functional $J$.
The cost functional $J$ is defined as:
\begin{align}
\label{eq:def_J}
    J(\NN_{[\mathfrak{NN},\Gamma]}) = \int_X c(x, \NN_{[\mathfrak{NN},\Gamma]}(x)) d\mu^{\NN}(x),    
\end{align}
where $c: X \times U \rightarrow \mathbb{R}$ is a state-action cost function and $\mu^{\NN}$ is the distribution of states induced by the nominal dynamics $f$ in~\eqref{eq:dyn} under the control of $\NN_{[\mathfrak{NN},\Gamma]}$. As an example, the cost functional can be a controller's energy $J(\NN_{[\mathfrak{NN},\Gamma]}) = \int_X \norm{\NN_{[\mathfrak{NN},\Gamma]}(x)}^2 d\mu^{\NN}(x)$. 

Let $\xi_{\NN_{[\mathfrak{NN},\Gamma]}}^x$ be a closed-loop trajectory of a robot that starts from the state $x \in X_0$ and evolves under the composed neural network $\NN_{[\mathfrak{NN},\Gamma]}$. We define the problem of interest as follows:
\begin{problem}
    \label{prob:main}
    Given a cost functional $J$, train a library of ReLU neural networks $\mathfrak{NN}$, and compute an activation map $\Gamma$ at runtime when a task $\mathcal{T} = (t, \varphi, \mathcal{W}, X_0)$ is revealed,  such that the composed neural network minimizes the cost $J(\NN_{[\mathfrak{NN},\Gamma]})$ and satisfies the specification $\varphi$ with probability at least $p$, i.e., $\pr \left(\xi_{\NN_{[\mathfrak{NN},\Gamma]}}^x \models \varphi \right) \geq p$ for any $x \in X_0$.
\end{problem}


\subsection{Overview of the Neurosymbolic Framework} \label{subsec:overview}
Our approach to designing the NN-based planner $\NN_{[\mathfrak{NN},\Gamma]}$ can be split into two stages: offline training and runtime selection. During the offline training phase, our algorithm obtains a library of networks $\mathfrak{NN}$. At runtime, and to fulfill \emph{unseen} tasks using a \emph{finite} set of neural networks $\mathfrak{NN}$, our neurosymbolic framework bridges ideas from symbolic LTL-based planning and machine learning. Similar to symbolic LTL-based planning, our framework uses a hierarchical approach that consists of a ``high-level'' discrete planner and a ``low-level'' continuous controller~\cite{shoukry2017linear,fainekos2005hybrid,bhatia2010sampling}. The ``high-level'' discrete planner focuses on ensuring the satisfaction of the LTL specification. At the same time, the ``low-level'' controllers compute control actions that steer the robot to satisfy the ``high-level'' plan. Unlike symbolic LTL-based planners, our framework uses neural networks as low-level controllers, thanks to their ability to handle complex nonlinear dynamic constraints. In particular, the ``high-level'' planner chooses the activation map $\Gamma$ to activate particular neural networks.

Nevertheless, to ensure the correctness of the proposed framework, it is essential to ensure that each neural network in $\mathfrak{NN}$ satisfies some ``formal'' property. This ``formal'' property allows the high-level planner to abstract the capabilities of each of the neural networks in $\mathfrak{NN}$ and hence choose the correct activation map $\Gamma$. To that end, in Section~\ref{sec:formal_training}, we formulate the sub-problem of ``formal NN training'' that guarantees the trained NNs satisfy certain formal properties, and solve it efficiently by introducing a NN weight projection operator. The solution to the formal training is used in Section~\ref{subsec:offline} to obtain the library of networks $\mathfrak{NN}$ offline. The associated formal property of each NN is used in Section~\ref{subsec:runtime} to design the activation map $\Gamma$.

\section{Formal Training of NNs}
\label{sec:formal_training}
In this section, we study the sub-problem of training NNs that are guaranteed to obey certain behaviors. In addition to the classical gradient-descent update of NN weights, we propose a novel ``projection'' operator that ensures the resulting NN obeys the selected behavior. We provide a theoretical analysis of the proposed projection operator in terms of correctness and computational complexity. 

%


\subsection{Formulation of Formal Training}
\label{subsec:fomulation_formal_training}
We start by recalling that every ReLU NN represents a Continuous Piece-Wise Affine (CPWA) function~\cite{montufar2014number}. Let $\Psi_\text{CPWA}: X \rightarrow \mathbb{R}^m$ denote a CPWA function of the form:
\begin{equation}
    \label{eq:cpwa}
    \Psi_\text{CPWA}(x) = K_i^\prime x + b_i^\prime\quad \text{if}\ x \in \mathcal{R}_i,\ i =1, \ldots, L,
\end{equation}
where the collection of polytopic subsets $\{\mathcal{R}_1, \ldots, \mathcal{R}_L\}$ is a partition of the set $X \subset \mathbb{R}^n$ such that $\bigcup_{i = 1}^L \mathcal{R}_i = X$ and $\mathrm{Int}(\mathcal{R}_i) \cap \mathrm{Int}(\mathcal{R}_j) = \emptyset$ if $i \neq j$. We call each polytopic subset $\mathcal{R}_i \subset X$ a linear region, and denote by $\mathbb{L}_{\Psi_\text{CPWA}}$ the set of linear regions associated to $\Psi_\text{CPWA}$, i.e.,
$\mathbb{L}_{\Psi_\text{CPWA}} = \{\mathcal{R}_1, \ldots, \mathcal{R}_L\}$.
In this paper, we confine our attention to CPWA controllers (and hence neural network controllers) that are selected from a bounded polytopic set {$\mathcal{P}^{K} \times \mathcal{P}^{b} \subset \mathbb{R}^{m\times n} \times \mathbb{R}^m$}, i.e., we assume that $K_i^\prime \in \mathcal{P}^{K}$ and $b_i^\prime \in \mathcal{P}^{b}$. %
For simplicity of notation, we use $\mathcal{P}^{K \times b} \subset \mathbb{R}^{m \times (n+1)}$ to denote the polytopic set $\mathcal{P}^K \times \mathcal{P}^b$, and use $K_i(x)$ with a single parameter $K_i \in \mathcal{P}^{K \times b}$ to denote $K'_i x_i + b'_i$ with the pair $(K_i^\prime, b_i^\prime) = K_i$. 
 
Let $\mathcal{P} \subseteq \mathcal{P}^{K \times b}$ be a bounded polytopic subset of the parameters $K_i$, then with some abuse of notation, we use the same notation $\mathcal{P}$ to denote the subset of CPWA functions whose parameters $K_i$ are chosen from $\mathcal{P}$. In other words, a CPWA function $\Psi_\text{CPWA} \in \mathcal{P}$ if and only if $K_i \in \mathcal{P}$ at all linear regions $\mathcal{R}_i \in \mathbb{L}_{\Psi_\text{CPWA}}$, where the CPWA function $\Psi_\text{CPWA}$ is in the form of~\eqref{eq:cpwa}. 

Using this notation, we define the formal training problem that ensures the trained NNs belong to subsets of CPWA functions $\mathcal{P} \subseteq \mathcal{P}^{K \times b}$ as follows:
\begin{problem}
    \label{prob:formal_training}
    Given a bounded polytopic subset $q \subseteq X$, a bounded subset of CPWA functions $\mathcal{P} \subseteq \mathcal{P}^{K \times b}$, and a cost functional $J$, find NN weights $\theta^*$ such that:
    \begin{equation}
        \label{eq:formal_training}
        \theta^* = \underset{\theta}{\text{argmin}}\; J(\NN^{\:\theta})\quad \text{s.t.}\quad \NN^{\:\theta}|_q \in \mathcal{P}.
    \end{equation}
\end{problem}

In Problem~\ref{prob:formal_training}, we use $\NN^{\:\theta}|_q$ to denote the restriction of $\NN^{\:\theta}$ to the subset $q$, i.e., $\NN^{\:\theta}|_q: q \rightarrow \mathbb{R}^m$ is given by $\NN^{\:\theta}|_q(x) = \NN^{\:\theta}(x)$ for $x \in q$. Consider the CPWA function $\NN^{\:\theta}$ is in the form of~\eqref{eq:cpwa}, then the constraint $\NN^{\:\theta}|_q \in \mathcal{P}$ requires that $K_i \in \mathcal{P}$ whenever the corresponding linear region $\mathcal{R}_i$ intersects the subset $q$, i.e.:
\begin{equation}
    \label{eq:formal_training_equivalent}
    \NN^{\:\theta}|_q \!\in\! \mathcal{P} \Longleftrightarrow
    K_i \!\in\! \mathcal{P},\ \forall \mathcal{R}_i \!\in\! \{\mathcal{R} \!\in\! \mathbb{L}_{\NN^{\:\theta}} | \mathcal{R} \cap q \neq \emptyset\}.
\end{equation}



\subsection{NN Weight Projection}
\label{subsec:project_weights}
To solve Problem~\ref{prob:formal_training}, we introduce a NN weight projection operator that can be incorporated into the training of neural networks. Algorithm~\ref{alg:safe_train} outlines our procedure for solving Problem~\ref{prob:formal_training}. As a projected-gradient algorithm, Algorithm~\ref{alg:safe_train} alternates the gradient descent based training (line~\ref{line:contrained_train} in Algorithm~\ref{alg:safe_train}) and the NN weight projection (line~\ref{line:projection}-\ref{line:update} in Algorithm~\ref{alg:safe_train}) up to a pre-specified maximum iteration $\text{max\_iter}$. Given a subset of CPWA functions $\mathcal{P} \subseteq \mathcal{P}^{K \times b}$, we denote by $\Pi_{\mathcal{P}}$ the NN weight projection operator that enforces a network $\NN^{\:\theta}$ to satisfy $\NN^{\:\theta}|_q \in \mathcal{P}$, i.e., the constraints~\eqref{eq:formal_training_equivalent}. In the following, we formulate this NN weight projection operator $\Pi_{\mathcal{P}}$ as an optimization problem. 

\begin{algorithm}
    \caption{\textsc{Foramal-Train} ($q, \mathcal{P}, J$)}
    \label{alg:safe_train}
    \begin{algorithmic}[1]
        \STATE Initialize neural network $\NN^{\:\theta}$, $i=1$ 
        \WHILE{$i \leq \text{max\_iter}$} \label{line:max_iter}
            \STATE $\NN^{\:\theta} = \texttt{gradient-descent}(\NN^{\:\theta}, \mathcal{P}, J)$ \label{line:contrained_train}
            \STATE $\widehat{W}^{(F)}, \widehat{b}^{(F)} = \Pi_{\mathcal{P}} (\NN^{\:\theta})$ \label{line:projection}
            \STATE Set the output layer weights of $\NN^{\:\theta}$ be $\widehat{W}^{(F)}, \widehat{b}^{(F)}$ \label{line:update}
            \STATE $i = i+1$
        \ENDWHILE
        \STATE \textbf{Return} $\NN^{\:\theta}$ 
    \end{algorithmic}  
\end{algorithm}

Consider a neural network $\NN^{\:\theta}$ with $F$ layers, including $F-1$ hidden layers and an output layer. Let $W^{(F)}$ and $b^{(F)}$ be the weight matrix and the bias vector of the output layer, respectively, i.e.:
\begin{align}
    \label{eq:theta_before}
    \theta = \left(\theta^{(1)}, \ldots, \theta^{(F-1)}, \; (W^{(F)}, b^{(F)}) \right)
\end{align}
Then, the NN weight projection $\Pi_{\mathcal{P}}$ updates the output layer weights $W^{(F)}$, $b^{(F)}$ to $\widehat{W}^{(F)}$, $\widehat{b}^{(F)}$ (line~\ref{line:projection}-\ref{line:update} in Algorithm~\ref{alg:safe_train}). As a result, the projected NN weights $\widehat{\theta}$ are given by:
\begin{align}
    \label{eq:theta_after}
    \widehat{\theta} = \left(\theta^{(1)}, \ldots, \theta^{(F-1)}, \; (\widehat{W}^{(F)}, \widehat{b}^{(F)}) \right).
\end{align}
%
We formulate the NN weight projection operator $\Pi_{\mathcal{P}}$ as the following optimization problem: 
\begin{align}
    &\underset{\widehat{W}^{(F)}, \widehat{b}^{(F)}}{\text{argmin}}\; \underset{x \in q}{\max}\; \norm{{\NN^{\:\widehat{\theta}}(x) - \NN^{\:\theta}(x)}}_1 \label{eq:proj_prototype_obj} \\
    &\text{s.t.}\; \widehat{K}_i \in \mathcal{P},\ \forall \mathcal{R}_i \in \{\mathcal{R} \in \mathbb{L}_{\NN^{\:\theta}}\; |\; \mathcal{R} \cap q \neq \emptyset\}. \label{eq:proj_prototype_const}
\end{align}%
In the constraints~\eqref{eq:proj_prototype_const}, we use $\widehat{K}_i$ to denote the affine function parameters of the CPWA function $\NN^{\:\widehat{\theta}}$. 

The optimization problem~\eqref{eq:proj_prototype_obj}-\eqref{eq:proj_prototype_const} tries to minimize the change of the NN's outputs due to the weight projection, where the change is measured by the largest 1-norm difference between the outputs given by $\NN^{\:\widehat{\theta}}$ and $\NN^{\:\theta}$ across the subset $q \subseteq X$, i.e., $\underset{x \in q}{\max}\; |\!|{\NN^{\:\widehat{\theta}}(x) - \NN^{\:\theta}(x)}|\!|_1$. In the following two subsections, we first upper bound the objective function~\eqref{eq:proj_prototype_obj} in terms of the change of the NN's weights, and then show that the optimization problem~\eqref{eq:proj_prototype_obj}-\eqref{eq:proj_prototype_const} can be solved efficiently.


\subsection{Bounding the Change of Control Actions}
First, we note that it is common to omit the ReLU activation functions from the NN's output layer. Since the proposed projection operator only modifies the output layer weights, it is straightforward to show that the NN weight projection operator does not affect the set of linear regions, i.e., $\mathbb{L}_{\NN^{\:\widehat{\theta}}} = \mathbb{L}_{\NN^{\:\theta}}$, but only updates the affine functions defined over these regions. The following proposition shows the relation between the change in the NN's outputs and the change made in the output layer weights. The proof of this proposition can be found in Appendix~\ref{app:formal_training}.
%
%
\begin{proposition} 
    \label{prop:weight_relation}
    Consider two $F$-layer neural networks $\NN^{\:\theta}$ and $\NN^{\:\widehat{\theta}}$ where $\theta$ and $\widehat{\theta}$ are as defined in~\eqref{eq:theta_before}-\eqref{eq:theta_after}.
    Then, the largest difference in the NNs' outputs across a subset $q \subseteq X$ is upper bounded as follows:
    \begin{align}
        &\underset{x \in q}{\max}\; |\!|{\NN^{\:\widehat{\theta}}(x) - \NN^{\:\theta}(x)}|\!|_1 \label{eq:ub_output_diff} \\
        &\leq \underset{x \in \mathrm{Vert}\left(\mathbb{L}_{\NN^{\:\theta} \cap q} \right)}{\max}\; \sum_{i=1}^{m} \sum_{j=1}^{\mathfrak{o}_{F-1}} |\Delta W_{ij}^{(F)}| h_j(x) + \sum_{i=1}^{m} |\Delta b_i^{(F)}|. \notag
    \end{align}
\end{proposition}

In Proposition~\ref{prop:weight_relation}, $m$ is the dimension of the NN's output, $\Delta W_{ij}^{(F)}$ and $\Delta b_i^{(F)}$ are the $(i,j)$-th and the $i$-th entry of $\Delta W^{(F)} = \widehat{W}^{(F)} - W^{(F)}$ and $\Delta b^{(F)} = \widehat{b}^{(F)} - b^{(F)}$, respectively. With the notation of layer functions~\eqref{eq:layer_fnc}, we use a single function $h: \mathbb{R}^n \rightarrow \mathbb{R}^{\mathfrak{o}_{F-1}}$ to represent all the hidden layers, i.e., $h(x) = (L_{\theta^{(F-1)}} \circ L_{\theta^{(F-2)}} \circ \dots \circ L_{\theta^{(1)}})(x)$, where $\mathfrak{o}_{F-1}$ is the number of neurons in the $(F-1)$-layer (the last hidden layer). Furthermore, we use $\mathbb{L}_{\NN^{\:\theta} \cap q}$ to denote the intersected regions between the linear regions in $\mathbb{L}_{\NN^{\:\theta}}$ and the subset $q \subseteq X$, i.e., $\mathbb{L}_{\NN^{\:\theta} \cap q} = \{\mathcal{R} \cap q | \mathcal{R} \in \mathbb{L}_{\NN^{\:\theta}}, \mathcal{R} \cap q \neq \emptyset\}$. Let $\mathrm{Vert}(\mathcal{R})$ be the set of vertices of a region $\mathcal{R}$, then $\mathrm{Vert}(\mathbb{L}_{\NN^{\:\theta} \cap q}) = \bigcup_{\mathcal{R} \in \mathbb{L}_{\NN^{\:\theta} \cap q}} \mathrm{Vert}(\mathcal{R})$ is the set of vertices of all regions in $\mathbb{L}_{\NN^{\:\theta} \cap q}$.

\subsection{Efficient Computation of the NN Projection Operator}
Now, we focus on how to compute the NN weight projection operator $\Pi_{\mathcal{P}}$ efficiently. In particular, Proposition~\ref{prop:weight_relation} proposes a direct way to solve the intended projection operator. 
In order to minimize the change of the NN's outputs~\eqref{eq:proj_prototype_obj} due to the weight projection, we minimize its upper bound given by~\eqref{eq:ub_output_diff}. 
Accordingly, we compute the NN weight projection operator $\Pi_{\mathcal{P}}$ by solving following optimization problem:
\begin{align}
    &\underset{\widehat{W}^{(F)}, \widehat{b}^{(F)}}{\text{argmin}} \;\;  \underset{x \in \mathrm{Vert}\left(\mathbb{L}_{\NN^{\:\theta} \cap q}\right)}{\max} \sum_{i=1}^{m} \!\!\sum_{j=1}^{\mathfrak{o}_{F-1}}\!\! \big|\Delta W_{ij}^{(F)}\big| h_j(x) \!+\!\! \sum_{i=1}^{m} \big|\Delta b_i^{(F)}\big| \label{eq:proj_operator_obj} \\
    &\text{s.t.}\; \widehat{K}_i \in \mathcal{P},\ \forall \mathcal{R}_i \in \{\mathcal{R} \in \mathbb{L}_{\NN^{\:\theta}}\; |\; \mathcal{R} \cap q \neq \emptyset\}.  \label{eq:proj_operator_const}
\end{align}

The next result establishes the computational complexity of solving the optimization problem above. The proof of the proposition is given in Appendix~\ref{app:formal_training}.
\begin{proposition}
    \label{prop:formal_training_linear}
    The optimization problem~\eqref{eq:proj_operator_obj}-\eqref{eq:proj_operator_const}
    is a linear program.
\end{proposition}


While Proposition~\ref{prop:formal_training_linear} ensures that solving the optimization problem can be done efficiently, we note that identifying the set of linear regions $\mathbb{L}_{\NN^{\:\theta}}$ of a ReLU neural network $\NN^{\:\theta}$ needs to enumerate the hyperplanes represented by $\NN^{\:\theta}$. For shallow NNs and other special NN architectures, this can be done in polynomial time (e.g.,~\cite{ferlez2021cdc} uses a poset for the enumeration). For general NNs, identifying linear regions may not be polynomial time, but there exist efficient tools such as \texttt{NNENUM}~\cite{bak2020cav} that uses star sets to enumerate all the linear regions. Moreover, as we will show in the following sections, each NN in the library $\mathfrak{NN}$ can contain a limited number of weights (and hence a limited number of linear regions), but their combination leads to NNs with a large number of linear regions and hence capable of implementing complex functions.

We conclude this section with the following result whose proof follows directly from Proposition~\ref{prop:formal_training_linear} and the equivalence in~\eqref{eq:formal_training_equivalent}. 
\begin{theorem}
    \label{thm:formal_training}
    Given a bounded polytopic subset $q \subseteq X$ and a bounded subset of CPWA functions $\mathcal{P} \subseteq \mathcal{P}^{K \times b}$. Consider a neural network $\NN^{\:\theta}$ whose output layer weights are given by the NN weight projection operator $\Pi_{\mathcal{P}}$ (i.e., the solution to~\eqref{eq:proj_operator_obj}-\eqref{eq:proj_operator_const}). Then, the network $\NN^{\:\theta}$ satisfies the constraint in~\eqref{eq:formal_training}, i.e., $\NN^{\:\theta}|_q \in \mathcal{P}$. Furthermore, the optimization problem~\eqref{eq:proj_operator_obj}-\eqref{eq:proj_operator_const} 
    is a linear program.
\end{theorem}

\section{Neurosymbolic Learning Framework}
\label{sec:algorithm}
As discussed in Section~\ref{subsec:overview}, our approach to designing the NN-based planner $\NN_{[\mathfrak{NN},\Gamma]}$ and solving Problem~\ref{prob:main} is split into two stages: offline training and runtime selection. During the offline training phase, our algorithm obtains a library of networks $\mathfrak{NN}$, where each NN is trained using the formal training Algorithm~\ref{alg:safe_train}.  At runtime, when the exact task $\mathcal{T} = (t, \varphi, \mathcal{W}, X_0)$ is observed, we use dynamic programming (DP) to compute an activation map $\Gamma$, which selects a subset of the trained NNs and combines them into a single planner. We provide details on these two stages in the following two subsections separately.

  
\subsection{Offline Training of a Library $\mathfrak{NN}$}
\label{subsec:offline}
Similar to standard LTL-based motion planners~\cite{fainekos2005hybrid,fainekos2006translating,kress2007s,shoukry2017linear,tabuada2009verification,belta2017formal}, we partition the continuous state space $X \subset \mathbb{R}^n$ into a finite set of abstract states $\mathbb{X} = \{q_1, \ldots, q_N\}$, where each abstract state $q_i \in \mathbb{X}$ is an infinity-norm ball in $\mathbb{R}^n$ with a pre-specified diameter $\eta_q \in \mathbb{R}^+$ (see Section~\ref{sec:speedups} for the choice of $\eta_q$). The partitioning satisfies $X = \bigcup_{q \in \mathbb{X}} q$ and $\Int(q_i) \cap \Int(q_j) = \emptyset$ if $i \neq j$. Let $\abs: X \rightarrow \mathbb{X}$ map a state $x \in X$ to the abstract state $\abs(x) \in \mathbb{X}$ that contains $x$, i.e., $x \in \abs(x)$, and $\ct_\mathbb{X}: \mathbb{X} \rightarrow X$ map an abstract state $q \in \mathbb{X}$ to its center $\ct_\mathbb{X}(q) \in X$, which is well-defined since abstract states are inifinity-norm balls. With some abuse of notation, we denote by $q$ both an abstract state, i.e., $q \in \mathbb{X}$, and a subset of states, i.e., $q \subseteq X$.

As mentioned in the above section, we consider CPWA controllers (and hence neural network controllers) selected from a bounded polytopic set (namely a controller space) $\mathcal{P}^{K \times b} \subset \mathbb{R}^{m \times (n+1)}$. We partition the controller space $\mathcal{P}^{K \times b} \subset \mathbb{R}^{m \times (n+1)}$ into a finite set of controller partitions $\mathbb{P} = \{\mathcal{P}_1, \ldots, \mathcal{P}_M\}$ with a pre-specified grid size $\eta_\mathcal{P} \in \mathbb{R}^+$ (see Section~\ref{sec:speedups} for the choice of $\eta_\mathcal{P}$). Each controller partition $\mathcal{P}_i \in \mathbb{P}$ is an infinity-norm ball centered around some $K_i \in \mathcal{P}^{K \times b}$ such that $\mathcal{P}^{K \times b} = \bigcup_{\mathcal{P} \in \mathbb{P}} \mathcal{P}$ and $\Int(\mathcal{P}_i) \cap \Int(\mathcal{P}_j) = \emptyset$ if $i \neq j$. Let $\ct_\mathbb{P}: \mathbb{P} \rightarrow \mathcal{P}^{K \times b}$ map a controller partition $\mathcal{P} \in \mathbb{P}$ to its center $\ct_\mathbb{P}(\mathcal{P}) \in \mathcal{P}^{K \times b}$. As mentioned in Section~\ref{subsec:fomulation_formal_training}, we use the same notation $\mathcal{P}$ to denote both a subset of the parameters $K_i \in \mathcal{P}^{K \times b}$ and a subset of CPWA functions whose parameters $K_i$ are chosen from $\mathcal{P}$. 

Algorithm~\ref{alg:offline_train} outlines the training of a library of neural networks $\mathfrak{NN}$. Without knowing the exact robot dynamics (i.e., the stochastic kernel $t$), the workspace $\mathcal{W}$, and the specification $\varphi$, we use the formal training Algorithm~\ref{alg:safe_train} to train one neural network $\NN^{\:\theta}_{(q, \mathcal{P})}$ corresponding to each combination of controller partitions $\mathcal{P} \in \mathbb{P}$ and abstract states $q \in \mathbb{X}$ (line~\ref{line:formal_train} in Algorithm~\ref{alg:offline_train}). Thanks to the NN weight projection operator $\Pi_\mathcal{P}$, the neural networks $\NN^{\:\theta}_{(q, \mathcal{P})}$ satisfy the constraint in~\eqref{eq:formal_training}, i.e., $\NN^{\:\theta}_{(q, \mathcal{P})}|_q \!\in\! \mathcal{P}$. In the following, we use the notation $\NN_{(q, \mathcal{P})}$ by dropping the superscript $\theta$ for simplicity and refer to each neural network $\NN_{(q, \mathcal{P})}$ a local network. 

To minimize the cost functional $J$, we implement the training approach $\texttt{gradient-descent}$ (line~\ref{line:contrained_train} in Algorithm~\ref{alg:safe_train}) based on Proximal Policy Optimization (PPO)~\cite{ppo} with the reward function as follows:
\begin{equation}
    r(x, u) = -w_1 c(x, u) - w_2 \norm{u - \kappa(x)},
\end{equation}%
where $\kappa = \ct_\mathbb{P}(\mathcal{P})$ is the center of the assigned controller partition $\mathcal{P} \in \mathbb{P}$, $w_1, w_2 \in \mathbb{R}^+$ are pre-specified weights, and the state-action cost function $c: X \times U \rightarrow \mathbb{R}$ is from the definition of $J$ in~\eqref{eq:def_J}. Maximizing the above reward minimizes the cost $c(x, u)$ and encourages choosing controllers from the assigned controller partition~$\mathcal{P}$. We assume access to a ``nominal'' simulator (i.e., the nominal dynamics $f$ in~\eqref{eq:dyn}) for updating the robot states. Algorithm~\ref{alg:offline_train} returns a library $\mathfrak{NN}$ of $M \times N$ local networks, where $M$ and $N$ are the number of abstract states and the number of controller partitions, respectively. In Section~\ref{sec:speedups}, we reduce the number of local networks that need to be trained by employing transfer learning. 

\begin{algorithm}
    \caption{\textsc{Train-Library-NNs} ($\mathbb{X}, \mathbb{P}, J$)}
    \label{alg:offline_train}
    \begin{algorithmic}[1]
        \STATE $\mathfrak{NN} = \{\}$
        \FOR{$q \in \mathbb{X}$}
            \FOR{$\mathcal{P} \in \mathbb{P}$} 
                \STATE $\NN_{(q, \mathcal{P})} = \texttt{Formal-Train}(q, \mathcal{P}, J)$ \label{line:formal_train}
                \STATE $\mathfrak{NN} = \mathfrak{NN}\; \cup\; \{\NN_{(q, \mathcal{P})} \}$
            \ENDFOR
        \ENDFOR
        \STATE \textbf{Return} $\mathfrak{NN}$
    \end{algorithmic}  
\end{algorithm}

\subsection{Runtime Selection of Local NNs}
\label{subsec:runtime}
In this subsection, we present our selection algorithm used at runtime when an arbitrary task $\mathcal{T} = (t, \varphi, \mathcal{W}, X_0)$ is given. The selection algorithm assigns one local neural network in the set $\mathfrak{NN}$ to each abstract state $\{q_1, \ldots, q_N\}$ in order to satisfy the given specification $\varphi$. Given a stochastic kernel $t$, our algorithm first computes a finite-state Markov Decision Process (MDP) that captures the closed-loop behavior of the robot under \emph{all} possible CPWA controllers. Transitions in this finite-state MDP correspond to different subsets of CPWA functions in $\mathbb{P} = \{\mathcal{P}_1, \ldots, \mathcal{P}_M \}$. 
Thanks to the fact that the neural networks in the library $\mathfrak{NN}$ were trained using the formal training algorithm (Algorithm~\ref{alg:safe_train}), \emph{each neural network now represents a transition (symbol) in the finite-state MDP}. In other words, although neural networks are hard to interpret due to their construction, the formal training algorithm ensures the one-to-one mapping between these black-box neural networks and the transitions in the finite-state symbolic model.

Next, we use standard techniques in LTL-based motion planning to construct a finite-state automaton that captures the satisfaction of mission specifications $\varphi$. By analyzing the product between the finite-state MDP (that abstracts the robot dynamics) and the automaton corresponding to the specification $\varphi$, our algorithm decides which local networks in the set $\mathfrak{NN}$ need to be activated. We present details on the selection algorithm in the three steps below. 

\subsubsection*{\textbf{Step 1: Compute Symbolic Model}}
%
We construct a finite-state Markov decision process (MDP) $\hat{\Sigma} = (\mathbb{X}, \mathbb{X}_0, \mathbb{P}, \hat{t})$
    of the robotic system $\Sigma = (X, X_0, U, t)$ 
    as:
    \begin{itemize}
        \item $\mathbb{X} = \{q_1, \ldots, q_N\}$ is the set of abstract states;
        \item $\mathbb{X}_0 = \{q \in \mathbb{X}\; |\; q \subseteq X_0\}$ is the set of initial states;
        \item $\mathbb{P} = \{\mathcal{P}_1, \ldots, \mathcal{P}_M\}$ is the set of controller partitions;
        \item The transition probability from state $q \in \mathbb{X}$ to state $q^\prime \in \mathbb{X}$ with label $\mathcal{P} \in \mathbb{P}$ is given by:
        \begin{equation}
            \label{eq:integral_hat_t}
            \hat{t}(q^\prime | q, \mathcal{P}) = \int_{q^\prime} t(dx^\prime|z, \kappa(z))
        \end{equation}
        where $z=\ct_\mathbb{X}(q)$, $\kappa=\ct_\mathbb{P}(\mathcal{P})$.
    \end{itemize}
As explained in Section~\ref{subsec:dynamcis}, the integral~\eqref{eq:integral_hat_t} can be easily computed since the stochastic kernel $t(\cdot | x, u)$ is a normal distribution, and we show techniques to accelerate the construction of the symbolic model $\hat{\Sigma}$ in Section~\ref{sec:speedups}. Such finite symbolic models have been used heavily in state-of-the-art LTL-based controller synthesis. Nevertheless, and unlike state-of-the-art LTL-based controllers, the control alphabet in $\hat{\Sigma}$ is \emph{controller partitions} (i.e., subsets of CPWA functions). This is in contrast to LTL-based controllers in the literature (e.g.,~\cite{tabuada2009verification,belta2017formal}) that use subsets of control signals as their control alphabet.

We emphasize that our trained NN controllers are used to control the robotic system $\Sigma$ with continuous state and action spaces, and the theoretical guarantees that we provide in Section~\ref{sec:guarantees} are also for the robotic system $\Sigma$, not for the finite-state MDP $\hat{\Sigma}$. As the motivation to introduce the symbolic model $\hat{\Sigma}$, our approach provides correctness guarantees for the NN-controlled robotic system $\Sigma$ through (i) analyzing the behavior of the finite-state MDP $\hat{\Sigma}$ (in this section), and (ii) bounding the difference in behavior between the finite-state MDP $\hat{\Sigma}$ and the NN-controlled robotic system $\Sigma$ (in Section~\ref{sec:guarantees}). Critical to the latter step is the ability to restrict the NN's behavior thanks to the formal training proposed in Section~\ref{sec:formal_training}.

\subsubsection*{\textbf{Step 2: Construct Product MDP}}
Given a mission specification $\varphi$ encoded in BLTL or scLTL formula, we construct the equivalent deterministic finite-state automaton (DFA) $\mathcal{A}_\varphi = (S, S_0, \mathbb{A}, G, \delta)$ as follows:
    \begin{itemize}
        \item $S$ is a finite set of states;
        \item $S_0 \subseteq S$ is the set of initial states;
        \item $\mathbb{A}$ is an alphabet;
        \item $G \subseteq S$ is the accepting set;
        \item $\delta: S \times \mathbb{A} \rightarrow S$ is a transition function. 
    \end{itemize}
Such translation of BLTL and scLTL specifications to the equivalent DFA can be done using off-the-shelf tools (e.g.,~\cite{latvala2003spin,gerth1996}).


Given the finite-state MDP capturing the robot dynamics $\hat{\Sigma} = (\mathbb{X}, \mathbb{X}_0, \mathbb{P}, \hat{t})$ and the DFA $\mathcal{A}_\varphi = (S, S_0, \mathbb{A}, G, \delta)$ of the mission specification $\varphi$, we construct the product MDP $\hat{\Sigma} \otimes \mathcal{A}_\varphi = (\mathbb{X}^\otimes, \mathbb{X}_0^\otimes, \mathbb{P}, \mathbb{X}_G^\otimes, \hat{t}^\otimes)$ as follows:
    \begin{itemize}
        \item $\mathbb{X}^\otimes = \mathbb{X} \times S$ is a finite set of states;
        \item $\mathbb{X}_0^\otimes = \{(q_0, \delta(s_0, \hat{L}(q_0)) | q_0 \in \mathbb{X}_0, s_0 \in S_0\}$ is the set of initial states, where $\hat{L}: \mathbb{X} \rightarrow \mathbb{A}$ is the labeling function that assigns to each abstract state $q \in \mathbb{X}$ the subset of atomic propositions $\hat{L}(q) \in \mathbb{A}$ that evaluate $true$ at $q$;
        \item $\mathbb{P}$ is the set of controller partitions;
        \item $\mathbb{X}_G^\otimes = \mathbb{X} \times G$ is the accepting set;
        \item The transition probability from state $(q, s) \in \mathbb{X}^\otimes$ to state $(q^\prime, s^\prime) \in \mathbb{X}^\otimes$ under $\mathcal{P} \in \mathbb{P}$ is given by:
        \begin{align}
            \hat{t}^\otimes(q^\prime, s^\prime | q, s, \mathcal{P})=
            \begin{cases}
                \hat{t}(q^\prime | q, \mathcal{P})\ &\text{if } s^\prime = \delta(s, \hat{L}(q^\prime)) \\
                0\ &\text{else}. \notag
            \end{cases}
        \end{align}
    \end{itemize}



\subsubsection*{\textbf{Step 3: Select Local NNs by Dynamic Programming}}
Once constructed the product MDP $\hat{\Sigma} \otimes \mathcal{A}_\varphi$, the next step is to assign one local network $\NN_{(q, \mathcal{P})} \in \mathfrak{NN}$ to each abstract state $q \in \mathbb{X}$. In particular, the selection of NNs aims to maximize the probability of the finite-state MDP $\hat{\Sigma}$ satisfying the given specification $\varphi$. 
This can be formulated as finding the optimal policy that maximizes the probability of reaching the accepting set $\mathbb{X}_G^\otimes$ in the product MDP $\hat{\Sigma} \otimes \mathcal{A}_\varphi$. To that end, we define the optimal value functions $\hat{V}_k^*: \mathbb{X}^\otimes \rightarrow [0, 1]$ that map a state $(q, s) \in \mathbb{X}^\otimes$ to the maximum probability of reaching the accepting set $\mathbb{X}_G^\otimes$ in $H-k$ steps from the state $(q, s)$. When $k=0$, the optimal value function $\hat{V}_0^*$ yields the maximum probability of reaching the accepting set $\mathbb{X}_G^\otimes$ in $H$ steps, i.e., the maximum probability of $\hat{\Sigma}$ satisfying $\varphi$. The optimal value functions can be solved by the following dynamic programming recursion:
\begin{align}
    \hat{Q}_k(q, s, \mathcal{P}) &= \mathbf{1}_{G}(s) \label{eq:abst_Q}\\
    &+ \mathbf{1}_{S \setminus G}(s) \sum_{(q^\prime, s^\prime) \in \mathbb{X}^\otimes} \hat{V}_{k+1}^* (q^\prime, s^\prime) \hat{t}^\otimes(q^\prime, s^\prime | q, s, \mathcal{P}) \notag \\
    \hat{V}_k^*(q, s) &= \max_{\mathcal{P} \in \mathbb{P}}\; \hat{Q}_k(q, s, \mathcal{P}) \label{eq:abst_V}
\end{align} 
with the initial condition $\hat{V}_H^* (q, s) =  \mathbf{1}_G(s)$ for all $(q, s) \in \mathbb{X}^\otimes$, where $k = H, \ldots, 0$.

Algorithm~\ref{alg:runtime_dp} summarizes the above three steps for selecting local NNs. Given a task $\mathcal{T}=(t, \varphi, \mathcal{W}, X_0)$ at runtime, Algorithm~\ref{alg:runtime_dp} first computes the symbolic model $\hat{\Sigma}$ based on the stochastic kernel $t$, translates the mission specification $\varphi$ to a DFA $\mathcal{A}_\varphi$ using off-the-shelf tools, and constructs the product MDP $\hat{\Sigma} \otimes \mathcal{A}_\varphi$ (line~\ref{line:mdp}-\ref{line:product} in Algorithm~\ref{alg:runtime_dp}). Then, Algorithm~\ref{alg:runtime_dp} solves the optimal policy for the product MDP $\hat{\Sigma} \otimes \mathcal{A}_\varphi$ using the DP recursion~\eqref{eq:abst_Q}-\eqref{eq:abst_V} (line~\ref{line:dp_start}-\ref{line:dp_end} in Algorithm~\ref{alg:runtime_dp}). At time step $k$, the optimal controller partition $\mathcal{P}^*$ at state $(q, s) \in \mathbb{X}^\otimes$ is given by the maximizer of $\hat{Q}_k(q, s, \mathcal{P})$ (line~\ref{line:p_star} in Algorithm~\ref{alg:runtime_dp}). The last step is to assign a corresponding neural network to be applied given the robot states $x \in X$ and the DFA states $s \in S$. To that end, let:
$$\Gamma_k(x, s) = \hat{\Gamma}_k(\abs(x), s),$$ 
where $\hat{\Gamma}_k$ maps the product MDP's states $(q, s) \in \mathbb{X}^\otimes$ to neural network's indices $(q, \mathcal{P}^*)$ (line~\ref{line:gamma} in Algorithm~\ref{alg:runtime_dp}). In other words, given the robot states $x \in X$ and the DFA states $s \in S$ at time step $k$, we first find the abstract state $q \in \mathbb{X}$ that contains $x$, i.e., $q = \abs(x)$, and then use the neural network $\NN_{(q, \mathcal{P}^*)} \in \mathfrak{NN}$ to control the robot at $x$, where $\hat{\Gamma}_k(q, s) = (q, \mathcal{P}^*)$. Recall that the neural networks in $\mathfrak{NN}$ are indexed as $(q, \mathcal{P})$ and hence the function $\Gamma(x,s) = \hat{\Gamma}_k(\abs(x), s)$ computes such indices. 

\begin{algorithm}
    \caption{\textsc{Runtime-Select} $\left(\mathcal{T}=(t, \varphi, \mathcal{W}, X_0)\right)$}
    \label{alg:runtime_dp}
    \begin{algorithmic}[1]
        \STATE Compute the symbolic model $\hat{\Sigma} = (\mathbb{X}, \mathbb{X}_0, \mathbb{P}, \hat{t})$ \label{line:mdp}
        \STATE Translate $\varphi$ to a DFA $\mathcal{A}_\varphi = (S, S_0, \mathbb{A}, G, \delta)$
        \STATE Construct the product MDP $\hat{\Sigma} \otimes \mathcal{A}_\varphi$ \label{line:product}
        \FOR{$(q, s) \in \mathbb{X}^\otimes$} \label{line:dp_start}
            \STATE $\hat{V}_H^*(q, s) =  \mathbf{1}_{G} (s)$
        \ENDFOR
        \STATE $k = H-1$
        \WHILE{$k \geq 0$}
            \FOR{$(q, s) \in \mathbb{X}^\otimes$}
                \IF{$s \in G$}
                    \STATE $\hat{Q}_k(q, s, \mathcal{P}) = 1$
                \ELSE
                    \STATE $\hat{Q}_k(q, s, \mathcal{P}) =\hspace{-3mm} \underset{(q^\prime, s^\prime) \in \mathbb{X}^\otimes}{\sum} \hat{V}_{k+1}^* (q^\prime, s^\prime) \hat{t}^\otimes(q^\prime, s^\prime | q, s, \mathcal{P})$
                \ENDIF
                \STATE $\hat{V}_k^*(q, s) = \underset{\mathcal{P} \in \mathbb{P}}{\max}\ \hat{Q}_k(q, s, \mathcal{P})$
                \STATE $\mathcal{P}^* = \underset{\mathcal{P} \in \mathbb{P}}{\text{argmax}}\ \hat{Q}_k(q, s, \mathcal{P})$ \label{line:p_star}
                \STATE $\hat{\Gamma}_{k}(q, s) = (q, \mathcal{P}^*)$ \label{line:gamma}
            \ENDFOR
            \STATE $k = k-1$
        \ENDWHILE \label{line:dp_end}
        \STATE \textbf{Return} $\{\hat{\Gamma}_{k}\}_{k \in \{0, \ldots, H-1\}}, \hat{V}^*_0, \hat{\Sigma} \otimes \mathcal{A}_\varphi$
    \end{algorithmic}  
\end{algorithm}

\subsection{Toy Example}
\label{subsec:toy}

\begin{figure*}[!ht]
    \center
    \resizebox{.99\textwidth}{!}{
    \begin{tabular}{cccc}
        \includegraphics[height=0.4\textwidth]{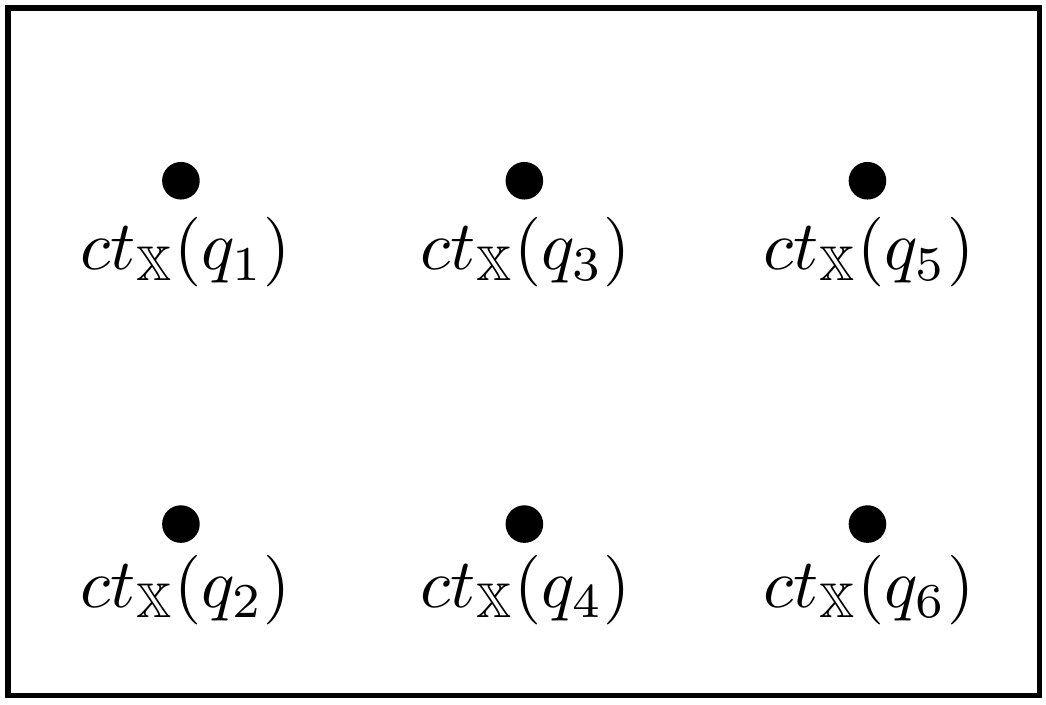}&
        \includegraphics[height=0.4\textwidth]{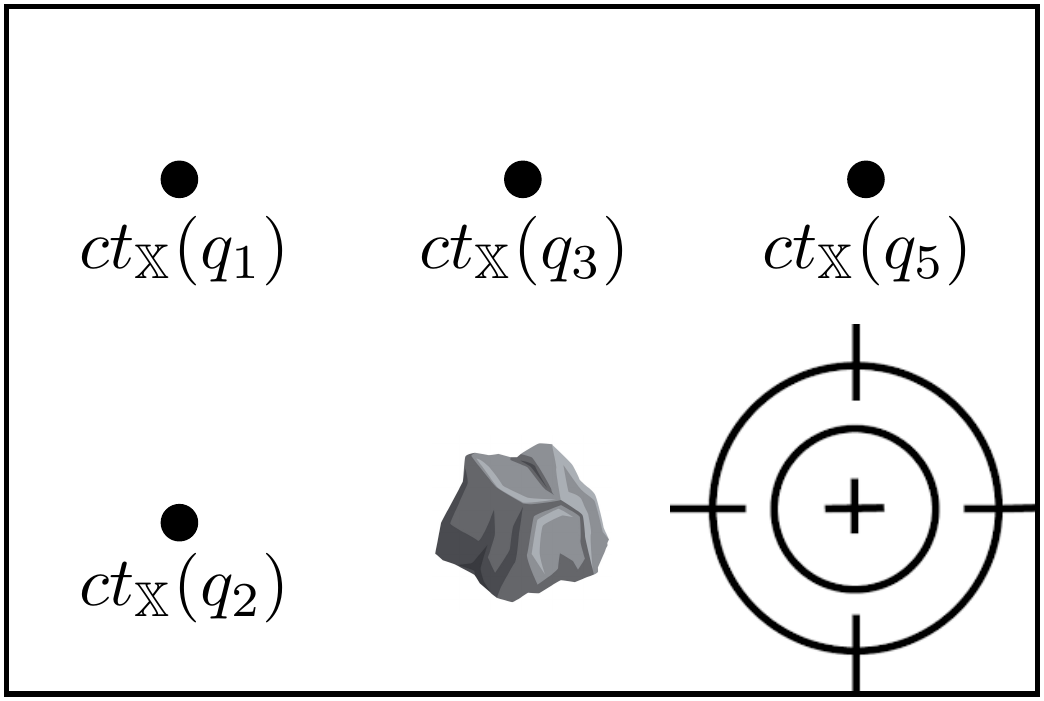}&
        \includegraphics[height=0.4\textwidth]{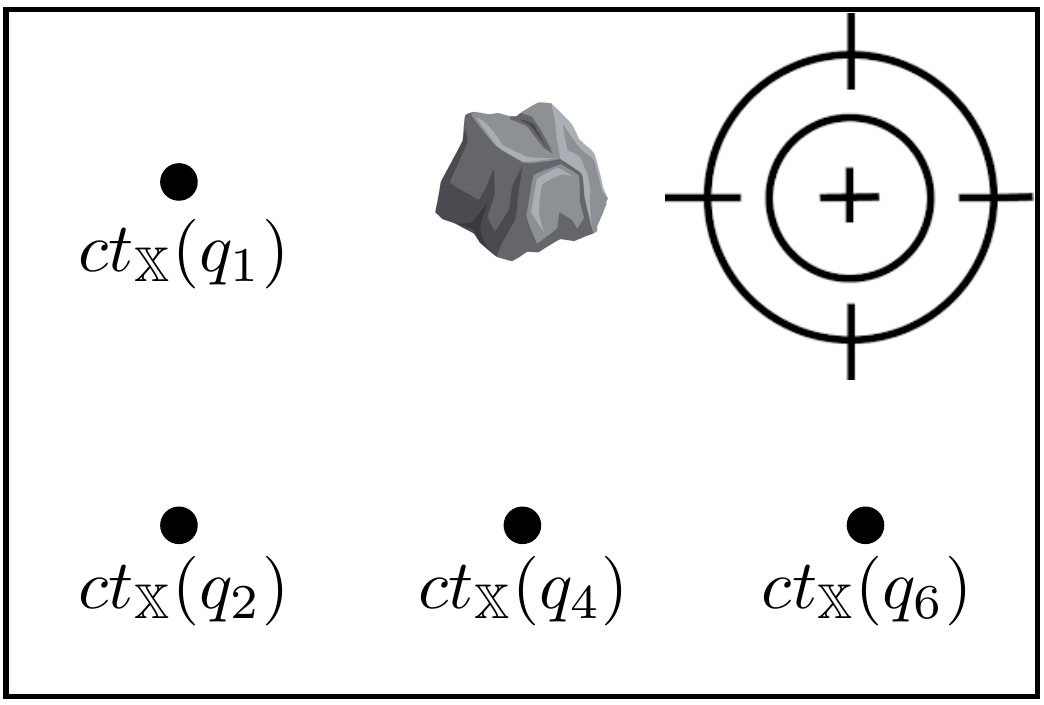}&
        \includegraphics[height=0.4\textwidth]{frame_new_3} 
        \\
        \includegraphics[height=0.4\textwidth]{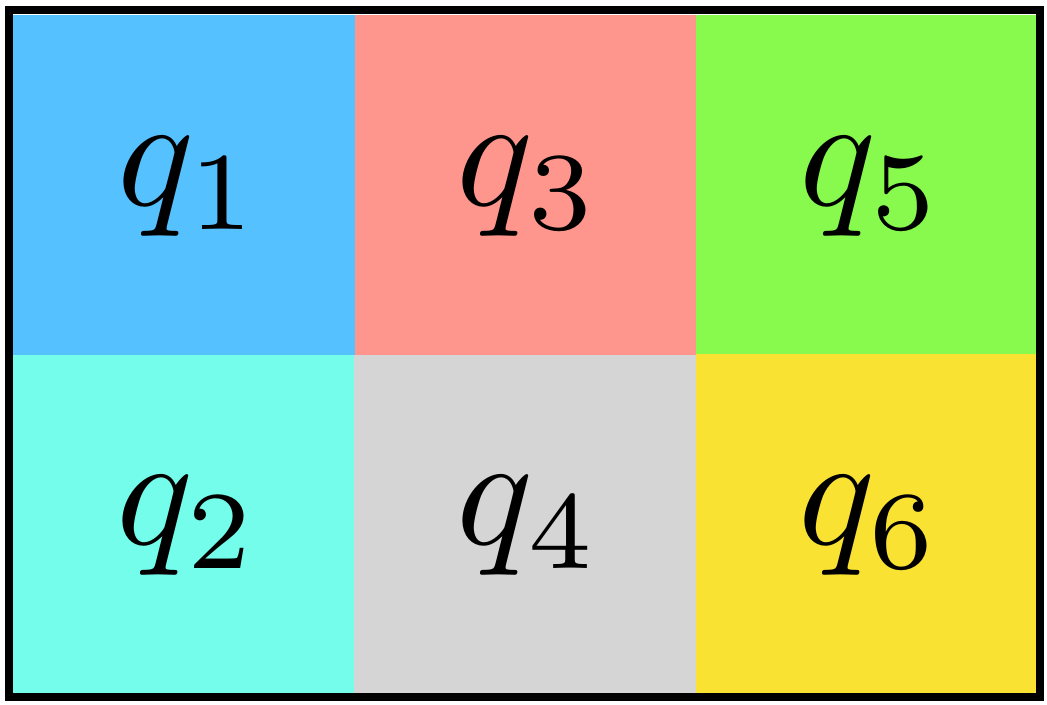}&
        \includegraphics[height=0.4\textwidth]{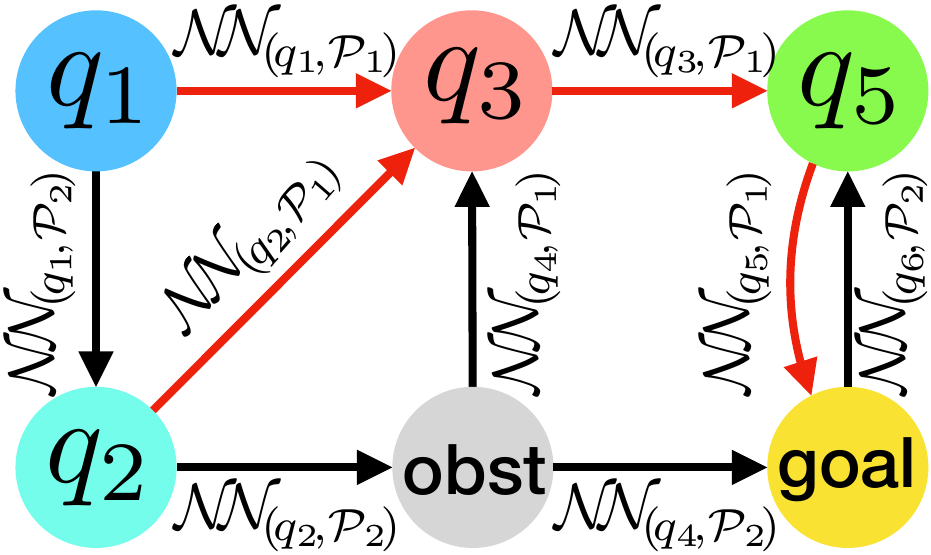}&
        \includegraphics[height=0.4\textwidth]{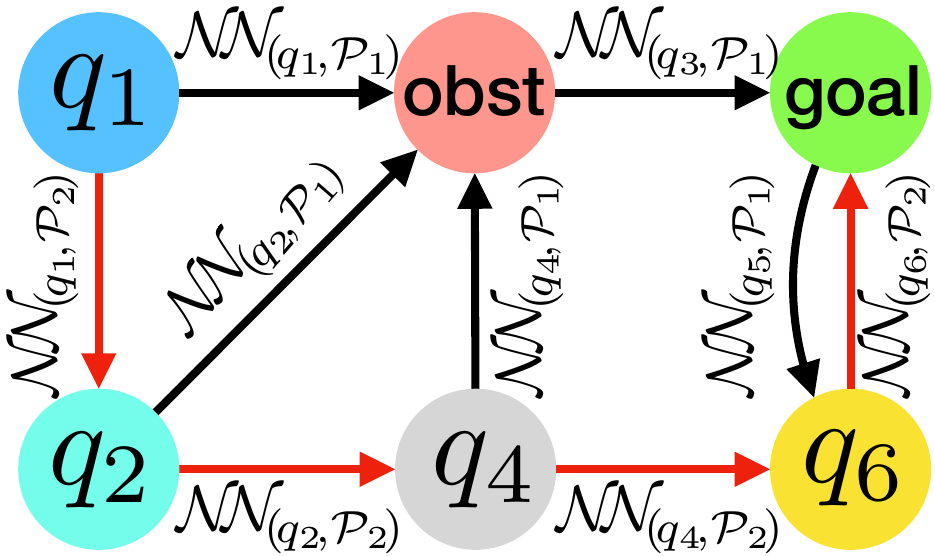}&
        \includegraphics[height=0.4\textwidth]{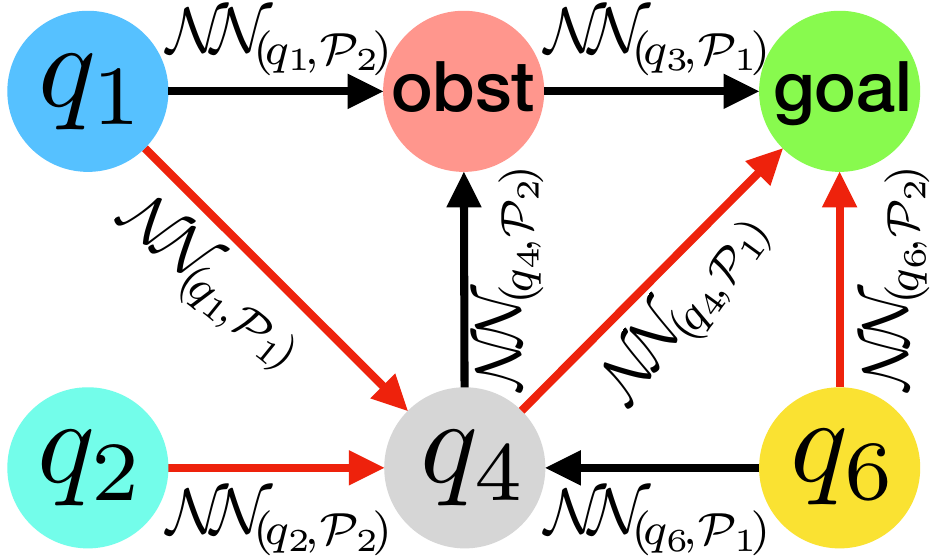}
        \\
        &&&
        \\
        \Huge{(a)} & \Huge{(b)} & \Huge{(c)} & \Huge{(d)}
    \end{tabular}  
    }  
    \caption{A toy example of a robot that navigates a two-dimensional workspace and needs to satisfy reach-avoid specifications $\varphi = \varphi_\text{liveness} \wedge \varphi_\text{safety}$ (see more details in Section~\ref{subsec:toy}).} 
    \label{fig:frame}  
\end{figure*}

We conclude this section by providing a toy example in Figure~\ref{fig:frame}. Consider a mobile robot that navigates a two-dimensional workspace. We partition the state space $X \subset \mathbb{R}^2$ into six abstract states $\mathbb{X} = \{q_1, \ldots, q_6\}$ and discretize the controller space $\mathcal{P}^{K \times b}$ into two controller partitions $\mathbb{P} = \{\mathcal{P}_1, \mathcal{P}_2\}$. Figure~\ref{fig:frame} (a) shows the state space (top) and the abstract states $q_1, \ldots, q_6$ resulted from the partitioning (bottom), where the centers of abstract states are $\ct_\mathbb{X}(q_1), \ldots, \ct_\mathbb{X}(q_6)$.

During the offline training (Section~\ref{subsec:offline}), we use the formal training Algorithm~\ref{alg:safe_train} to obtain a library $\mathfrak{NN}$ consisting of 12 neural networks, i.e., $\mathfrak{NN} = \{\NN_{(q_i, \mathcal{P}_j)} | i \in \{1, \ldots, 6\}, j \in \{1, 2\} \}$. 


We consider three different tasks $\mathcal{T}_1$, $\mathcal{T}_2$, and $\mathcal{T}_3$ that only become available at runtime after all the neural networks in $\mathfrak{NN}$ have been trained. Figure~\ref{fig:frame} (b), (c), and (d) show the workspaces for these three tasks, respectively. The specifications for these three tasks are 
$\varphi_1 = \lozenge_{[0, 3]} \left(x \in q_6\right) \land \square_{[0, 3]} \neg \left(x \in q_4\right)$, $\varphi_2 = \lozenge_{[0, 4]} \left(x \in q_5\right) \land \square_{[0, 4]} \neg \left(x \in q_3\right)$, and $\varphi_3 = \lozenge_{[0, 3]} \left(x \in q_5\right) \land \square_{[0, 3]} \neg \left(x \in q_3\right)$, respectively. 
Finally, the three tasks have different robot dynamics $t$. 
Figure~\ref{fig:frame} (b)-(d) also depict the transitions in the resulting symbolic models, where we assume that all the transition probabilities $\hat{t}$ are $1$ for simplicity (the transition probabilities $\hat{t}$ are computed as the integral of $t$ in~\eqref{eq:integral_hat_t}). Thanks to the formal training Algorithm~\ref{alg:safe_train}, the neural networks in $\mathfrak{NN}$ are guaranteed to be members of the CPWA functions in $\{\mathcal{P}_1, \mathcal{P}_2\}$. Hence, we label the transitions in the MDPs in Figure~\ref{fig:frame} (b)-(d) using $\NN_{(q_i, \mathcal{P}_j)}$ instead of $\{\mathcal{P}_1, \mathcal{P}_2\}$. While the transitions in the MDPs in Figure~\ref{fig:frame} (b) and (c) are the same, the MDP in Figure~\ref{fig:frame} (d) is different from that in Figure~\ref{fig:frame} (b) and (c) due to the difference in the robot dynamics in this task.


When the tasks $\mathcal{T}_1$, $\mathcal{T}_2$, and $\mathcal{T}_3$ become available, we use the runtime selection algorithm (Algorithm~\ref{alg:runtime_dp}) to obtain the selection functions $\Gamma_k$. In Figure~\ref{fig:frame} (b)-(d), the selected NNs are the labels of the transitions marked in red. For example, in Figure~\ref{fig:frame} (b), our algorithm selects $\NN_{(q_1,\mathcal{P}_1)}$ to be used at all states $x \in q_1$. It is clear from the figures that the selected NNs are guaranteed to satisfy the given specifications $\varphi_1$, $\varphi_2$, and $\varphi_3$, respectively, regardless of the difference in the workspaces and robot dynamics.

\section{Theoretical Guarantees}
\label{sec:guarantees}
In this section, we study the theoretical guarantees of the proposed approach. We first provide a probabilistic guarantee for our NN-based planners on satisfying mission specifications given at runtime, then bound the difference between the NN-based planner and the optimal controller that maximizes the probability of satisfying the given specifications. The proof of the theoretical guarantees (Theorem~\ref{thm:nn_v} and Theorem~\ref{thm:nn_optimal}) can be found in Appendix~\ref{app:guarantees}.

\subsection{Generalization to Unseen Tasks}
For an arbitrary task $\mathcal{T} = (t, \varphi, \mathcal{W}, X_0)$, let $\NN_{[\mathfrak{NN},\Gamma]}$ be the corresponding NN-based planner, where the library of networks $\mathfrak{NN}$ is trained by Algorithm~\ref{alg:offline_train} without knowing the task $\mathcal{T}$, and the activation map $\Gamma$ denotes the time-dependent functions $\Gamma_k$ obtained from Algorithm~\ref{alg:runtime_dp}. As a key feature of $\NN_{[\mathfrak{NN},\Gamma]}$, the activation map $\Gamma$ selects NNs based on both the robot states and the states of the $\mathcal{A}_\varphi$  DFA. This allows the NN-based planner $\NN_{[\mathfrak{NN},\Gamma]}$ to take into account the specification $\varphi$ by tracking states of the DFA $\mathcal{A}_\varphi$. In comparison, a single state-feedback neural network $\NN: X \rightarrow U$ is not able to track the DFA states and hence cannot be trained to satisfy BLTL or scLTL specifications in general.

We denote by $\xi_{\NN_{[\mathfrak{NN},\Gamma]}}^{(x, s)}$ the closed-loop trajectory of a robot under the NN-based planner $\NN_{[\mathfrak{NN},\Gamma]}$ with the robot starting from state $x \in X_0$ and the DFA $\mathcal{A}_\varphi$ starting from state $s \in S_0$. Notice that though the symbolic model $\hat{\Sigma}$ is a finite-state MDP, the NN-based planner $\NN_{[\mathfrak{NN},\Gamma]}$ is used to control the robotic system $\Sigma$ with continuous state and action spaces. The following theorem provides a probabilistic guarantee for the NN-controlled robotic system to satisfy mission specifications given at runtime. 
\begin{theorem}
    \label{thm:nn_v}
    Let $\hat{V}^*_0$ be the optimal value function returned by Algorithm~\ref{alg:runtime_dp}. For arbitrary states $x \in X_0$ and $s \in S_0$, the probability of the closed-loop trajectory $\xi_{\NN_{[\mathfrak{NN},\Gamma]}}^{(x, s)}$ satisfying the given mission specification $\varphi$ is bounded as follows: 
    \begin{equation}
        \label{eq:nn_v}
        \left| \pr \left(\xi_{\NN_{[\mathfrak{NN},\Gamma]}}^{(x, s)} \models \varphi \right) - \hat{V}^*_0(q, s) \right| \leq  H Z \Delta^\NN
    \end{equation}
    where $q = \abs(x)$ and 
    \begin{equation}
        \label{eq:delta_nn}
        \Delta^\NN = \max_{i \in \{1, \ldots, N\}} \left(\Lambda_i \eta_q + B_i L_i \eta_q + \sqrt{m(n+1)} \mathcal{L}_X B_i \eta_\mathcal{P} \right).
    \end{equation}
\end{theorem}

Recall that $\eta_q$ and $\eta_\mathcal{P}$ are the grid sizes used for partitioning the state space and the controller space, respectively. The upper bound $H Z \Delta^\NN$ in Theorem~\ref{thm:nn_v} can be arbitrarily small by tuning the grid sizes $\eta_q$ and $\eta_\mathcal{P}$. In~\eqref{eq:nn_v}-\eqref{eq:delta_nn}, $H$ is the time horizon, $N = |\mathbb{X}|$ is the number of abstract states, and $Z = |S|$ is the number of the $\mathcal{A}_\varphi$ DFA states. The parameters $\Lambda_i$ and $B_i$ are given by $\Lambda_i = \int_X \lambda_i(y) \mu(dy)$ and $B_i = \int_X \beta_i(y) \mu(dy)$, where $\lambda_i(y)$ and $\beta_i(y)$ are the Lipschitz constants of the stochastic kernel $t: \mathcal{B}(X) \times X \times U \rightarrow [0, 1]$, i.e., $\forall x, x^\prime \in q_i$, $\forall u \in U$:
\begin{equation*}
    |t(dy|x^\prime, u) - t(dy|x, u)| \leq \lambda_i (y) \norm{x^\prime - x} \mu(dy),
\end{equation*}
and $\forall x \in q_i$, $\forall u, u^\prime \in U$:
\begin{equation*}
    |t(dy|x, u^\prime) - t(dy|x, u)| \leq \beta_i (y) \norm{u^\prime - u} \mu(dy).
\end{equation*}
Furthermore, $L_i$ is the Lipschitz constant of the local neural networks at abstract state $q_i \in \mathbb{X}$, i.e., $\forall \mathcal{P} \in \mathbb{P}$, $\forall x, x^\prime \in q_i$:
\begin{equation*}
    \norm{\NN_{(q_i, \mathcal{P})}(x) - \NN_{(q_i, \mathcal{P})}(x^\prime)} \leq L_i \norm{x - x^\prime}.
\end{equation*}
Finally, $\underset{x \in X}{\sup} \norm{x} \leq \mathcal{L}_X$, $\underset{K \in \mathcal{P}^{K \times b}}{\sup} \norm{K} \leq \mathcal{L}_\mathcal{P}$, and $n$, $m$ are the dimensions of $X \subset \mathbb{R}^n$, $U \subset \mathbb{R}^m$, respectively.

\subsection{Optimality Guarantee}
Next, we compare our NN-based planner $\NN_{[\mathfrak{NN},\Gamma]}$ with the optimal controller (not necessarily a neural network) that maximizes the probability of satisfying the given specification $\varphi$. To that end, we provide an upper bound on the difference in the probabilities of satisfying $\varphi$ without explicit computing of the optimal controller. Let $\mathcal{C}_\varphi^*: X \times S \rightarrow U$ be the optimal controller and $\xi_{\mathcal{C}_\varphi^*}^{(x,s)}$ be the closed-loop trajectory of the robotic system $\Sigma = (X, X_0, U, t)$ controlled by $\mathcal{C}_\varphi^*$. Similar to the NN-based planner $\NN_{[\mathfrak{NN},\Gamma]}$, the optimal controller $\mathcal{C}_\varphi^*$ applies to the robotic system $\Sigma$ with continuous state and action spaces, and takes the DFA states $s \in S$ into consideration when computing control actions. Synthesizing the optimal controller $\mathcal{C}_\varphi^*$ for a mission specification $\varphi$ is computationally prohibitive due to the continuous state and action spaces. Without explicitly computing $\mathcal{C}_\varphi^*$, the following theorem tells how close our NN-based planner $\NN_{[\mathfrak{NN},\Gamma]}$ is to the optimal controller $\mathcal{C}_\varphi^*$ in terms of satisfying the specification $\varphi$. By tuning the grid sizes $\eta_q$ and $\eta_\mathcal{P}$, our NN-based planner $\NN_{[\mathfrak{NN},\Gamma]}$ can be arbitrarily close to the optimal controller $\mathcal{C}_\varphi^*$. 
\begin{theorem}
    \label{thm:nn_optimal}
    For arbitrary states $x \in X_0$ and $s \in S_0$, the difference in the probabilities of the closed-loop trajectories $\xi_{\NN_{[\mathfrak{NN},\Gamma]}}^{(x, s)}$ and $\xi_{\mathcal{C}_\varphi^*}^{(x,s)}$ satisfying the given mission specification $\varphi$ is upper bounded as follows: 
    \begin{equation}
        \label{eq:nn_optimal}
        \left| \pr \left(\xi_{\NN_{[\mathfrak{NN},\Gamma]}}^{(x, s)} \models \varphi \right) - \pr \left(\xi_{\mathcal{C}_\varphi^*}^{(x, s)} \models \varphi \right) \right| \leq  H Z (\Delta^\NN + \Delta^*)
    \end{equation}
    where $\Delta^\NN$ is given by~\eqref{eq:delta_nn} and 
    \begin{equation}
        \label{eq:delta_optimal}
        \Delta^* = \max_{i \in \{1, \ldots, N\}} \left(\Lambda_i \eta_q + B_i \mathcal{L}_\mathcal{P} \eta_q + 2 \sqrt{m(n+1)} \mathcal{L}_X B_i \eta_\mathcal{P} \right).
    \end{equation}
\end{theorem}

\section{Effective Adaptation}
\label{sec:speedups}
In this section, we focus on practical issues of the proposed approach and present some key elements for performance improvement while maintaining the same theoretical guarantees as Section~\ref{sec:guarantees}. Firstly, we show that the proposed composition of neural networks leads to an effective way to adapt previous learning experiences to unseen tasks. In particular, instead of training the whole library of neural networks $\mathfrak{NN}$ in Algorithm~\ref{alg:offline_train}, we only train a subset of networks $\mathfrak{NN}_\text{part} \subseteq \mathfrak{NN}$ based on tasks provided for training. Obtaining this subset $\mathfrak{NN}_\text{part}$ can be viewed as a systematic way to store learning experiences, which are adapted to unseen tasks via transfer learning (see Section~\ref{subsec:transfer}). Secondly, we propose a data-driven approach to accelerate the construction of the symbolic model $\hat{\Sigma}$ (see Section~\ref{subsec:learn_abstract}). Finally, we comment on the choice of grid sizes $\eta_q$ and $\eta_\mathcal{P}$ for partitioning the state and action spaces (see Section~\ref{subsec:adaptive_partitioning}).

\subsection{Accelerate by Transfer Learning}
\label{subsec:transfer}
Consider a meta-RL problem with a set of training tasks $\{\mathcal{T}_1, \mathcal{T}_2, \ldots, \mathcal{T}_d\}$ that are provided for training neural networks in the hope of fast adaptation to unseen tasks $\mathcal{T}_\text{test}$ during the test phase, where each task is a tuple $\mathcal{T} = (t, \varphi, \mathcal{W}, X_0)$ as defined before. We consider the problem of how to leverage the learning experiences from the training tasks to accelerate the learning of the unseen test tasks. Our intuition is that when the training tasks have enough variety, the local behavior for fulfilling a test task $\mathcal{T}_\text{test}$ should be close to the local behavior for fulfilling some training task $\mathcal{T}_\text{train} \in \{\mathcal{T}_1, \mathcal{T}_2, \ldots, \mathcal{T}_d\}$. In other words, the controller needed by a robot to fulfill the test task $\mathcal{T}_\text{test}$ should be close to the controller used for fulfilling some training task $\mathcal{T}_\text{train} \in \{\mathcal{T}_1, \mathcal{T}_2, \ldots, \mathcal{T}_d\}$, where the training task $\mathcal{T}_\text{train}$ can be \emph{different} in different subsets of the state space $X$. This is more general than the prevalent assumption in the meta-RL literature that the test task's controller is close to the \emph{same} training task's controller everywhere in the state space. As a result, our approach requires less variety of the training tasks $\{\mathcal{T}_1, \mathcal{T}_2, \ldots, \mathcal{T}_d\}$ for fast adaptation to unseen tasks. 

The form of the composed NN-based planner $\NN_{[\mathfrak{NN},\Gamma]}$ provides a systematic way to store learning experiences from all the training tasks and enables to select which training task should be adapted to the test task based on the current state of the robot. Given a set of training tasks $\{\mathcal{T}_1, \mathcal{T}_2, \ldots, \mathcal{T}_d\}$, Algorithm~\ref{alg:train_transfer} trains a subset of local networks $\mathfrak{NN}_\text{part} \subseteq \mathfrak{NN}$ suggested by the training tasks. For each training task $\mathcal{T}_\text{train} \in \{\mathcal{T}_1, \mathcal{T}_2, \ldots, \mathcal{T}_d\}$, Algorithm~\ref{alg:train_transfer} first calls \texttt{Runtime-Select} (i.e., Algorithm~\ref{alg:runtime_dp}) to compute the corresponding activation maps $\hat{\Gamma}_k$ (line~\ref{line:train_transfer_runtime} in Algorithm~\ref{alg:train_transfer}). The activation maps $\hat{\Gamma}_k$ are then used to determine which local networks $\NN_{(q, \mathcal{P})}$ need to be trained at each state $(q, s) \in \mathbb{X}^\otimes$ of the product MDP $\hat{\Sigma} \otimes \mathcal{A}_\varphi$ (line~\ref{line:train_transfer_select} in Algorithm~\ref{alg:train_transfer}). The local neural networks are trained using the method \texttt{Formal-Train} given by Algorithm~\ref{alg:safe_train} (line~\ref{line:train_transfer_train} in Algorithm~\ref{alg:train_transfer}). Compared to Algorithm~\ref{alg:offline_train} that trains all the neural networks to obtain the library $\mathfrak{NN}$, Algorithm~\ref{alg:train_transfer} reduces the number of NNs need to be trained by leveraging the training tasks $\{\mathcal{T}_1, \mathcal{T}_2, \ldots, \mathcal{T}_d\}$. 

During the test phase, we adapt previous learning experiences stored in the subset of networks $\mathfrak{NN}_\text{part}$ to test tasks $\mathcal{T}_\text{test}$ by employing transfer learning. In particular, if a local NN needed by the test task $\mathcal{T}_\text{test}$ has not been trained, we fast learn it by fine-tuning the ``closest'' NN to it in the subset $\mathfrak{NN}_\text{part}$. Thanks to the fact that each local network $\NN_{(q, \mathcal{P})}$ is associated with an abstract state $q \in \mathbb{X}$ and a controller partition $\mathcal{P} \in \mathbb{P}$, we can define the distance between two local networks $\NN_{(q_1, \mathcal{P}_1)}$ and $\NN_{(q_2, \mathcal{P}_2)}$ as follows:
\begin{align}
    \label{eq:nn_dist}
    &\text{Dist} \left(\NN_{(q_1, \mathcal{P}_1)}, \NN_{(q_2, \mathcal{P}_2)} \right) = \alpha_1 \norm{\ct_\mathbb{X}(q_1) - \ct_\mathbb{X}(q_2)}  \notag \\  
    &\hspace{30mm} + \alpha_2 \norm{\ct_\mathbb{P}(\mathcal{P}_1) - \ct_\mathbb{P}(\mathcal{P}_2)}_{\max} 
\end{align}
with pre-specified weights $\alpha_1, \alpha_2 \in \mathbb{R}^+$. Given a test task $\mathcal{T}_\text{test}$, Algorithm~\ref{alg:run_transfer} first computes the corresponding activation maps $\hat{\Gamma}_k$ (line~\ref{line:run_transfer_gamma} in Algorithm~\ref{alg:run_transfer}), and then selects local networks $\NN_{(q, \mathcal{P})}$ to be applied at each time step until reaching the product MDP's accepting set $\mathbb{X}^\otimes_G$ (line~\ref{line:run_transfer_not_reach}-\ref{line:run_transfer_select} in Algorithm~\ref{alg:run_transfer}). If the needed network $\NN_{(q, \mathcal{P})}$ has not been trained, Algorithm~\ref{alg:run_transfer} initializes the missing network $\NN_{(q, \mathcal{P})}$ using the weights of the closest network $\NN_{(q^*, \mathcal{P}^*)}$ to it in the subset $\mathfrak{NN}_\text{part}$, where the distance metric between neural networks is given by~\eqref{eq:nn_dist} (line~\ref{line:run_transfer_missing}-\ref{line:run_transfer_init_nn} in Algorithm~\ref{alg:run_transfer}). After that, the algorithm trains the missing network $\NN_{(q, \mathcal{P})}$ using PPO with only a few episodes for fine-tuning (line~\ref{line:run_transfer_ppo_update} in Algorithm~\ref{alg:run_transfer}). Thanks to the NN weight projection operator $\Pi_\mathcal{P}$, the resulting NN-based planner enjoys the same theoretical guarantees presented in Section~\ref{sec:guarantees} (line~\ref{line:run_transfer_project}-\ref{line:run_transfer_set_weights} in Algorithm~\ref{alg:run_transfer}).

\begin{algorithm}
    \caption{\textsc{Train-Transfer} $\left(\{\mathcal{T}_1, \mathcal{T}_2, \ldots, \mathcal{T}_d\}, J\right)$}
    \label{alg:train_transfer}
    \begin{algorithmic}[1]
        \STATE $\mathfrak{NN}_\text{part} = \{\}$
        \FOR{$\mathcal{T}_\text{train} \in \{\mathcal{T}_1, \mathcal{T}_2, \ldots, \mathcal{T}_d\}$}
            \STATE $\hat{\Gamma}_{k}, \hat{V}^*_0, \hat{\Sigma} \otimes \mathcal{A}_\varphi = \texttt{Runtime-Select}(\mathcal{T}_\text{train})$ \label{line:train_transfer_runtime}
            \FOR{$(q, s) \in \mathbb{X}^\otimes$, $k \in \{0, \ldots, H-1\}$}
                \STATE $(q, \mathcal{P}) = \hat{\Gamma}_k(q, s)$ \label{line:train_transfer_select}
                \IF{$\NN_{(q, \mathcal{P})} \not \in \mathfrak{NN}_\text{part}$}
                    \STATE $\NN_{(q, \mathcal{P})} = \texttt{Formal-Train}(q, \mathcal{P}, J)$ \label{line:train_transfer_train}
                    \STATE $\mathfrak{NN}_\text{part} = \mathfrak{NN}_\text{part}\; \cup\; \{\NN_{(q, \mathcal{P})}\}$
                \ENDIF
            \ENDFOR
        \ENDFOR
        \STATE \textbf{Return} $\mathfrak{NN}_\text{part}$
    \end{algorithmic}   
\end{algorithm}

\begin{algorithm}
    \caption{\textsc{Runtime-Transfer} ($\mathcal{T}_\text{test}, \mathfrak{NN}_\text{part}, J, x, s$)}
    \label{alg:run_transfer}
    \begin{algorithmic}[1]
        \STATE $\hat{\Gamma}_{k}, \hat{V}^*_0, \hat{\Sigma} \otimes \mathcal{A}_\varphi = \texttt{Runtime-Select}(\mathcal{T}_\text{test})$ \label{line:run_transfer_gamma}
        \STATE $k = 0$, $q = \abs(x)$
        \WHILE{$(q, s) \not \in \mathbb{X}_G^\otimes$} \label{line:run_transfer_not_reach}
            \STATE $(q, \mathcal{P}) = \hat{\Gamma}_k(q, s)$ \label{line:run_transfer_select}
            \IF{$\NN_{(q, \mathcal{P})} \not \in \mathfrak{NN}_\text{part}$} \label{line:run_transfer_missing}
                \STATE $\NN_{(q^*, \mathcal{P}^*)} =\hspace{-2mm} \underset{\NN_{(q_1, \mathcal{P}_1)} \in \mathfrak{NN}_\text{part}}{\text{argmin}}\hspace{-2mm} \text{Dist} \left( \NN_{(q_1, \mathcal{P}_1)}, \NN_{(q, \mathcal{P})} \right)$ \label{line:run_transfer_closest_nn}
                \STATE $\NN_{(q, \mathcal{P})} = \texttt{initialize}(\NN_{(q^*, \mathcal{P}^*)})$ \label{line:run_transfer_init_nn}
                \STATE $\NN_{(q, \mathcal{P})} = \texttt{PPO-update}(\NN_{(q, \mathcal{P})}, J)$ \label{line:run_transfer_ppo_update}
                \STATE $\widehat{W}^{(F)}, \widehat{b}^{(F)} = \Pi_{\mathcal{P}} (\NN_{(q, \mathcal{P})})$ \label{line:run_transfer_project}
                \STATE Set $\NN_{(q, \mathcal{P})}$ output layer weights be $\widehat{W}^{(F)}, \widehat{b}^{(F)}$ \label{line:run_transfer_set_weights}
                \STATE $\mathfrak{NN}_\text{part} = \mathfrak{NN}_\text{part}\; \cup\; \{\NN_{(q, \mathcal{P})}\}$
            \ENDIF    
            \STATE $u = \NN_{(q, \mathcal{P})}(x)$
            \STATE Apply action $u$, observe the new state $x$ 
            \STATE $q = \abs(x)$, $s = \delta(s, L(x))$
            \STATE $k = k+1$
        \ENDWHILE
    \end{algorithmic}  
\end{algorithm}

\subsection{Data-Driven Symbolic Model}
\label{subsec:learn_abstract}
Recall that in Algorithm~\ref{alg:runtime_dp}, after knowing the robot dynamics (i.e., the stochastic kernel $t$), the first step is to construct the symbolic model $\hat{\Sigma} = (\mathbb{X}, \mathbb{X}_0, \mathbb{P}, \hat{t})$ (line~\ref{line:mdp} in Algorithm~\ref{alg:runtime_dp}). The construction of $\hat{\Sigma}$ requires to compute the transition probabilities $\hat{t}(q^\prime | q, \mathcal{P}) = \int_{q^\prime} t(dx^\prime|z, \kappa(z))$ with all controller partitions $\mathcal{P} \in \mathbb{P}$ at each abstract state $q \in \mathbb{X}$, where $z=\ct_\mathbb{X}(q)$, $\kappa=\ct_\mathbb{P}(\mathcal{P})$. Reducing the computation of transition probabilities is tempting when the number of controller partitions is large, especially if the stochastic kernel $t(\cdot | x, u)$ is not a normal distribution and needs numerical integration. In this subsection, we accelerate the construction of $\hat{\Sigma}$ in a data-driven manner. 

For a given task $\mathcal{T}$, we consider our algorithm has access to a set of expert-provided trajectories $\mathcal{D} = \{\xi_1, \xi_2, \ldots, \xi_c\}$, such as human demonstrations that fulfill the task $\mathcal{T}$. Instead of computing all the transition probabilities $\hat{t}(q^\prime | q, \mathcal{P})$, we use the set of expert trajectories $\mathcal{D}$ to guide the computation of transitions. The resulting symbolic model can be viewed as a symbolic representation of the expert trajectories in $\mathcal{D}$. 

In Algorithm~\ref{alg:construct_mdp}, we first use imitation learning to train a neural network $\NN$ by imitating the expert trajectories in $\mathcal{D}$ (line~\ref{line:construct_mdp_imitation} in Algorithm~\ref{alg:construct_mdp}). Though the neural network $\NN$ trained using a limited dataset $\mathcal{D}$ may not always fulfill the task $\mathcal{T}$, the network $\NN$ contains relevant control actions that can be used to obtain the final controller. In particular, at each abstract state $q \in \mathbb{X}$, we only compute transition probabilities $\hat{t}(q^\prime | q, \mathcal{P})$ with controller partitions $\mathcal{P}$ suggested by the network $\NN$. To be specific, let $u^*$ be the control actions given by the network $\NN$ at the centers of abstract states $q \in \mathbb{X}$ (line~\ref{line:construct_mdp_u} in Algorithm~\ref{alg:construct_mdp}). Then, Algorithm~\ref{alg:construct_mdp} selects a subset $P_q \subseteq \mathbb{P}$ consists of $I$ controller partitions that yield control actions close to the NN's output $u^*$, where $I \in \mathbb{N}$ is a user-defined parameter (line~\ref{line:construct_mdp_start}-\ref{line:construct_mdp_end} in Algorithm~\ref{alg:construct_mdp}). Finally, Algorithm~\ref{alg:construct_mdp} computes a symbolic model $\hat{\Sigma}$ with only transitions under the controller partitions in the subset $P_q$ (line~\ref{line:construct_mdp_probability} in Algorithm~\ref{alg:construct_mdp}). The symbolic model $\hat{\Sigma}$ contains more transitions by increasing the parameter $I$ at the cost of computational efficiency. The choice of $I$ can be adaptively determined as discussed in the next subsection. 

\begin{algorithm}
    \caption{\textsc{Construct-Symbol-Model} ($\mathcal{T}, \mathcal{D}, \mathbb{X}, \mathbb{P}, I$)}
    \label{alg:construct_mdp}
    \begin{algorithmic}[1]
        \STATE $\NN = \texttt{imitation-learning}(\mathcal{D})$ \label{line:construct_mdp_imitation}
        \FOR{$q \in \mathbb{X}$}
            \STATE $u^* = \NN(z)$, where $z = \ct_\mathbb{X}(q)$ \label{line:construct_mdp_u}
            \STATE $P_q = \{\}$ \label{line:construct_mdp_start}
            \FOR{$i = 1, \ldots, I$}
                \STATE $\mathcal{P}^*\! = \underset{\mathcal{P} \in \mathbb{P} \setminus P_q  }{\text{argmin}} \norm{\kappa(z) - u^*}$, s.t. $\kappa=\ct_\mathbb{P}(\mathcal{P}), z=\ct_\mathbb{X}(q)$
                \STATE $P_q = P_q \cup \{\mathcal{P}^*\}$
            \ENDFOR \label{line:construct_mdp_end}
            \STATE Compute $\hat{t}(q^\prime | q, \mathcal{P})$ with $\mathcal{P} \in P_q$ \label{line:construct_mdp_probability}
        \ENDFOR
        \STATE \textbf{Return} $\hat{\Sigma}$
    \end{algorithmic}   
\end{algorithm}

\subsection{Adaptive Partitioning}
\label{subsec:adaptive_partitioning}

Recall that during the offline training, we partition the state space $X \subset \mathbb{R}^n$ and the controller space $\mathcal{P}^{K \times b} \subset \mathbb{R}^{m \times (n+1)}$ using the pre-specified parameters $\eta_q$ and $\eta_\mathcal{P}$, respectively (see Section~\ref{subsec:offline}). In this subsection, we comment on the choice of the grid sizes $\eta_q$ and $\eta_\mathcal{P}$. In particular, our framework can directly incorporate the discretization techniques from the literature of abstraction-based controller synthesis (e.g.~\cite{soudjani2013siam, hsu2018hscc}). To that end, we provide a simple yet efficient example of adaptive partitioning in Algorithm~\ref{alg:runtime_adapt}, which enables the update of gird sizes $\eta_q$ and $\eta_\mathcal{P}$ at runtime using transfer learning. 

The first part of Algorithm~\ref{alg:runtime_adapt} aims to partition the state and controller spaces such that the resulting probabilities $\hat{V}^*_0(q, s)$ of satisfying the specification $\varphi$ are greater than the pre-specified threshold $p$ at all initial states $(q, s) \in \mathbb{X}_0 \times S_0$ (line~\ref{line:runtime_adapt_partition_start}-\ref{line:runtime_adapt_partition_end} in Algorithm~\ref{alg:runtime_adapt}). In particular, if the probability $\hat{V}^*_0(q, s)$ is less than $p$ at some state $(q, s) \in \mathbb{X}_0 \times S_0$, Algorithm~\ref{alg:runtime_adapt} decreases the current grid sizes $\eta_q$ and $\eta_\mathcal{P}$ by half and increases the parameter $I$ (line~\ref{line:runtime_adapt_grid} in Algorithm~\ref{alg:runtime_adapt}).
After having such a partitioning of the state and controller spaces, Algorithm~\ref{alg:runtime_adapt} trains the corresponding locals networks by fine-tuning the NNs in the provided library of networks $\mathfrak{NN}_\text{part}$ (line~\ref{line:runtime_adapt_train_start}-\ref{line:runtime_adapt_train_end} in Algorithm~\ref{alg:runtime_adapt}). The following theoretical guarantee for the resulting NN-based planner to satisfy the given specification $\varphi$ directly follows Theorem~\ref{thm:nn_v}.
\begin{corollary}
    Consider Algorithm~\ref{alg:runtime_adapt} returns a library of local networks $\mathfrak{NN}_\text{part}$ and an activation map $\Gamma$ (denoting the functions $\hat{\Gamma}_k$). Then, the NN-based planner $\NN_{[\mathfrak{NN}_\text{part},\Gamma]}$ satisfying $\pr \left(\xi_{\NN_{[\mathfrak{NN}_\text{part}, \Gamma]}}^{(x, s)} \models \varphi \right) \geq p - \varepsilon$ for any $x \in X_0$ and $s \in S_0$, where $\varepsilon = HZ\Delta^\NN$ and $\Delta^\NN$ is given by~\eqref{eq:delta_nn}.
\end{corollary}

\begin{algorithm}
    \caption{\textsc{Adapt-Partition} ($\mathcal{T}, \mathcal{D}, \mathfrak{NN}_\text{part}, J, \eta_q, \eta_\mathcal{P}, I$)}
    \label{alg:runtime_adapt}
    \begin{algorithmic}[1]
        \WHILE{$\hat{V}^*_{\min} < p$} \label{line:runtime_adapt_partition_start}
            \STATE $\mathbb{X}\hspace{-0.5mm} =\hspace{-0.5mm} \texttt{partition} (X, \eta_q)$,\hspace{-1mm} $\mathbb{P}\hspace{-0.5mm} =\hspace{-0.5mm} \texttt{partition} (\mathcal{P}^{K \times b}, \eta_\mathcal{P})$
            \STATE $\hat{\Sigma} = \texttt{Construct-Symbol-Model} (\mathcal{T}, \mathcal{D}, \mathbb{X}, \mathbb{P}, I)$
            \STATE $\hat{\Gamma}_{k}, \hat{V}^*_0, \hat{\Sigma} \otimes \mathcal{A}_\varphi = \texttt{Runtime-Select}(\mathcal{T})$
            \STATE $\hat{V}^*_{\min} = \underset{(q, s) \in \mathbb{X}_0 \times S_0}{\min} \hat{V}^*_0(q, s)$
            \STATE $\eta_q = \eta_q / 2,\ \eta_\mathcal{P}= \eta_\mathcal{P} / 2,\ I = 2 I$ \label{line:runtime_adapt_grid}
        \ENDWHILE \label{line:runtime_adapt_partition_end}
        \FOR{$(q, s) \in \mathbb{X}^\otimes$, $k \in \{0, \ldots, H-1\}$} \label{line:runtime_adapt_train_start}
            \STATE $(q, \mathcal{P}) = \hat{\Gamma}_k(q, s)$ 
            \IF{$\NN_{(q, \mathcal{P})} \not \in \mathfrak{NN}_\text{part}$} 
                \STATE $\NN_{(q^*, \mathcal{P}^*)} =\hspace{-2mm} \underset{\NN_{(q_1, \mathcal{P}_1)} \in \mathfrak{NN}_\text{part}}{\text{argmin}}\hspace{-2mm} \text{Dist} \left( \NN_{(q_1, \mathcal{P}_1)}, \NN_{(q, \mathcal{P})} \right)$ 
                \STATE $\NN_{(q, \mathcal{P})} = \texttt{initialize}(\NN_{(q^*, \mathcal{P}^*)})$ 
                \STATE $\NN_{(q, \mathcal{P})} = \texttt{PPO-update}(\NN_{(q, \mathcal{P})}, J)$ 
                \STATE $\widehat{W}^{(F)}, \widehat{b}^{(F)} = \Pi_{\mathcal{P}} (\NN_{(q, \mathcal{P})})$ 
                \STATE Set $\NN_{(q, \mathcal{P})}$ output layer weights be $\widehat{W}^{(F)}, \widehat{b}^{(F)}$
                \STATE $\mathfrak{NN}_\text{part} = \mathfrak{NN}_\text{part}\; \cup\; \{\NN_{(q, \mathcal{P})}\}$
            \ENDIF   
        \ENDFOR \label{line:runtime_adapt_train_end}
        \STATE \textbf{Return} $\mathfrak{NN}_\text{part}, \{\hat{\Gamma}_{k}\}_{k \in \{0, \ldots, H-1\}}$
    \end{algorithmic}  
\end{algorithm}

\section{Results}
\label{sec:results}
We evaluated the proposed framework both in simulation and on a robotic vehicle. All experiments were executed on a single Intel Core i9 2.4-GHz processor with 32 GB of memory. Our open-source implementation of the proposed neurosymbolic framework can be found at {\small \texttt{https://github.com/rcpsl/Neurosymbolic\_planning}}.

\subsection{Controller Performance in Simulation}
Consider a wheeled robot with the state vector $x = [\zeta_x, \zeta_y, \theta]^\top \in X \subset \mathbb{R}^3$, where $\zeta_x$, $\zeta_y$ denote the coordinates of the robot and $\theta$ is the heading direction. The priori known nominal model $f$ in the form of~\eqref{eq:dyn} is given by:
\begin{align}
    \zeta_x^{(t+\Delta t)} &= \zeta_x^{(t)} + \Delta t\ v\ \text{cos}(\theta^{(t)}) \notag \\
    \zeta_y^{(t+\Delta t)} &= \zeta_y^{(t)} + \Delta t\ v\ \text{sin}(\theta^{(t)}) \label{eq:dubin_car}\\
    \theta^{(t+\Delta t)} &= \theta^{(t)} + \Delta t\ u^{(t)} \notag
\end{align}
where the speed $v=0.3$m/s and the time step $\Delta t=1$s. 
We train NNs to control the robot, i.e., $u^{(t)} = \text{NN}(x^{(t)}), \; \text{NN} \in \mathcal{P}^{K \times b} \subset \mathbb{R}^{1 \times 4}$ with the controller space $\mathcal{P}^{K \times b}$ being a hyperrectangle.

\begin{figure}[!h]
    \center
    \resizebox{.49\textwidth}{!}{
    \begin{tabular}{c|c}
        \rotatebox{90}{$\quad$\Large{\textbf{Our Neurosymbolic Framework}}}
         & 
        \includegraphics[height=0.4\textwidth]{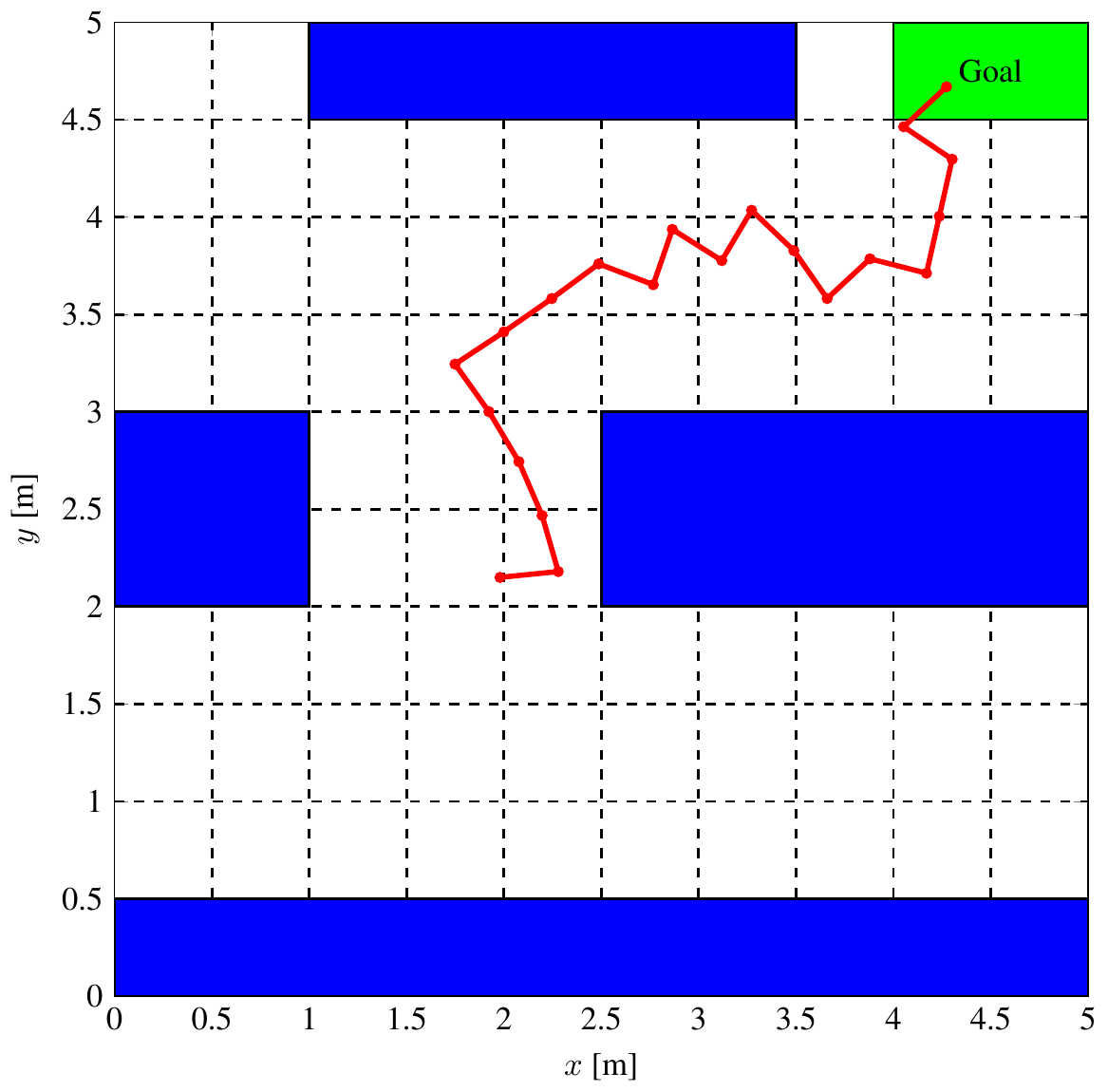}
        \includegraphics[height=0.4\textwidth]{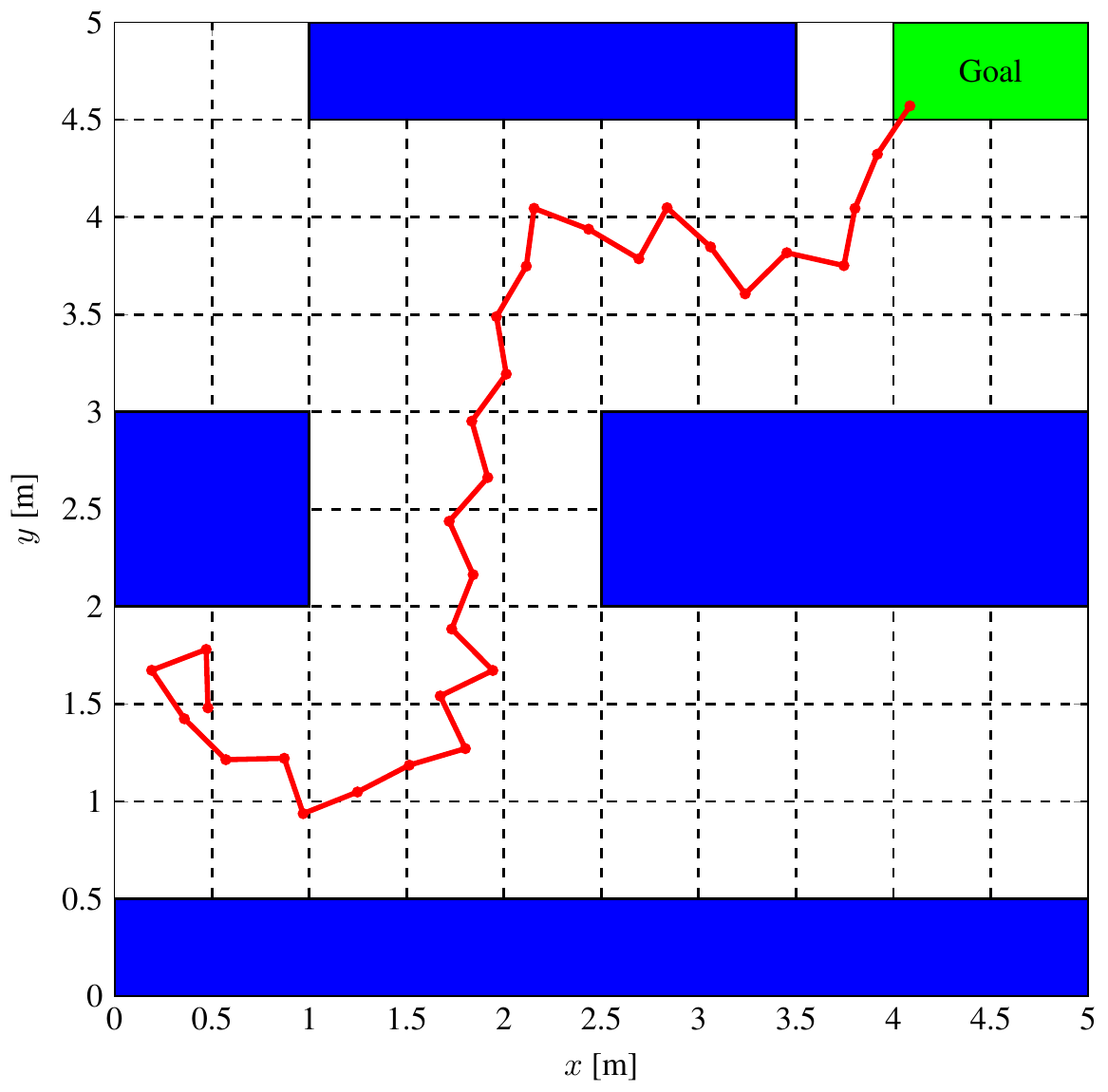}
        \includegraphics[height=0.4\textwidth]{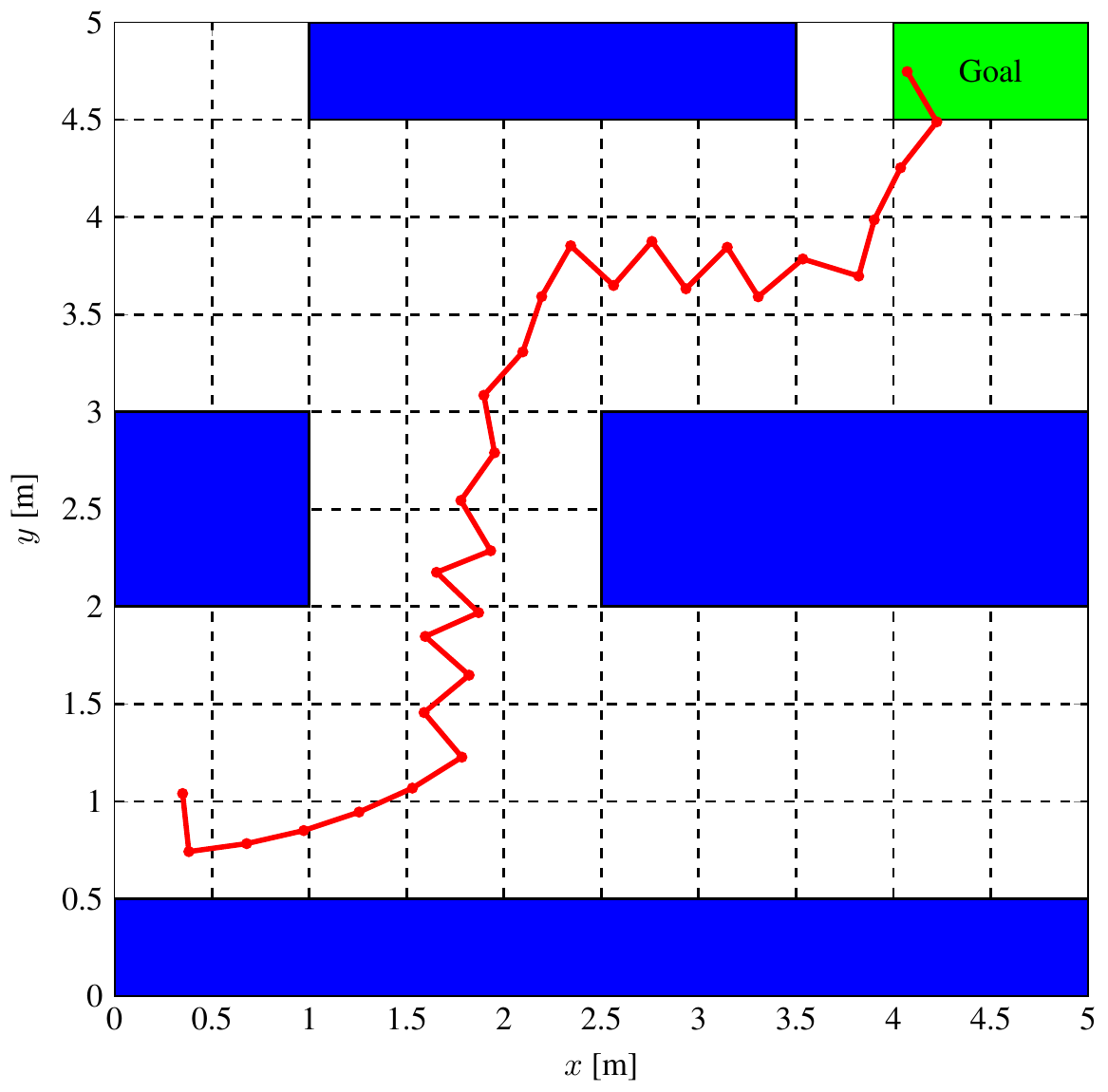}
        \\ \hline
        \rotatebox{90}{$\quad$\Large{\textbf{Standard Imitation Learning}}}
         & 
        \includegraphics[height=0.4\textwidth]{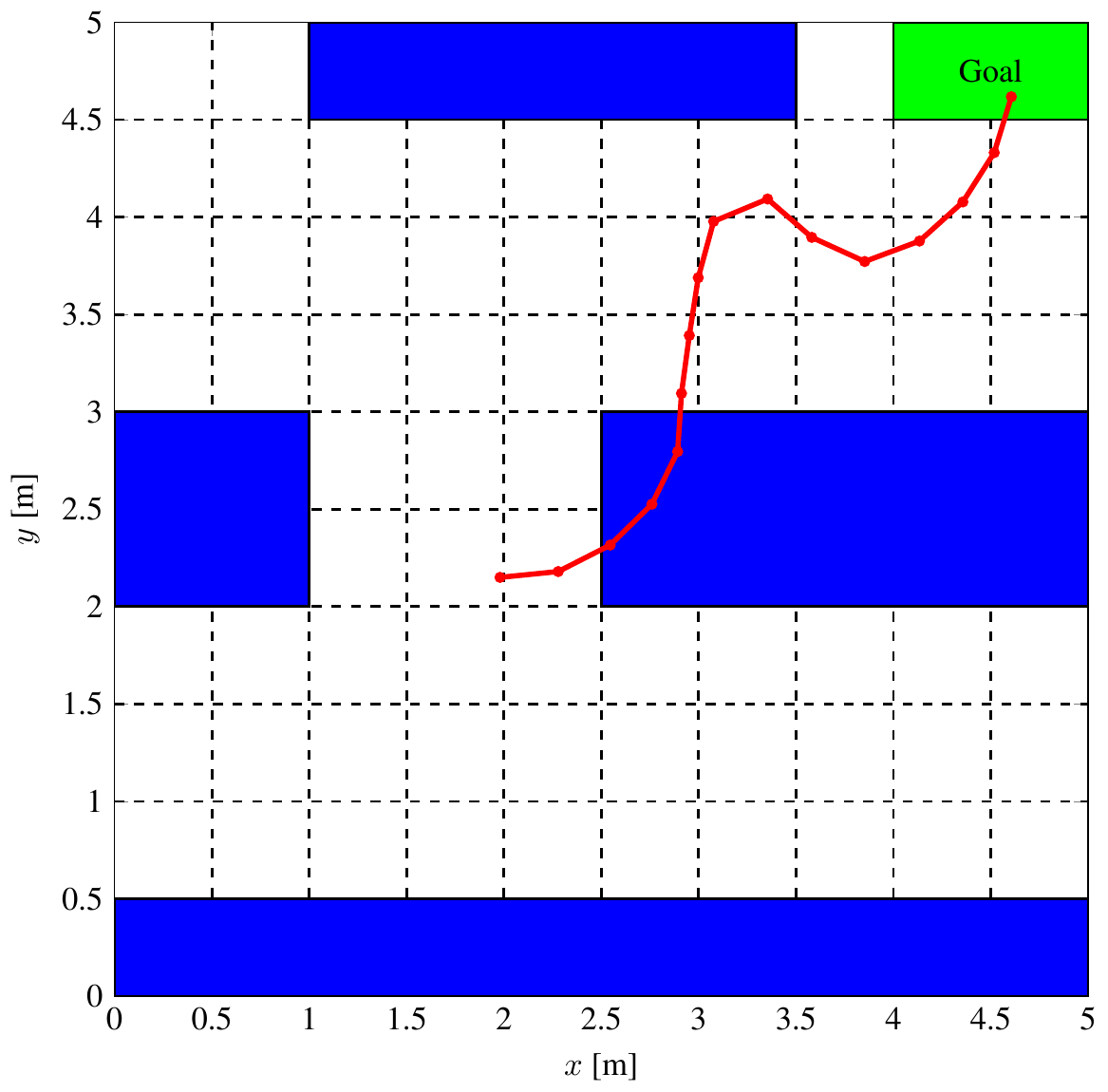}
        \includegraphics[height=0.4\textwidth]{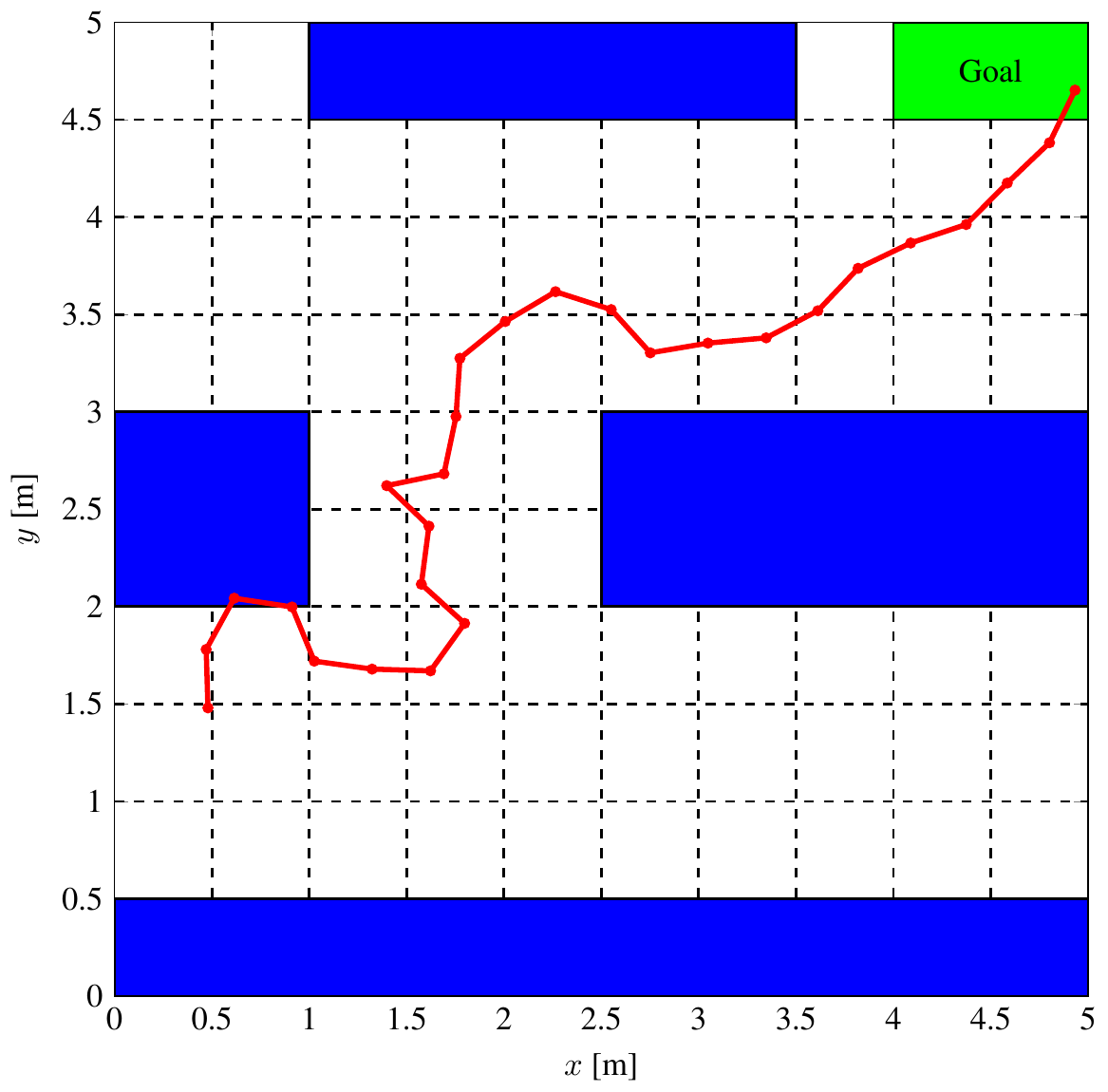} 
        \includegraphics[height=0.4\textwidth]{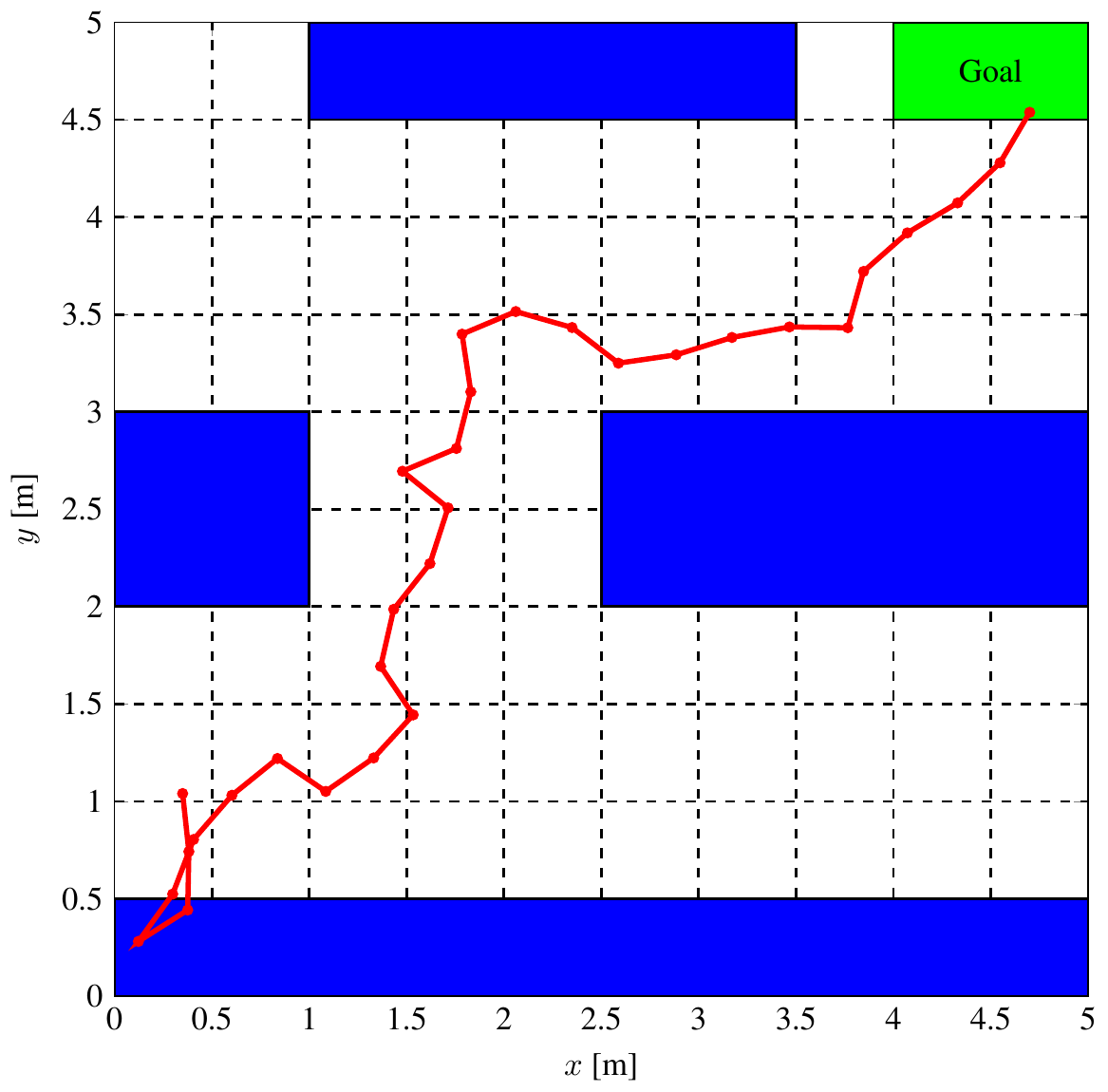}
    \end{tabular}  
    }  
    \caption{The upper row shows trajectories resulting from NN-based planners trained using our framework. The lower row shows trajectories under the control of NNs trained by standard imitation learning, where the NN architectures are (left) $2$ hidden layers with $10$ neurons per layer, (middle) $2$ hidden layers with $64$ neurons per layer, and (right) $3$ hidden layers with $128$ neurons per layer. With the same initial states (two subfigures in the same column), only NN-based planners trained by our framework lead to collision-free trajectories.} 
    \label{fig:compare_imitation}  
\end{figure}

As the first step of our framework, we discretized the state space $X \subset \mathbb{R}^3$ and the controller space $\mathcal{P}^{K \times b} \subset \mathbb{R}^{1 \times 4}$ as described in Section~\ref{subsec:offline}. Specifically, we partitioned the range of heading direction $\theta \in [0, 2\pi)$ uniformly into $8$ intervals, and the partitions in the $x$, $y$ dimensions are shown as the dashed lines in Figure~\ref{fig:compare_imitation}. We uniformly partitioned the controller space $\mathcal{P}^{K \times b}$ into $240$ hyperrectangles.

\noindent\textbf{Study\#1: Comparison against standard NN training for a fixed task.}
The objective of this study is to compare the proposed framework against standard NN training when the task is known during training time. We aim to show the ability of our framework to guarantee the safety and correctness of achieving the task compared with standard NN training. To that end, we considered the workspace shown in Figure~\ref{fig:compare_imitation} and a simple reach-avoid specification, i.e., reach the goal area (green) while avoiding the obstacles (blue).

We collected data by observing the control actions of an expert controller operating in this workspace while varying the initial position of the robot. We trained several NNs using imitation learning for a wide range of NN architectures and a number of episodes to achieve the best performance. 


We then trained a library of neural networks $\mathfrak{NN}$ using Algorithm~\ref{alg:offline_train}, and we used the dataset---used to train NNs with imitation learning---to accelerate the runtime selection as detailed in Algorithm~\ref{alg:construct_mdp} (recall that line 1 in Algorithm~\ref{alg:construct_mdp} uses imitation-learning). 

We report the trajectories of the proposed neurosymbolic framework in the first row of  Figure~\ref{fig:compare_imitation} and the results of the top performing NNs obtained from imitation learning in the second row of Figure~\ref{fig:compare_imitation}. As shown in the figure, we were able to find initial states from which the imitation-learning-based NNs failed to guarantee the safety of the robot (and hence failed to satisfy the mission goals). However, 
as shown in the figure (and supported by our theoretical analysis in Theorem~\ref{thm:nn_optimal}), our framework was capable of always achieving the mission goals and steering the robot safely to the goal.


\noindent\textbf{Study\#2: Generalization to unknown workspace/tasks using transfer learning.} This experiment aims to study our framework's ability to generalize to unseen tasks even when the library of neural networks is not complete. In other words, the trained local networks in $\mathcal{NN}$ cannot cover all possible transitions in the symbolic model, and hence a transfer learning needs to be performed during the runtime selection phase.

\begin{figure}[!h]
    \center
    \resizebox{.49\textwidth}{!}{
    \begin{tabular}{c}
        \includegraphics[height=0.5\textwidth]{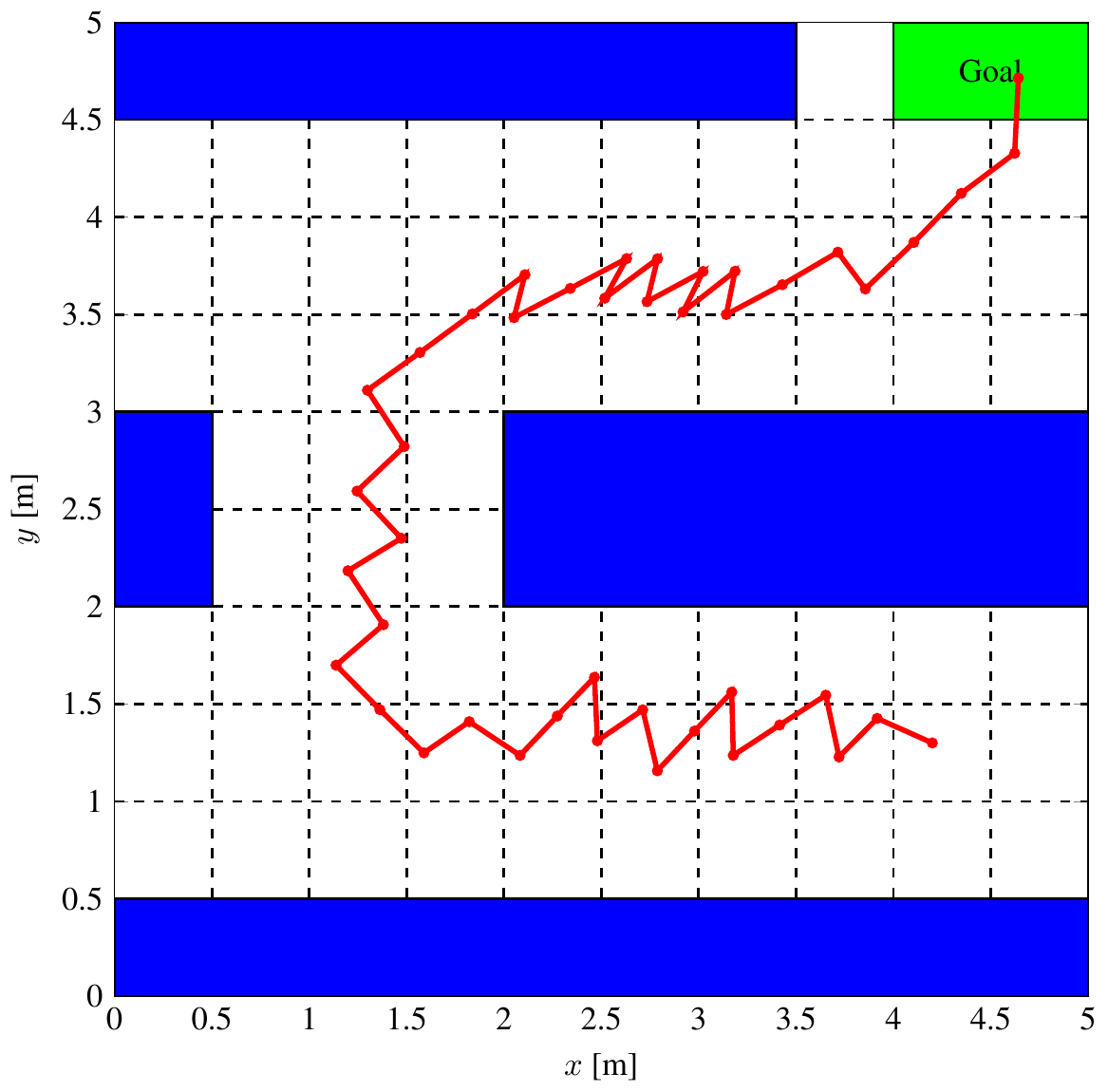}
        \includegraphics[height=0.5\textwidth]{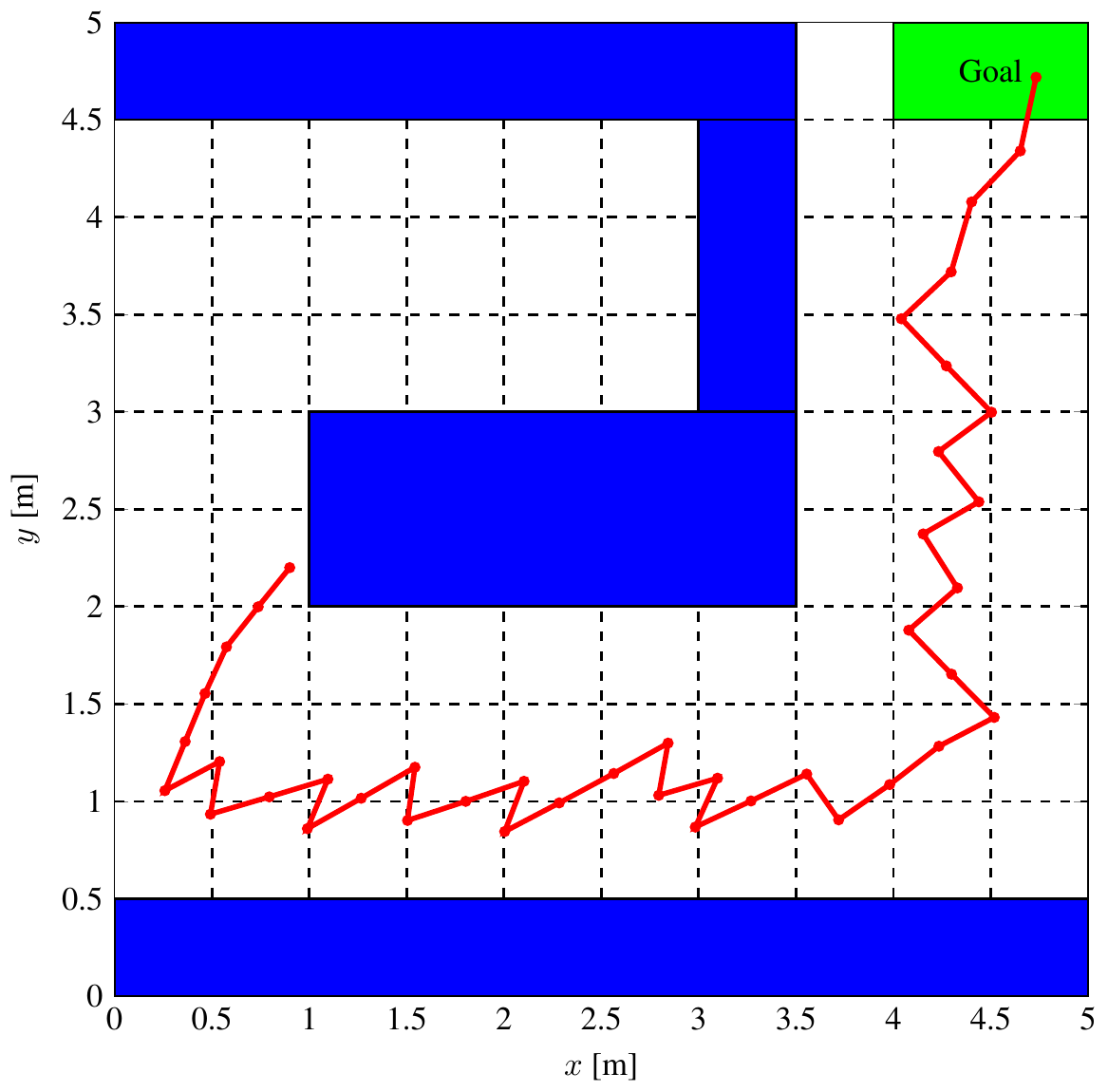}
        \includegraphics[height=0.5\textwidth]{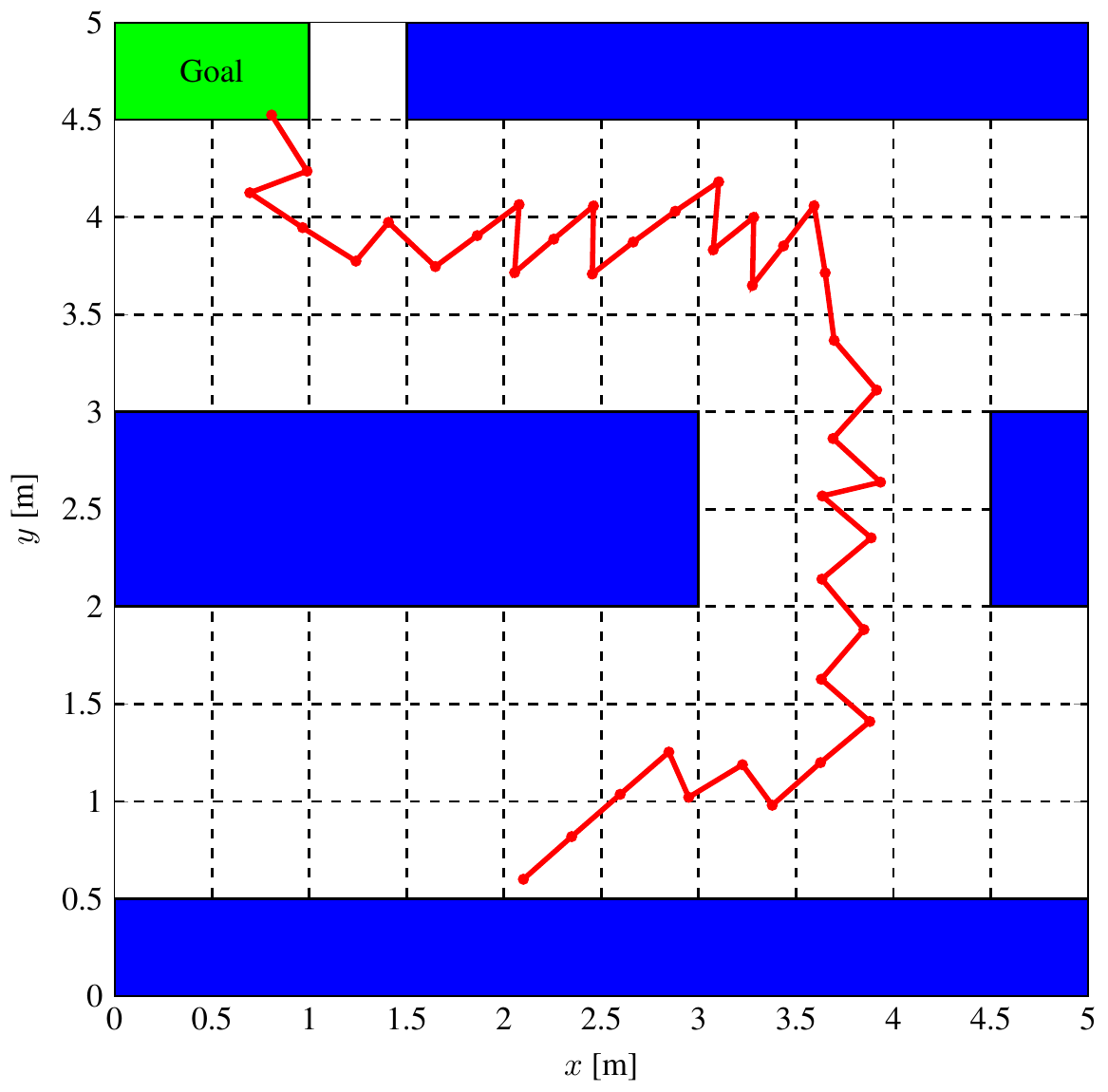} 
        \\ 
        \includegraphics[height=0.5\textwidth]{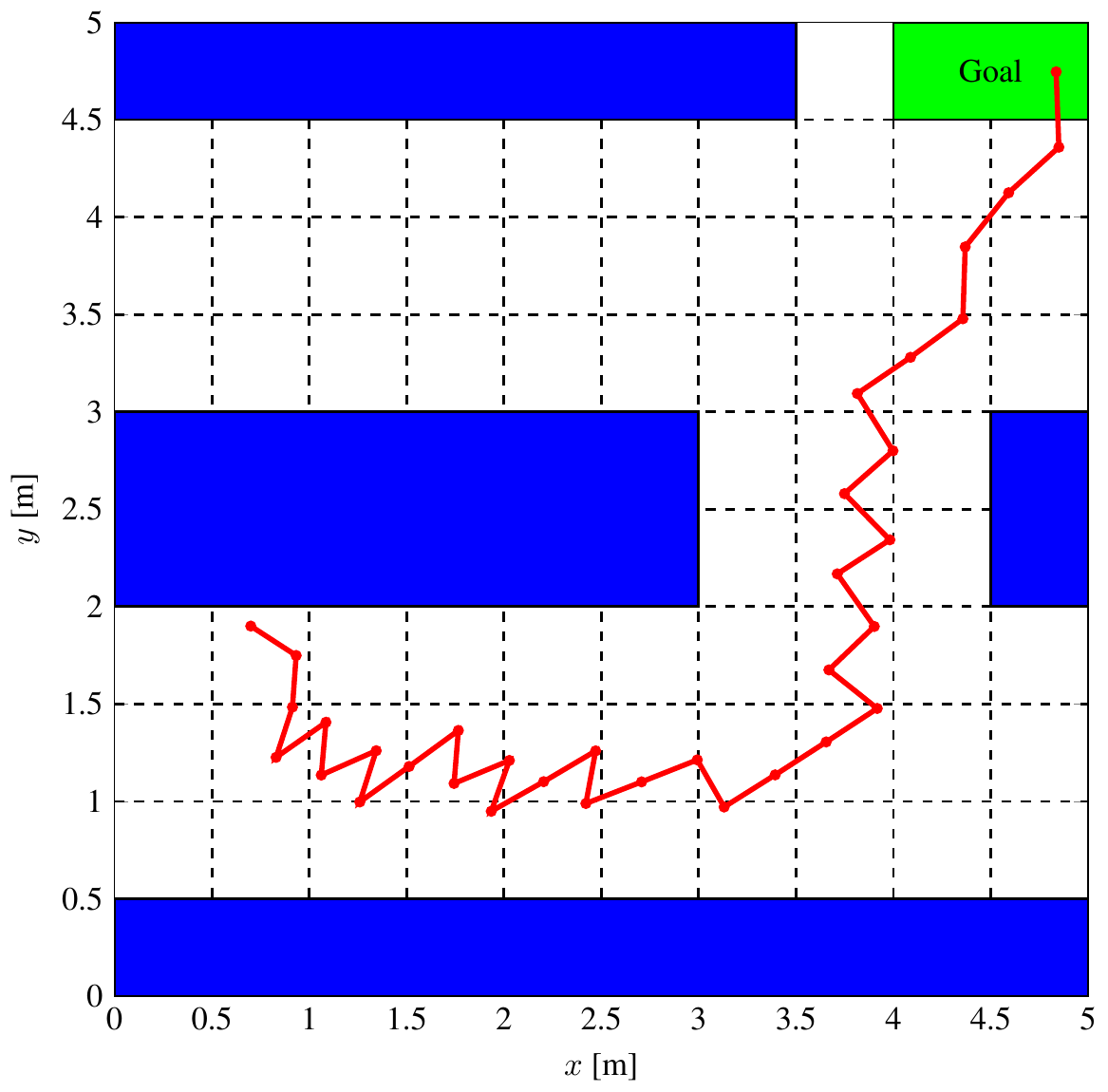}
        \includegraphics[height=0.5\textwidth]{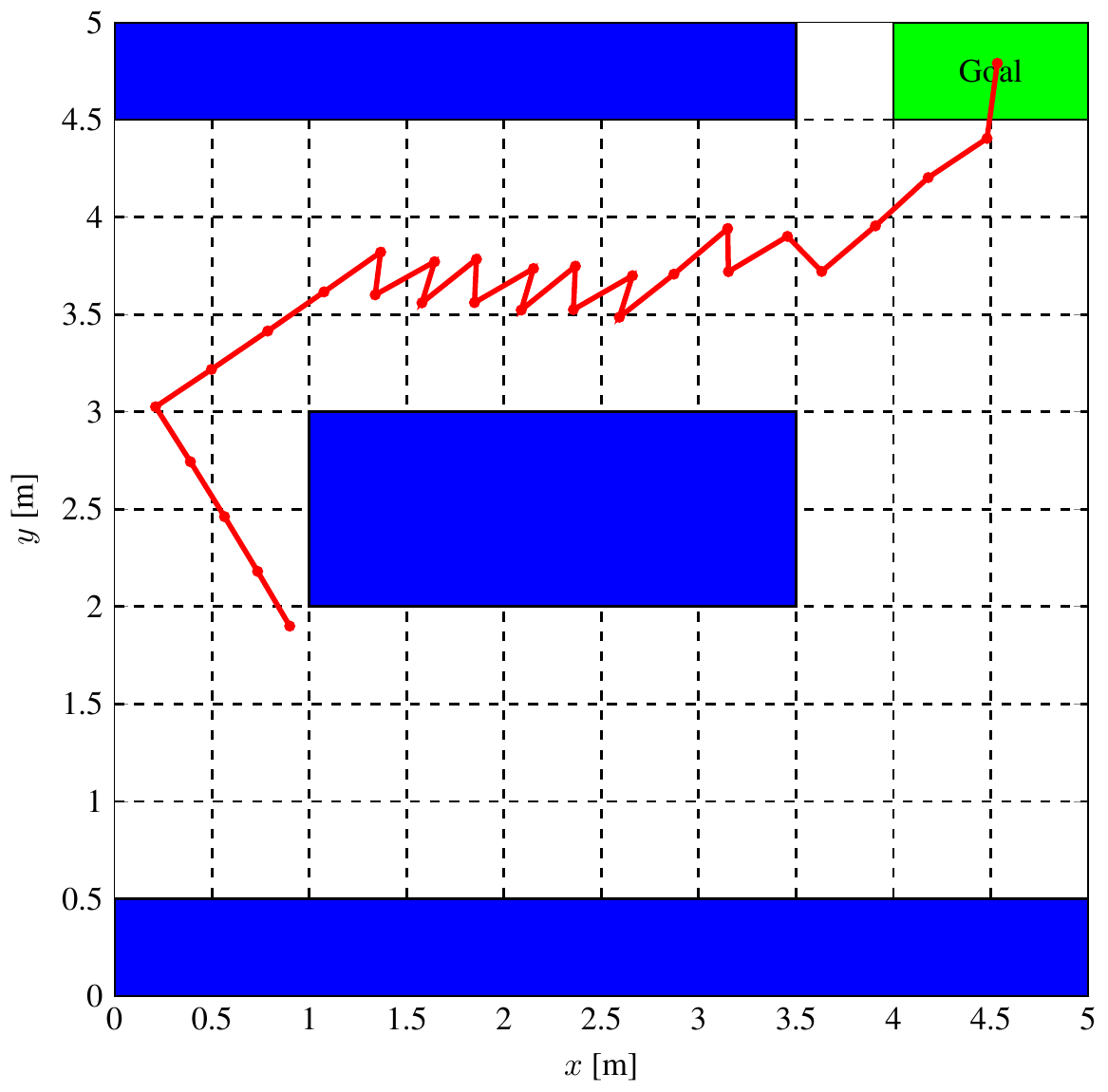}
        \includegraphics[height=0.5\textwidth]{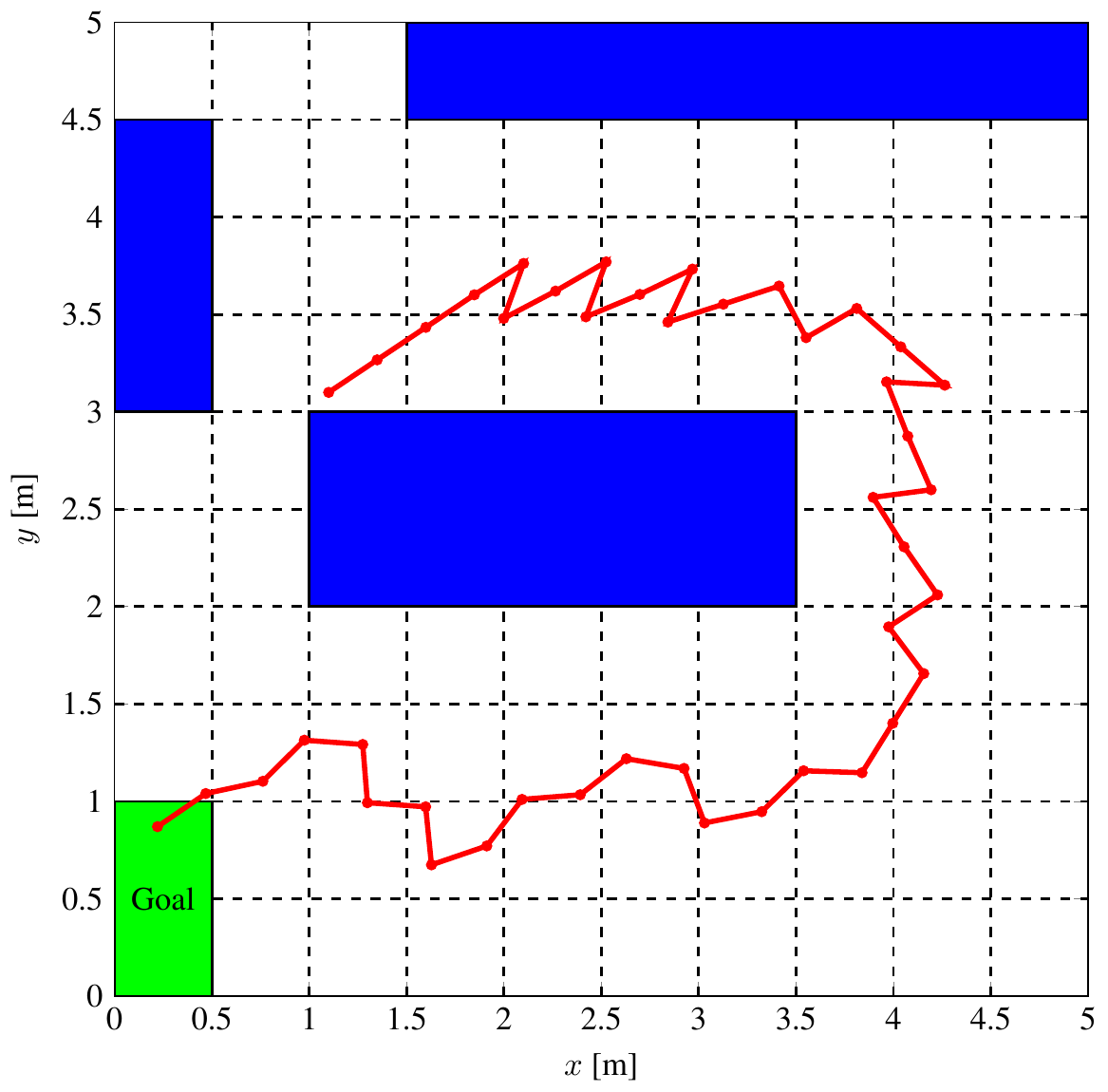}
    \end{tabular}
    }
    \caption{The upper row shows trajectories in workspaces $\mathcal{W}_1$, $\mathcal{W}_3$, $\mathcal{W}_5$, and the lower row corresponds to workspaces $\mathcal{W}_2$, $\mathcal{W}_4$, $\mathcal{W}_6$. The subset of local networks $\mathfrak{NN}_\text{part}$ is trained in workspace $\mathcal{W}_1$ and the rest five workspaces are given at runtime. Trajectories in all the workspaces satisfy both the safety specification $\varphi_\text{safety}$ (blue areas are obstacles) and the liveness specification $\varphi_\text{liveness}$ for reaching the goal (green area).}%
    \label{fig:simulation}  
\end{figure}

During the offline training, we trained a subset of local networks $\mathfrak{NN}_\text{part}$ by following Algorithm~\ref{alg:train_transfer} in Section~\ref{subsec:transfer}. Specifically, the local NNs are trained in the workspace $\mathcal{W}_1$ (the first subfigure in the upper row of Figure~\ref{fig:simulation}). The set $\mathfrak{NN}_\text{part}$ consists of $658$ local NNs, where each local NN has only one hidden layer with $6$ neurons. We used Proximal Policy Optimization (PPO) implemented in Keras~\cite{chollet2015keras} to train each local NN for $800$ episodes, and projected the NN weights at the end of training. The total time for training and projecting weights of the $658$ local networks in $\mathfrak{NN}_\text{part}$ is $2368$ seconds.

At runtime, we tested the trained NN-based planner in five unseen workspaces $\mathcal{W}_i$, $i=2, \ldots, 6$, and the corresponding trajectories are shown in Figure~\ref{fig:simulation}. For each of the workspaces, our framework computes an activation map $\Gamma$ that assigns a controller partition $\mathcal{P} \in \mathbb{P}$ to each abstract state $q \in \mathbb{X}$ through dynamical programming (Algorithm~\ref{alg:runtime_dp} in Section~\ref{subsec:runtime}). The local NNs corresponding to the assigned controller partitions may not have been trained offline. If this was the case, we  follow Algorithm~\ref{alg:run_transfer} that employs transfer learning to learn the missing NNs at runtime efficiently. Specifically, after initializing a missing NN using its closest NN in the set $\mathfrak{NN}_\text{part}$, we trained it for $80$ episodes, which is much less than the number of episodes used in the offline training. For example, for the workspace $\mathcal{W}_2$ (the first subfigure in the lower row of Figure~\ref{fig:simulation}), the length of the corresponding trajectory is $35$ steps, and $28$ local NNs used along the trajectory are not in the set $\mathfrak{NN}_\text{part}$. Our algorithm efficiently trains these $28$ local NNs in $10.5$ seconds, which shows the capability of our framework in real-time applications.

\subsection{Actual Robotic Vehicle}   
\begin{figure}[!ht]  
    \center
    \resizebox{.45\textwidth}{!}{
        \raisebox{1ex}{\includegraphics[width=0.15\linewidth, height=0.18\linewidth]{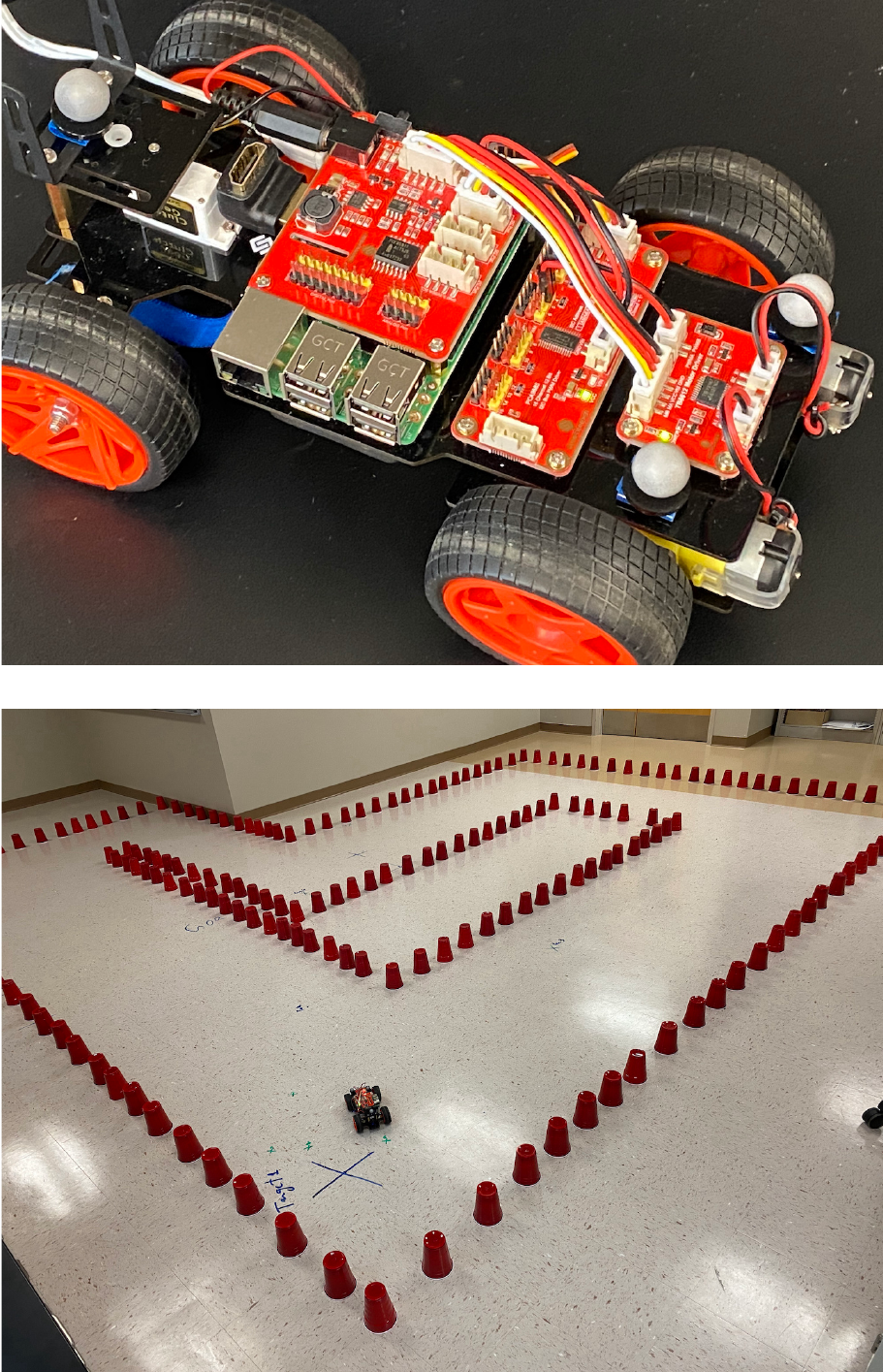}}
       \includegraphics[width=0.19\linewidth, height=0.2\linewidth]{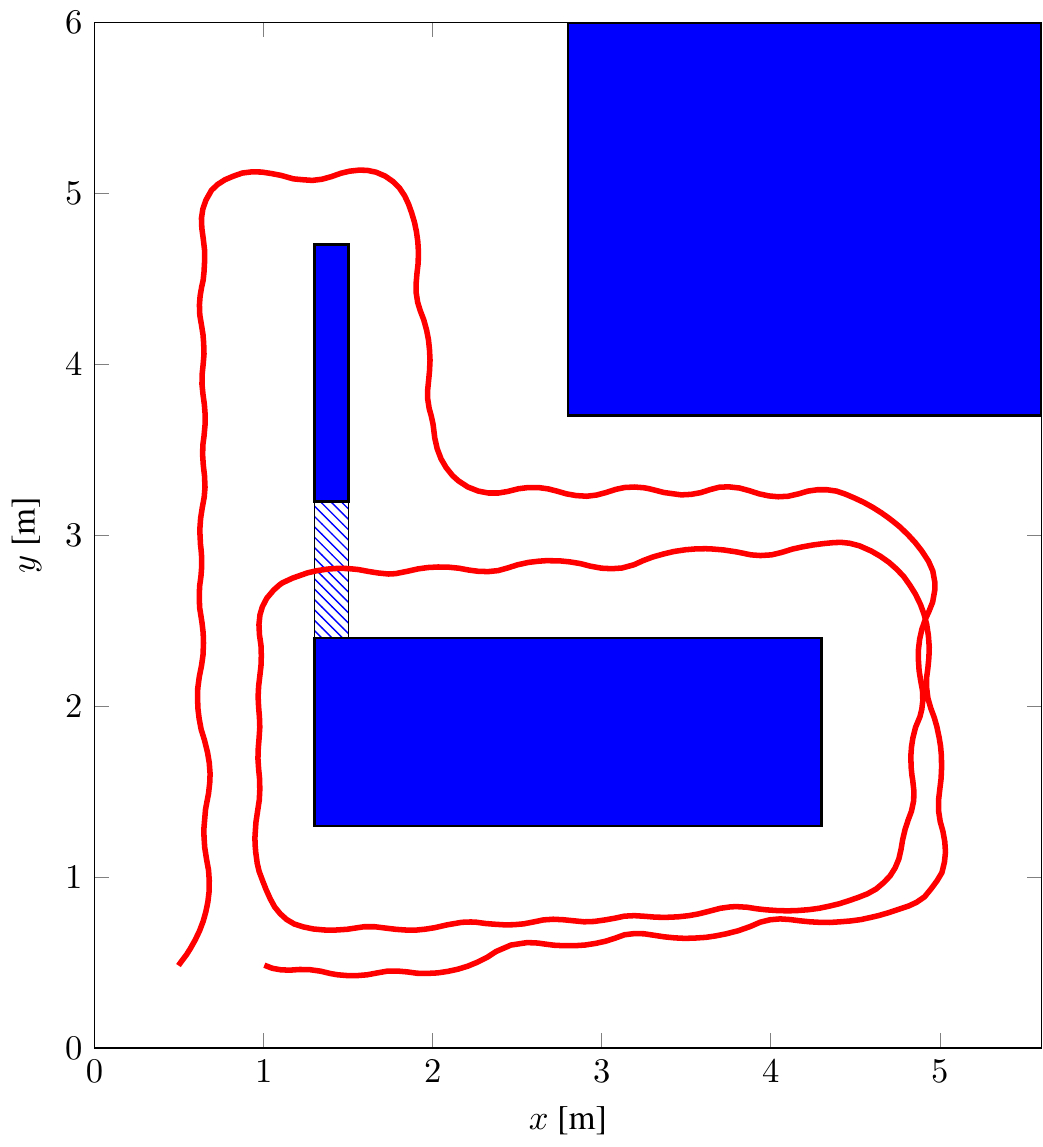}
    }
    \caption{(Left) PiCar and workspace. (Right) The PiCar's trajectory (red) for two loops, where the striped blue obstacle is removed after the first loop.} 
    \label{fig:loop} 
\end{figure}

We tested the proposed framework on a small robotic vehicle called PiCar, which carries a Raspberry Pi that runs the NNs trained by our framework. We used a Vicon motion capture system to measure the states of the PiCar in real-time. Figure~\ref{fig:loop} (left) shows the PiCar and our experimental setup. We modeled the PiCar's dynamics using the rear-wheel bicycle drive~\cite{rearwheel} and used GP regression to learn the model-error.

\noindent\textbf{Study\#3: Dynamic changes in the workspace.}
We study the ability of our framework to adapt, at runtime, to changes in the workspace. This is critical in cases when the workspace is dynamic and changes over time. To that end, we trained NNs in the workspace shown in Figure~\ref{fig:loop} (right). The part of the obstacle colored in striped blue was considered an obstacle during the training, but was removed at runtime after the PiCar finished running the first loop. Thanks to the DP recursion that selects the optimal NNs at runtime 
(Algorithm~\ref{alg:runtime_dp} in Section~\ref{subsec:runtime}), the PiCar was capable of updating its optimal selection of neural networks and found a better trajectory to achieve the mission.

\begin{figure}[!ht]
    \center
    \resizebox{.49\textwidth}{!}{
    \begin{tabular}{c|c}
        \rotatebox{90}{$\qquad$\Large{\textbf{Workspaces Used for Training NNs}}}
        & 
        \includegraphics[height=0.5\textwidth]{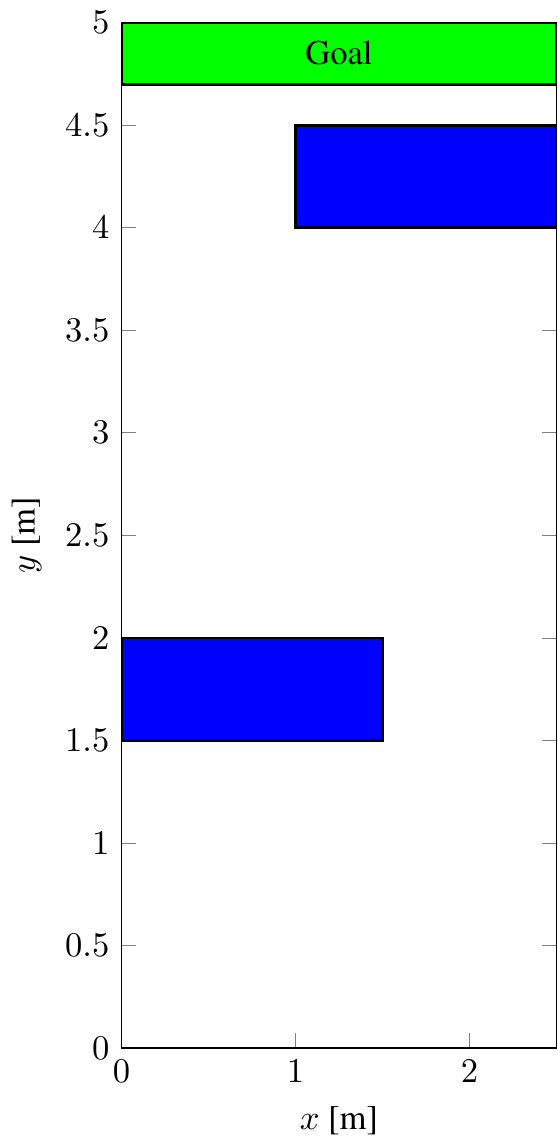}
        \includegraphics[height=0.5\textwidth]{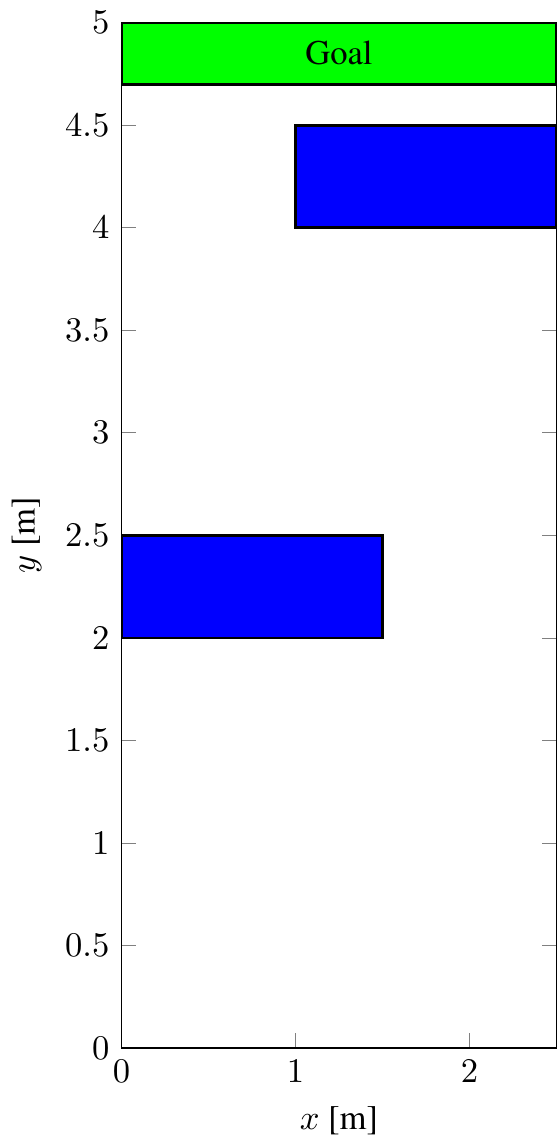} 
        \includegraphics[height=0.5\textwidth]{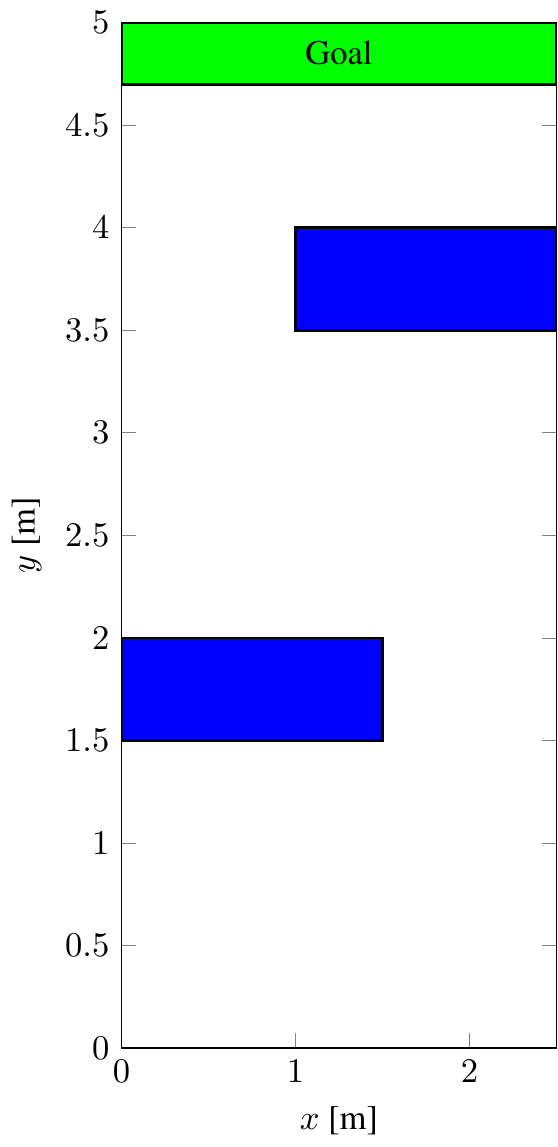}
        \includegraphics[height=0.5\textwidth]{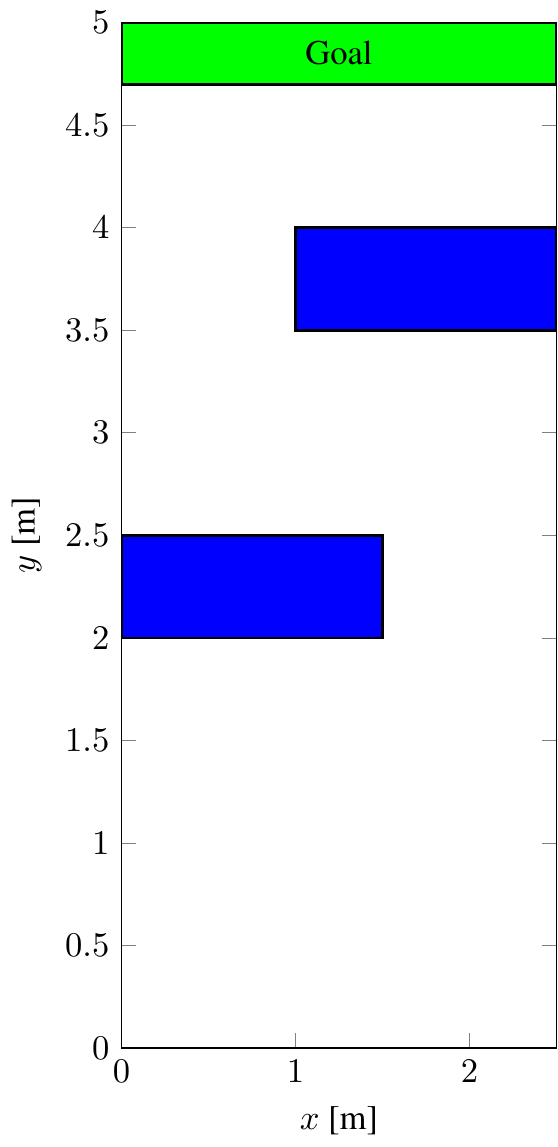}
        \\ \hline
        \rotatebox{90}{$\qquad$\Large{\textbf{Our Neurosymbolic Framework}}}
        & 
        \includegraphics[height=0.5\textwidth]{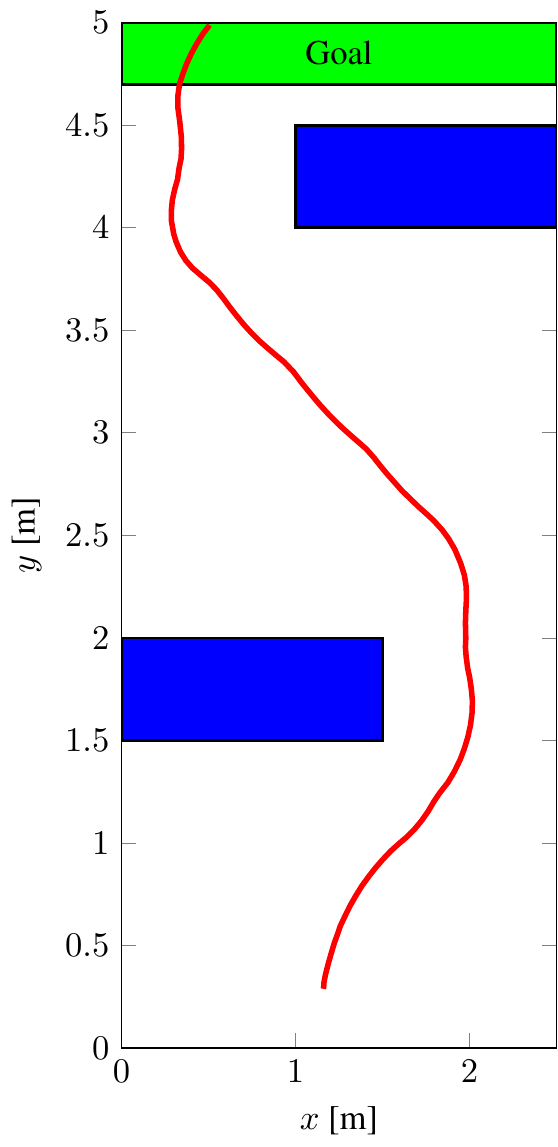}
        \includegraphics[height=0.5\textwidth]{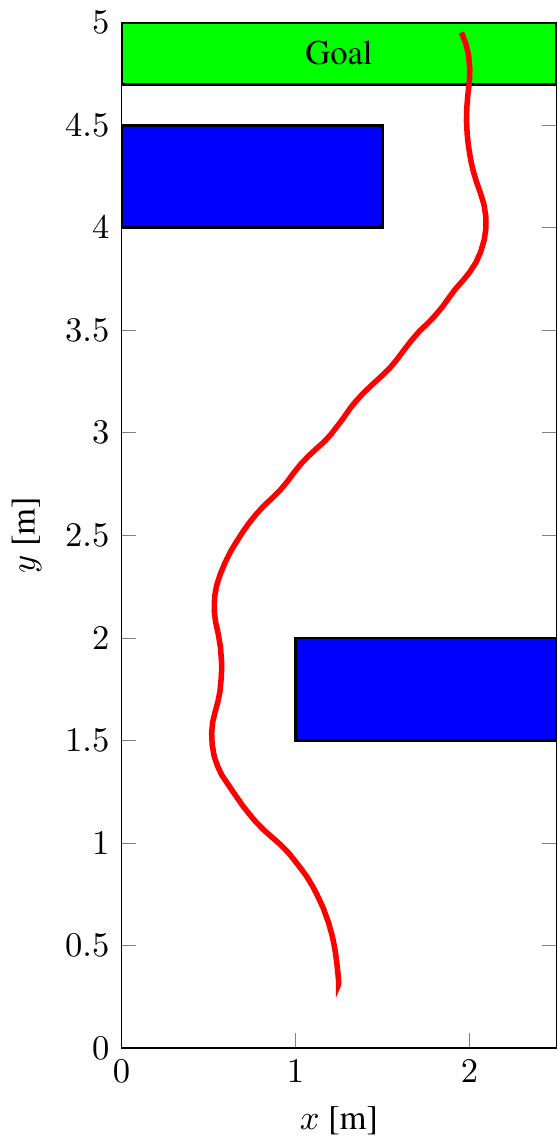}
        \includegraphics[height=0.5\textwidth]{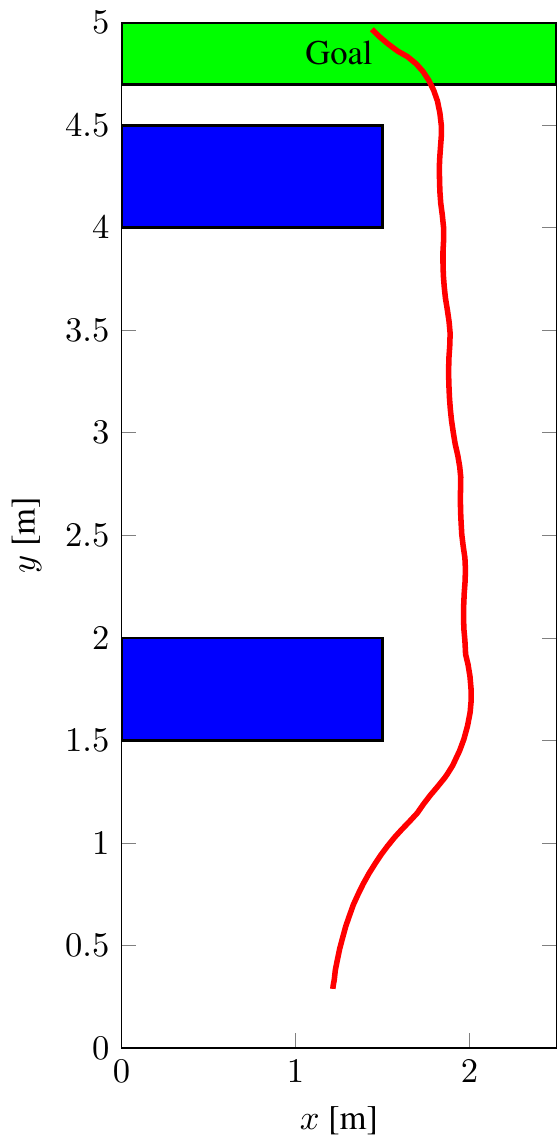}
        \includegraphics[height=0.5\textwidth]{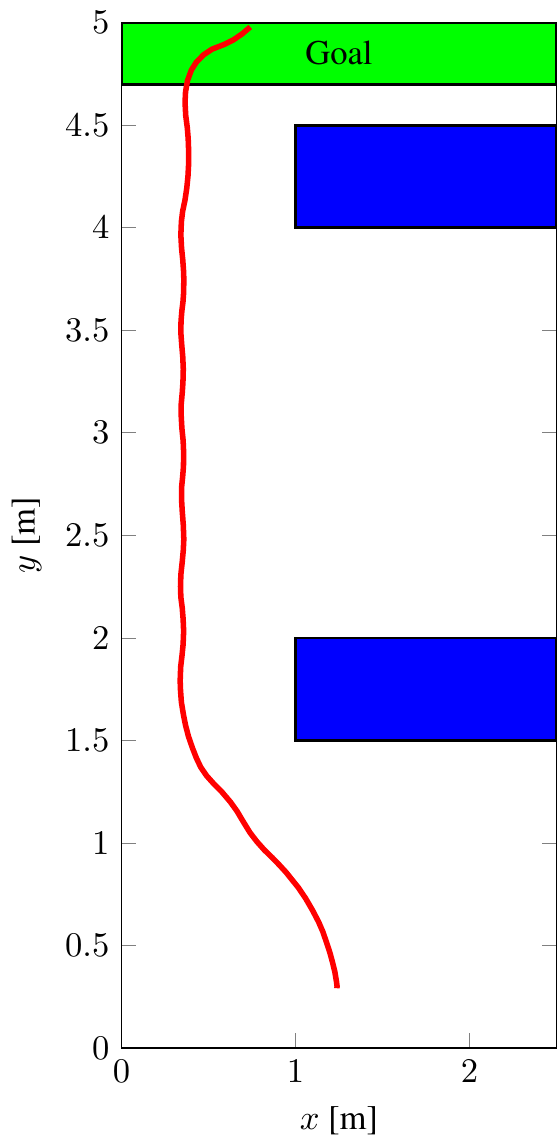}
        \\ \hline
        \rotatebox{90}{$\qquad$\Large{\textbf{PEARL (State-of-the-Art Meta-RL)}}}
        &
        \includegraphics[height=0.5\textwidth]{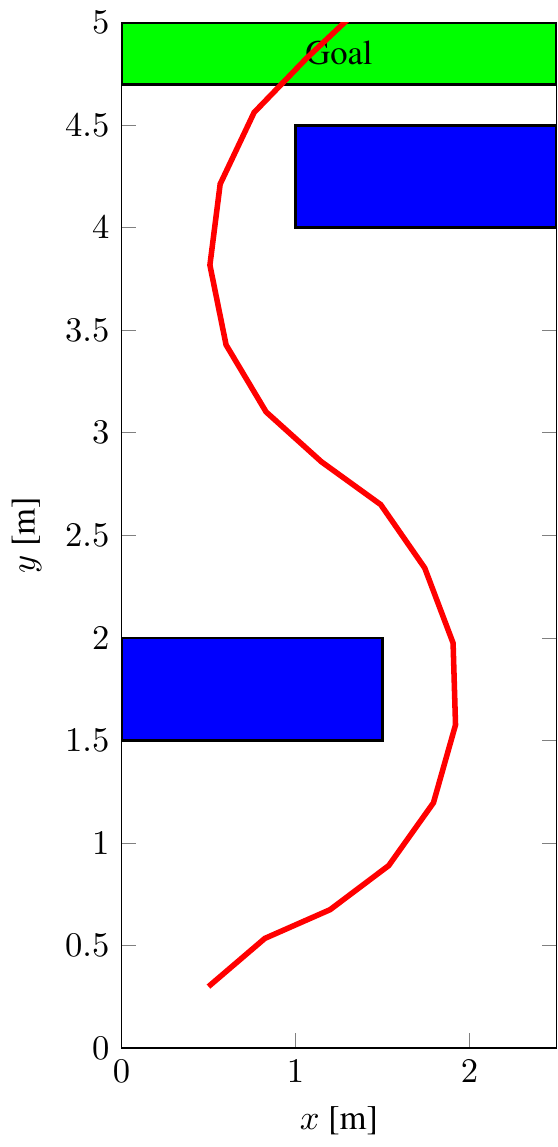}
        \includegraphics[height=0.5\textwidth]{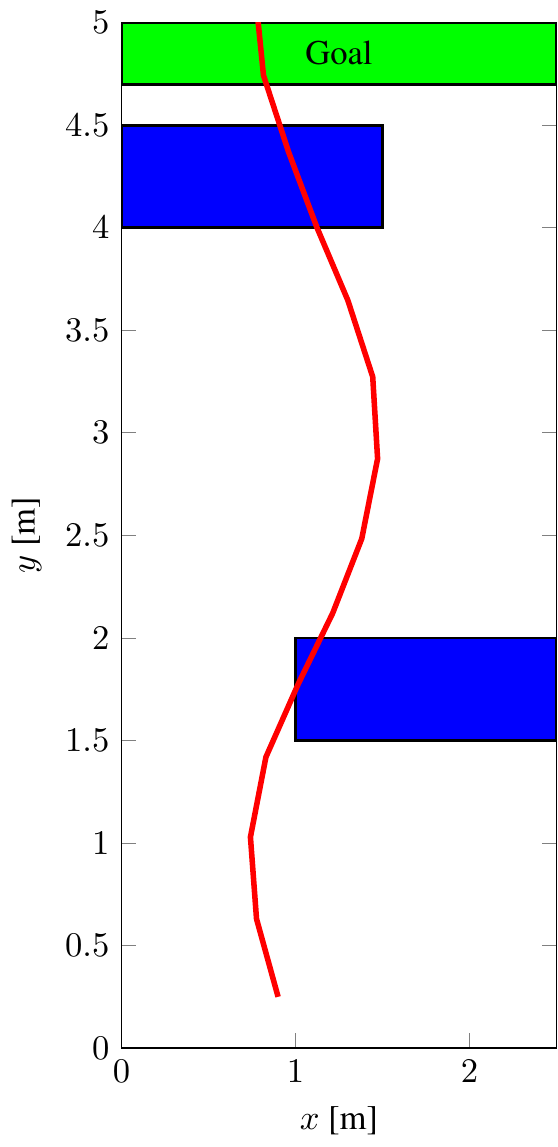}
        \includegraphics[height=0.5\textwidth]{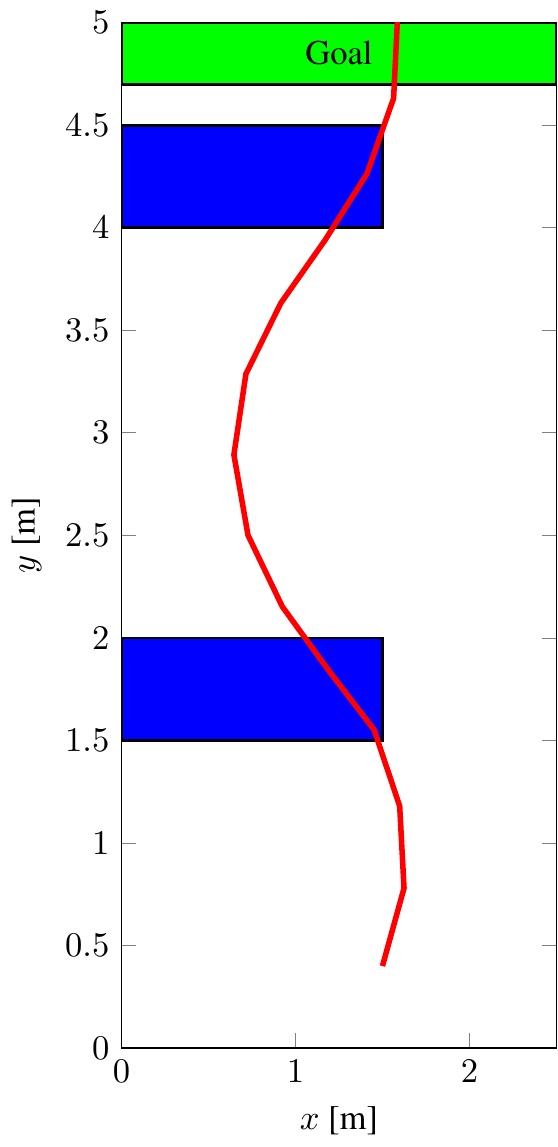}
        \includegraphics[height=0.5\textwidth]{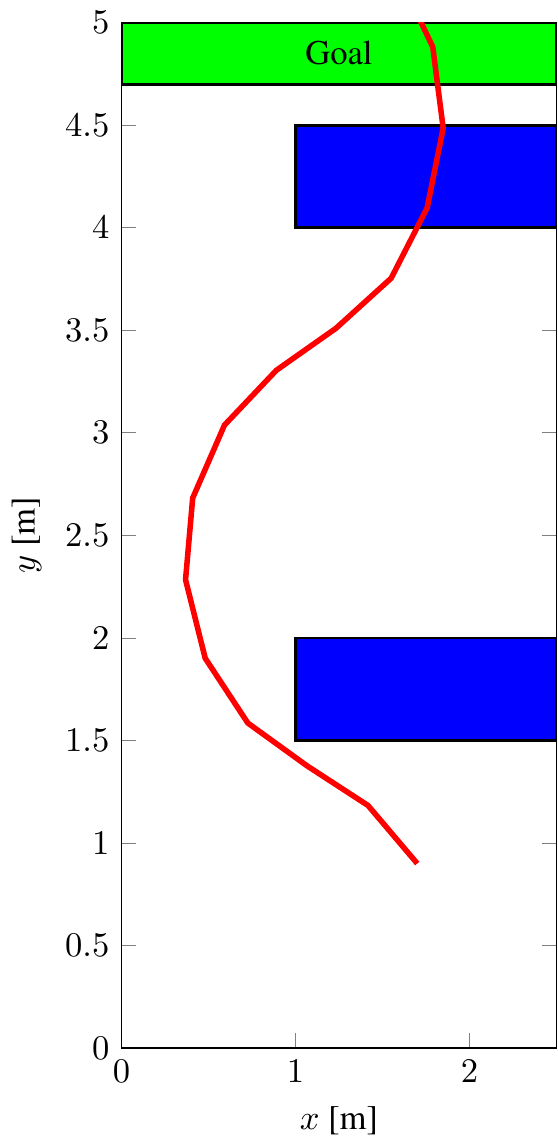}
    \end{tabular}  
    }  
    \caption{Performance comparison between our neurosymbolic framework and a state-of-the-art meta-RL algorithm PEARL. The first row shows the four workspaces used for training NNs. The second row shows the PiCar's trajectories under the NN-based planner trained by our neurosymbolic framework. All the trajectories satisfy reach-avoid specifications even in unseen workspaces. The third row shows trajectories resulting from NN controllers trained by PEARL, where the trajectory is only safe in the training workspace (the first subfigure in the third row) but unsafe in the three unseen workspaces (the rest three subfigures in the third row).} 
    \label{fig:compare_meta}  
\end{figure}

\noindent\textbf{Study\#4: Comparison against meta-RL in terms of generalization to unknown workspace/tasks.}
The objective of this study is to show the ability of our framework to generalize to unseen tasks, even in scenarios that are known to be hard for state-of-the-art meta-RL algorithms. We conducted our second experiment with the workspaces in Figure~\ref{fig:compare_meta}. In particular, the four subfigures in the first row of Figure~\ref{fig:compare_meta} are the workspaces considered for training. These four training workspaces differ in the y-coordinate of the two obstacles (blue areas). During runtime, we use the workspaces shown in the second/third row of Figure~\ref{fig:compare_meta}. Specifically, the first subfigure in the second/third rows of Figure~\ref{fig:compare_meta} corresponds to a workspace that has appeared in training. The rest three subfigures in the second/third row of Figure~\ref{fig:compare_meta} are unseen workspaces, i.e., they are not present in training and only become known at runtime.
%
Indeed, as demonstrated in~\cite{cao2021ral}, existing meta-RL algorithms are limited by the ability to adapt across homotopy classes (in Figure~\ref{fig:compare_meta}, the training tasks and the unseen tasks are in different homotopy classes since trajectories satisfying a training task cannot be continuously deformed to trajectories satisfying an unseen task without intersecting the obstacles). 

We show the PiCar's trajectories under the NN-based planner trained by our neurosymbolic framework in the second row of Figure~\ref{fig:compare_meta}. 
By following Algorithm~\ref{alg:run_transfer} with transfer learning, the PiCar's trajectories satisfy the reach-avoid specifications in all four workspaces, including the three unseen ones. Thanks to the fact that our NN-based planner is composed of local networks, our framework enables easy adaptation across homotopy classes by updating the activation map $\Gamma$ based on the revealed task (Algorithm~\ref{alg:runtime_dp}).


As a comparison, we assessed NN controllers trained by a state-of-the-art meta-RL algorithm PEARL~\cite{rakelly2019icml} in the above workspaces. Given the four training workspaces (the first row of Figure~\ref{fig:compare_meta}), we use PEARL to jointly learn a probabilistic encoder~\cite{kingma2014icml} ($3$ hidden layers with $20$ neurons per layer) and a NN controller ($3$ hidden layers with $30$ neurons per layer). The probabilistic encoder accumulates information about tasks into a vector of probabilistic context variables $z \in \mathbb{R}^5$, and the NN controller $\NN$ takes both the robot states $x$ and the context variables $z$ as input and outputs control actions $\NN(x, z)$.

When applying the trained NN controller to a task (either a training task or an unseen task) at runtime, PEARL needs to first update the posterior distribution of the context variables $z \in \mathbb{R}^5$ by collecting trajectories from the corresponding task. The third row of Figure~\ref{fig:compare_meta} shows trajectories under the control of neural networks trained by PEARL. Specifically, the first subfigure in the third row of Figure~\ref{fig:compare_meta} corresponds to a workspace that has appeared in training, and the presented trajectory is obtained after updating the posterior distribution of $z$ with $2$ trajectories collected from this workspace. The rest three subfigures in the third row of Figure~\ref{fig:compare_meta} show trajectories in unseen workspaces, where the trajectories cannot be safe even after updating the posterior distribution of $z$ with $100$ trajectories collected from the corresponding unseen workspace. By comparing trajectories resulting from our neurosymbolic framework and PEARL (the second and third rows in Figure~\ref{fig:compare_meta}), NN-based planners trained by our algorithm show the capability of adapting to unseen tasks that can be very different from training tasks. 

\subsection{Scalability Study} 
We study the scalability of our framework with respect to both partition granularity and system dimension. In this experiment, we construct the symbolic models $\hat{\Sigma}$ and assign controller partitions by following Algorithm~\ref{alg:runtime_dp}. Table~\ref{tab:granularity} reports the execution time that grows with the increasing number of abstract states and controller partitions. 
In Table~\ref{tab:dimension}, we show the scalability by increasing the system dimension $n$. To conveniently increase the system dimension, we consider a chain of integrators represented as the linear system $x^{(t+1)} = Ax^{(t)} + Bu^{(t)}$, where $A \in \mathbb{R}^{n \times n}$ is the identity matrix and $u^{(t)} \in \mathbb{R}^2$. Note that our algorithms is not aware of the linearity of the dynamics constraints nor is exploiting this fact. The algorithm has access to a simulator (the function $f$ in~\eqref{eq:dyn}) that it can use to construct the symbolic model $\hat{\Sigma}$.

To construct the symbolic models $\hat{\Sigma}$ efficiently, we adopt Algorithm~\ref{alg:construct_mdp} and only consider local controller partitions by setting the range parameter $I$ be $25$. The execution time show that our algorithm can handle a high-dimensional system in a reasonable amount of time. Although we conducted all the experiments on a single CPU core, we note that our framework is highly parallelizable. For example, both computing transition probabilities in the symbolic model $\hat{\Sigma}$ and training local networks $\NN_{(q, \mathcal{P})}$ can be parallelized. 

\begin{table}[!ht]
    \caption{Scalability with respect to Partition Granularity}
    \begin{center}
    \resizebox{.95\columnwidth}{!}{
    \begin{tabular}{|c|c|c|c|}
    \hline
    \textbf{Number of}        &\textbf{Number of}             &\textbf{Build Symbolic}           &\textbf{Assign Controller} \\ 
    \textbf{Abstract States}  &\textbf{Controller Partitions} &\textbf{Model $\hat{\Sigma}$ [s]} &\textbf{Partitions [s]} \\
    \hline\hline
    1000 & 100 & 10.1  & 21.8 \\ \cline{1-4}
    1000 & 324 & 11.3  & 69.8 \\ \cline{1-4}
    1000 & 900 & 13.3  & 193.2 \\ \cline{1-4}
    2197 & 100 & 41.6  & 74.2 \\ \cline{1-4}
    2197 & 324 & 44.7  & 227.5 \\ \cline{1-4}
    2197 & 900 & 51.3  & 673.45 \\ \cline{1-4}
    4096 & 100 & 145.6 & 383.8 \\ \cline{1-4}
    4096 & 324 & 151.2 & 1210.64 \\ \cline{1-4}
    4096 & 900 & 164.6 & 3444.43 \\ \cline{1-4}
\end{tabular}       
}
\end{center}
\label{tab:granularity} 
\end{table}

\begin{table}[!ht]
    \caption{Scalability with respect to System Dimension}
    \begin{center}
    \resizebox{.95\columnwidth}{!}{
    \begin{tabular}{|c|c|c|c|}
    \hline
    \textbf{System}        &\textbf{Number of}       &\textbf{Build Symbolic}           &\textbf{Assign Controller} \\ 
    \textbf{Dimension $n$} &\textbf{Abstract States} &\textbf{Model $\hat{\Sigma}$ [s]} &\textbf{Partitions [s]} \\
    \hline\hline
    2  & 324   & 2.1    & 1.8 \\ \cline{1-4}
    4  & 1296  & 9.4    & 10.4 \\ \cline{1-4}
    6  & 4096  & 70.3   & 62.9 \\ \cline{1-4}
    8  & 16384 & 311.2. & 158.4 \\ \cline{1-4}
    10 & 59049 & 1581.9 & 441.7 \\ \cline{1-4}
\end{tabular}       
}
\end{center}
\label{tab:dimension} 
\end{table}

\section{Conclusion}
This paper proposed a neurosymbolic framework of motion and task planning for mobile robots with respect to temporal logic formulas. By incorporating a symbolic model into the training of NNs and restricting the behavior of NNs, the resulting NN-based planner can be generalized to unseen tasks with correctness guarantees. Compared to existing techniques, our framework results in provably correct NN-based planners removing the need for online monitoring, predictive filters, barrier functions, or post-training formal verification. An interesting topic for future research is extending the framework to multiple agents with high-bandwidth sensor perception of the environment.

\section*{Acknowledgments}
This work was partially sponsored by the NSF awards \#CNS-2002405 and \#CNS-2013824.




{\appendices
\section{Section~\ref{sec:formal_training} Proofs}
\label{app:formal_training}
In this appendix, we provide proofs in Section~\ref{sec:formal_training}. 

\subsection{Proof of Proposition~\ref{prop:weight_relation}} 
\begin{proof}
    Let $h: \mathbb{R}^n \rightarrow \mathbb{R}^{\mathfrak{o}_{F-1}}$ represent all the hidden layers, then the neural networks before and after the change of the output layer weights are given by $\NN^{\:\theta}: x \mapsto W^{(F)} h(x) + b^{(F)}$ and $\NN^{\:\widehat{\theta}}: x \mapsto \widehat{W}^{(F)} h(x) + \widehat{b}^{(F)}$, respectively. The change in the NN's outputs is bounded as follows:  
    \begin{align}
        &\underset{x \in q}{\max}\; |\!|{\NN^{\:\widehat{\theta}}(x) - \NN^{\:\theta}(x)}|\!|_1  \\
        &= \underset{x \in q}{\max}\; \sum_{i=1}^{m} |\sum_{j=1}^{\mathfrak{o}_{F-1}} \Delta W_{ij}^{(F)} h_j(x) + \Delta b_i^{(F)}| \label{eq:ineq1}  \\
        &\leq \underset{x \in q}{\max}\; \sum_{i=1}^{m} \sum_{j=1}^{\mathfrak{o}_{F-1}} |\Delta W_{ij}^{(F)}| h_j(x) + \sum_{i=1}^{m} |\Delta b_i^{(F)}| \label{eq:ineq2} \\
        &\hspace{-2mm}=\underset{x \in \mathrm{Vert}(\mathbb{L}_{\NN^{\:\theta} \cap q})}{\max} \sum_{i=1}^{m} \sum_{j=1}^{\mathfrak{o}_{F-1}} |\Delta W_{ij}^{(F)}| h_j(x) + \sum_{i=1}^{m} |\Delta b_i^{(F)}| \label{eq:ineq3}
    \end{align}
    where~\eqref{eq:ineq1} directly follows the form of $\NN^{\:\theta}$ and $\NN^{\:\widehat{\theta}}$,~\eqref{eq:ineq2} swaps the order of taking the absolute value and the summation, and uses the fact that the hidden layers satisfy $h(x) \geq 0$ due to the ReLU activation function. When $x$ is restricted to each linear region of $\NN^{\:\theta}$, the hidden layer function $h$ is affine, and hence~\eqref{eq:ineq2} is a linear program whose optimal solution is attained at extreme points. Therefore, in~\eqref{eq:ineq3}, the maximum can be taken over a \emph{finite} set of states that are vertices of the linear regions in $\mathbb{L}_{\NN^{\:\theta} \cap q}$.
\end{proof}

\subsection{Proof of Proposition~\ref{prop:formal_training_linear}}
\begin{proof}
    We write the optimization problem~\eqref{eq:proj_operator_obj}-\eqref{eq:proj_operator_const} in its equivalent epigraph form:
    \begin{align}
        &\underset{\widehat{W}^{(F)}, \widehat{b}^{(F)}, t, s_{ij}, v_i}{\min} t \qquad \text{such that} \notag\\
        &\sum_{i=1}^{m} \sum_{j=1}^{\mathfrak{o}_{F-1}} s_{ij} h_j(x) + \sum_{i=1}^{m} v_i \leq t,\; \forall x \in \mathrm{Vert}(\mathbb{L}_{\NN^{\:\theta} \cap q}) \label{eq:const_t} \\
        &|\widehat{W}_{ij}^{(F)} - W_{ij}^{(F)}| \leq s_{ij},\; i =1, \ldots, m,\; j =1, \ldots, \mathfrak{o}_{F-1}\\
        &|\widehat{b}_i^{(F)} - b_i^{(F)}| \leq v_i,\; i =1, \ldots, m\\
        &\widehat{K}_i \in \mathcal{P},\; \forall \mathcal{R}_i \in \{\mathcal{R} \in \mathbb{L}_{\NN^{\:\theta}}\; |\; \mathcal{R} \cap q \neq \emptyset\}. \label{eq:const_p}
    \end{align}
    The inequalities in~\eqref{eq:const_t} are affine since the hidden layer function $h$ is known and does not depend on the optimization variables. The number of inequalities in~\eqref{eq:const_t} is finite since the set of vertices $\mathrm{Vert}(\mathbb{L}_{\NN^{\:\theta} \cap q})$ is finite. To see the constraints~\eqref{eq:const_p} are affine, consider the neural network $\NN^{\:\widehat{\theta}}: x \mapsto \widehat{W}^{(F)} h(x) + \widehat{b}^{(F)}$ with the output layer weights $\widehat{W}^{(F)}$, $\widehat{b}^{(F)}$ and the hidden layer function $h$. The CPWA function $\NN^{\:\widehat{\theta}}$ can also be written in the form of~\eqref{eq:cpwa}, i.e., $\NN^{\:\widehat{\theta}}: x \mapsto \widehat{K}_i(x)$ at each linear region $\mathcal{R}_i \in \mathbb{L}_{\NN^{\:\theta}}$, where we use the notation $\widehat{K}_i(x)$ to denote $\widehat{K}^\prime_i x + \widehat{b}^\prime_i$. Since the hidden-layer function $h$ restricted to each linear region $\mathcal{R}_i \in \mathbb{L}_{\NN^{\:\theta}}$ is a known affine function of $x$, the parameters $\widehat{K}_i$ affinely depend on $\widehat{W}^{(F)}$ and $\widehat{b}^{(F)}$. Therefore, the constraints $\widehat{K}_i \in \mathcal{P}$ are affine constraints of  $\widehat{W}^{(F)}$ and $\widehat{b}^{(F)}$.
\end{proof}

\section{Section~\ref{sec:guarantees} Proofs}
\label{app:guarantees}
In this appendix, we provide proofs of Theorem~\ref{thm:nn_v} and Theorem~\ref{thm:nn_optimal} in Section~\ref{sec:guarantees}. Let $\Sigma = (X, X_0, U, t)$ be a robotic system with continuous state and action spaces and $\mathcal{A}_\varphi = (S, S_0, \mathbb{A}, G, \delta)$ be the DFA of a mission specification $\varphi$. Similar to the product MDP $\hat{\Sigma} \otimes \mathcal{A}_\varphi$, the product between $\Sigma$ and $\mathcal{A}_\varphi$ is given by $\Sigma \otimes \mathcal{A}_\varphi = (X^\otimes, X_0^\otimes, U, X_G^\otimes, t^\otimes)$, where:
\begin{itemize}
    \item $X^\otimes = X \times S$ is the state space;
    \item $X_0^\otimes = \{(x_0, \delta(s_0, L(x_0)) | x_0 \in X_0, s_0 \in S_0\}$ is the set of initial states, where $L: X \rightarrow \mathbb{A}$ is the labeling function that assigns to each state $x \in X$ the subset of atomic propositions $L(x) \in \mathbb{A}$ that evaluate $true$ at $x$;
    \item $U \subset \mathbb{R}^m$ is the control action space;
    \item $X_G^\otimes = X \times G$ is the accepting set;
    \item The stochastic kernel $t^\otimes$ is given by:
    \begin{align}
        t^\otimes(dx^\prime, s^\prime | x, s, u)=
        \begin{cases}
            t(dx^\prime | x, u)\ &\text{if } s^\prime = \delta(s, L(x^\prime)) \\
            0\ &\text{else}. \notag
        \end{cases}
    \end{align}
\end{itemize}

\subsection{Proof of Theorem~\ref{thm:nn_v}} 
\begin{proof} 
    Given the NN-based planner $\NN_{[\mathfrak{NN},\Gamma]}$ obtained using our framework, we define functions $V_k^\NN: X^\otimes \rightarrow [0, 1]$ that map a state $(x, s) \in X^\otimes$ to the probability of reaching the accepting set $X_G^\otimes$ in $H-k$ steps from the state $(x, s)$ and under the control of $\NN_{[\mathfrak{NN},\Gamma]}$. With this notation, we have $V_0^\NN (x, s) = \pr \left(\xi_{\NN_{[\mathfrak{NN},\Gamma]}}^{(x, s)} \models \varphi \right)$ since reaching the accepting set $X_G^\otimes$ in $H$ steps in the product MDP $\Sigma \otimes \mathcal{A}_\varphi$ is equivalent to $\Sigma$ satisfying $\varphi$. In the following, we show that for any $x \in q$ and $k = 0, \ldots, H$:
    \begin{equation}
        \label{eq:proof1_v_nn_v_hat}
        |V_k^\NN(x, s) - \hat{V}^*_k(q, s)| \leq  (H-k) Z \Delta^\NN, 
    \end{equation}
    which yields~\eqref{eq:nn_v} by letting $k = 0$. By the definition of $V_k^\NN$, the probabilities of reaching the accepting set $X_G^\otimes$ under the NN-based planner $\NN_{[\mathfrak{NN},\Gamma]}$ can be expressed as:
    \begin{align}
        \label{eq:proof1_v_nn}
        &V_k^\NN(x, s) = \mathbf{1}_{G}(s) \notag \\
        &+ \mathbf{1}_{S \setminus G}(s) \sum_{s^\prime \in S} \int_X V_{k+1}^\NN (x^\prime, s^\prime) t^\otimes(dx^\prime, s^\prime | x, s, \NN(x)) 
    \end{align} 
    with the initial condition $V_H^\NN(x, s) = \mathbf{1}_{G}(s)$. In the stochastic kernel $t^\otimes$ in~\eqref{eq:proof1_v_nn}, we use $\NN$ to denote the local network selected by the activation map $\Gamma_{k+1}$ at the state $(x, s)$ for simplicity. Though solving~\eqref{eq:proof1_v_nn} is intractable due to the continuous state space, we can bound the difference between $V_k^\NN$ and $\hat{V}_k^*$ as~\eqref{eq:proof1_v_nn_v_hat} by induction. 

    For the base case $k = H$,~\eqref{eq:proof1_v_nn_v_hat} trivially holds since $V_H^\NN(x, s) = \mathbf{1}_{G}(s)$ and $\hat{V}_H^*(q, s) = \mathbf{1}_{G}(s)$. For the induction hypothesis, suppose for $k+1$ it holds that:
    \begin{equation}
        \label{eq:proof1_hypothesis}
        |V_{k+1}^\NN(x, s) - \hat{V}^*_{k+1}(q, s)| \leq  (H-k-1) Z \Delta^\NN.
    \end{equation}
    Let $\bar{V}^*_k$ be a piecewise constant interpolation of $\hat{V}^*_k$ defined by $\bar{V}^*_k(x, s) = \hat{V}^*_k(q, s)$ for any $x \in q$ and any $s \in S$. Then,
    \begin{align}
        \label{eq:proof1_induction}
        &|V_k^\NN(x, s) - \hat{V}^*_k(q, s)|  \notag \\
        &\leq |V_k^\NN(x, s) - V_k^\NN(z, s)| + |V_k^\NN(z, s) - \bar{V}_k^*(z, s)| 
    \end{align}
    where $z = \ct_\mathbb{X}(q)$ and $x \in q$. For the first term on the RHS:
    \begin{align}
        &|V_k^\NN(x, s) - V_k^\NN(z, s)| \notag \\
        &= |\mathbf{1}_{G}(s)\hspace{-1mm} +\hspace{-1mm} \mathbf{1}_{S \setminus G}(s)\hspace{-1mm} \sum_{s^\prime \in S} \int_{X}\hspace{-1mm} V_{k+1}^\NN (x^\prime, s^\prime) t^\otimes(dx^\prime, s^\prime |x, s, \NN(x)) \notag \\
        &\hspace{1.5mm}- \mathbf{1}_{G}(s)\hspace{-1mm} +\hspace{-1mm} \mathbf{1}_{S \setminus G}(s)\hspace{-1mm} \sum_{s^\prime \in S} \int_{X}\hspace{-1mm} V_{k+1}^\NN (x^\prime, s^\prime) t^\otimes(dx^\prime, s^\prime |z, s, \NN(z))| \notag \\
        &\leq \sum_{s^\prime \in S} \int_{X} V_{k+1}^\NN (x^\prime, s^\prime) |t^\otimes(dx^\prime, s^\prime | x, s, \NN(x)) \notag \\
        &\hspace{30mm} - t^\otimes(dx^\prime, s^\prime | z, s, \NN(z)| \notag \\
        &\leq Z \int_{X} |t(dx^\prime | x, \NN(x)) - t(dx^\prime | z, \NN(z)| \notag \\
        &\leq Z \int_{X} |t(dx^\prime|x, \NN(x)) - t(dx^\prime|z, \NN(x))| \notag \\
        &\hspace{9mm} + |t(dx^\prime|z, \NN(x)) - t(dx^\prime|z, \NN(z))| \notag \\
        &\leq Z \Lambda_i \norm{x - z} + Z B_i \norm{\NN(x) - \NN(z)} \notag \\
        &\leq Z \Lambda_i \eta_q + Z B_i L_i \eta_q. \label{eq:proof1_1st}
    \end{align}
    For the second term on the RHS of~\eqref{eq:proof1_induction}:
        \begin{align}
        &|V_k^\NN(z, s) - \bar{V}_k^*(z, s)| \notag \\
        &= |\mathbf{1}_{G}(s)\hspace{-1mm} +\hspace{-1mm} \mathbf{1}_{S \setminus G}(s)\hspace{-1mm} \sum_{s^\prime \in S}\int_{X}\hspace{-1mm} V_{k+1}^\NN (x^\prime, s^\prime) t^\otimes(dx^\prime, s^\prime | z, s, \NN(z)) \notag \\
        &\hspace{1.5mm} - \mathbf{1}_{G}(s)\hspace{-1mm} +\hspace{-1mm} \mathbf{1}_{S \setminus G}(s) \max_{\mathcal{P} \in \mathbb{P}}\hspace{-2mm} \sum_{(q^\prime, s^\prime) \in \mathbb{X}^\otimes}\hspace{-2mm} \hat{V}_{k+1}^* (q^\prime, s^\prime) \hat{t}^\otimes(q^\prime, s^\prime | q, s, \mathcal{P})| \label{eq:proof1_2nd_eq1}\\
        &\leq |\sum_{s^\prime \in S}\int_{X} V_{k+1}^\NN (x^\prime, s^\prime) t^\otimes(dx^\prime, s^\prime | z, s, \NN(z)) \notag \\
        &\hspace{2mm} - \sum_{s^\prime \in S} \sum_{q^\prime \in \mathbb{X}} \hat{V}_{k+1}^* (q^\prime, s^\prime) \hat{t}^\otimes(q^\prime, s^\prime | q, s, \mathcal{P}^*)| \label{eq:proof1_2nd_eq2}\\
        &\leq |\sum_{s^\prime \in S}\int_{X} V_{k+1}^\NN (x^\prime, s^\prime) t^\otimes(dx^\prime, s^\prime | z, s, \NN(z)) \notag \\
        &\hspace{2mm} - \sum_{s^\prime \in S} \int_X \bar{V}_{k+1}^* (x^\prime, s^\prime) t^\otimes(dx^\prime, s^\prime | z, s, \ct_\mathbb{P}(\mathcal{P}^*)(z))| \label{eq:proof1_2nd_eq3}\\
        &\leq \sum_{s^\prime \in S} \int_X |V_{k+1}^\NN (x^\prime, s^\prime) - \bar{V}_{k+1}^* (x^\prime, s^\prime) | t^\otimes(dx^\prime, s^\prime | z, s, \NN(z)) \notag \\
        &+ \sum_{s^\prime \in S} \int_{X} \bar{V}_{k+1}^* (x^\prime, s^\prime) |t^\otimes(dx^\prime, s^\prime | z, s, \NN(z)) \notag \\
        &\hspace{30mm}- t^\otimes(dx^\prime, s^\prime | z, s, \ct_\mathbb{P}(\mathcal{P}^*)(z))| \label{eq:proof1_2nd_eq4} \\
        &\leq (H-k-1) Z \Delta^\NN + Z \sqrt{m(n+1)} \mathcal{L}_X B_i \eta_\mathcal{P} \label{eq:proof1_2nd_eq5}
    \end{align}
    where~\eqref{eq:proof1_2nd_eq1} uses the DP recursion~\eqref{eq:abst_Q}-\eqref{eq:abst_V}, in~\eqref{eq:proof1_2nd_eq2} $\mathcal{P}^*$ denotes the maximizer, and~\eqref{eq:proof1_2nd_eq3} uses the definition of $\hat{t}$ in~\eqref{eq:integral_hat_t} with $z = \ct_\mathbb{X}(q)$. In~\eqref{eq:proof1_2nd_eq5}, we use the induction hypothesis~\eqref{eq:proof1_hypothesis}, and the inequality $\norm{K(x) - K^\prime(x)} \leq \norm{K - K^\prime} \norm{x} \leq \sqrt{m(n+1)} \norm{K - K^\prime}_{\max} \mathcal{L}_X \leq \sqrt{m(n+1)} \eta_\mathcal{P} \mathcal{L}_X$, where $\norm{K - K^\prime}_{\max} \leq \eta_\mathcal{P}$ since the local network $\NN$ selected by the activation map $\Gamma$ represents a CPWA function from the maximizer $\mathcal{P}^\star$, i.e., $K, K^\prime \in \mathcal{P}^\star \subset \mathbb{R}^{m \times (n+1)}$. Substitute~\eqref{eq:proof1_1st} and~\eqref{eq:proof1_2nd_eq5} into~\eqref{eq:proof1_induction} yields~\eqref{eq:proof1_v_nn_v_hat}.
\end{proof}

\subsection{Proof of Theorem~\ref{thm:nn_optimal}} 


\begin{proof}
Let functions $V_k^*: X^\otimes \rightarrow [0, 1]$ map a state $(x, s) \in X^\otimes$ to the probability of reaching the accepting set $X_G^\otimes$ in $H-k$ steps from the state $(x, s)$ and under the optimal controller $\mathcal{C}_\varphi^*: X \times S \rightarrow U$. Then, $V_0^* (x, s) = \pr \left(\xi_{\mathcal{C}_\varphi^*}^{(x, s)} \models \varphi \right)$ since reaching the accepting set $X_G^\otimes$ in $H$ steps in the product MDP $\Sigma \otimes \mathcal{A}_\varphi$ is equivalent to $\Sigma$ satisfying $\varphi$. The optimal probabilities of reaching the accepting set $X_G^\otimes$ can be expressed using DP recursion: 
\begin{align}
    Q_k(x, s, u) &= \mathbf{1}_{G}(s) \label{eq:continuous_Q}\\
    &+ \mathbf{1}_{S \setminus G}(s) \sum_{ s^\prime \in S} \int_X V_{k+1}^* (x^\prime, s^\prime) t^\otimes(dx^\prime, s^\prime | x, s, u) \notag \\
    V_k^*(x, s) &= \max_{u \in U}\; Q_k(x, s, u) \label{eq:continuous_V}
\end{align} 
Though solving $V_{k}^*$ and the corresponding optimal controller $\mathcal{C}_\varphi^*$ is intractable due to the continuous state and action spaces, we can bound the difference between $V_{k}^*$ and $\hat{V}_{k}^*$ by induction similar to the proof of~\eqref{eq:proof1_v_nn_v_hat} in Theorem~\ref{thm:nn_v}. We skip the details and directly give the following bound: 
\begin{equation}
    \label{eq:proof2_v_star_v_hat}
    |V_{k}^*(x, s) - \hat{V}^*_{k}(q, s)| \leq  (H-k) Z \Delta^*,
\end{equation}
where $x \in q$ and $\Delta^*$ is given by~\eqref{eq:delta_optimal}. With~\eqref{eq:proof1_v_nn_v_hat} and~\eqref{eq:proof2_v_star_v_hat}, we have:
\begin{align}
    &|V_{k}^\NN(x, s) - V^*_{k}(x, s)| \notag \\
    &\leq |V_{k}^\NN(x, s) - \hat{V}^*_{k}(q, s)| + |V_{k}^*(x, s) - \hat{V}^*_{k}(q, s)| \notag \\
    &\leq  (H-k) Z (\Delta^\NN + \Delta^*),
\end{align}
which yields~\eqref{eq:nn_optimal} by letting $k = 0$. 
\end{proof}

}

\bibliographystyle{IEEEtran} 
\bibliography{biblio.bib}

\begin{thebibliography}{10}
\providecommand{\url}[1]{#1}
\csname url@samestyle\endcsname
\providecommand{\newblock}{\relax}
\providecommand{\bibinfo}[2]{#2}
\providecommand{\BIBentrySTDinterwordspacing}{\spaceskip=0pt\relax}
\providecommand{\BIBentryALTinterwordstretchfactor}{4}
\providecommand{\BIBentryALTinterwordspacing}{\spaceskip=\fontdimen2\font plus
\BIBentryALTinterwordstretchfactor\fontdimen3\font minus
  \fontdimen4\font\relax}
\providecommand{\BIBforeignlanguage}[2]{{%
\expandafter\ifx\csname l@#1\endcsname\relax
\typeout{** WARNING: IEEEtran.bst: No hyphenation pattern has been}%
\typeout{** loaded for the language `#1'. Using the pattern for}%
\typeout{** the default language instead.}%
\else
\language=\csname l@#1\endcsname
\fi
#2}}
\providecommand{\BIBdecl}{\relax}
\BIBdecl

\bibitem{bltl}
A.~Biere, K.~Heljanko, T.~Junttila, T.~Latvala, and V.~Schuppan, ``Linear
  encodings of bounded \textsc{LTL} model checking,'' \emph{Logical Methods in
  Computer Science}, vol.~2, no. 5:5, pp. 1--64, 2006.

\bibitem{scltl}
O.~Kupferman and M.~Y. Vardi, ``Model checking of safety properties,''
  \emph{Formal Methods in System Design}, pp. 19:291--314, 2001.

\bibitem{kress2011correct}
H.~Kress-Gazit, T.~Wongpiromsarn, and U.~Topcu, ``Correct, reactive, high-level
  robot control,'' \emph{IEEE Robotics \& Automation Magazine}, vol.~18, no.~3,
  pp. 65--74, 2011.

\bibitem{kress2018synthesis}
H.~Kress-Gazit, M.~Lahijanian, and V.~Raman, ``Synthesis for robots: Guarantees
  and feedback for robot behavior,'' \emph{Annual Review of Control, Robotics,
  and Autonomous Systems}, vol.~1, pp. 211--236, 2018.

\bibitem{bhatia2011motion}
A.~Bhatia, M.~R. Maly, L.~E. Kavraki, and M.~Y. Vardi, ``Motion planning with
  complex goals,'' \emph{IEEE Robotics \& Automation Magazine}, vol.~18, no.~3,
  pp. 55--64, 2011.

\bibitem{garrett2018ffrob}
C.~R. Garrett, T.~Lozano-Perez, and L.~P. Kaelbling, ``Ffrob: Leveraging
  symbolic planning for efficient task and motion planning,'' \emph{The
  International Journal of Robotics Research}, vol.~37, no.~1, pp. 104--136,
  2018.

\bibitem{guo2013motion}
M.~Guo, K.~H. Johansson, and D.~V. Dimarogonas, ``Motion and action planning
  under ltl specifications using navigation functions and action description
  language,'' in \emph{2013 IEEE/RSJ International Conference on Intelligent
  Robots and Systems}.\hskip 1em plus 0.5em minus 0.4em\relax IEEE, 2013, pp.
  240--245.

\bibitem{bhatia2010sampling}
A.~Bhatia, L.~E. Kavraki, and M.~Y. Vardi, ``Sampling-based motion planning
  with temporal goals,'' in \emph{2010 IEEE International Conference on
  Robotics and Automation}.\hskip 1em plus 0.5em minus 0.4em\relax IEEE, 2010,
  pp. 2689--2696.

\bibitem{fainekos2005hybrid}
G.~E. Fainekos, H.~Kress-Gazit, and G.~J. Pappas, ``Hybrid controllers for path
  planning: A temporal logic approach,'' in \emph{Proceedings of the 44th IEEE
  Conference on Decision and Control}.\hskip 1em plus 0.5em minus 0.4em\relax
  IEEE, 2005, pp. 4885--4890.

\bibitem{fainekos2006translating}
G.~E. Fainekos, S.~G. Loizou, and G.~J. Pappas, ``Translating temporal logic to
  controller specifications,'' in \emph{Proceedings of the 45th IEEE Conference
  on Decision and Control}.\hskip 1em plus 0.5em minus 0.4em\relax IEEE, 2006,
  pp. 899--904.

\bibitem{kress2007s}
H.~Kress-Gazit, G.~E. Fainekos, and G.~J. Pappas, ``Where's waldo? sensor-based
  temporal logic motion planning,'' in \emph{Proceedings 2007 IEEE
  International Conference on Robotics and Automation}.\hskip 1em plus 0.5em
  minus 0.4em\relax IEEE, 2007, pp. 3116--3121.

\bibitem{shoukry2017linear}
Y.~Shoukry, P.~Nuzzo, A.~Balkan, I.~Saha, A.~L. Sangiovanni-Vincentelli, S.~A.
  Seshia, G.~J. Pappas, and P.~Tabuada, ``Linear temporal logic motion planning
  for teams of underactuated robots using satisfiability modulo convex
  programming,'' in \emph{2017 IEEE 56th annual conference on decision and
  control (CDC)}.\hskip 1em plus 0.5em minus 0.4em\relax IEEE, 2017, pp.
  1132--1137.

\bibitem{tabuada2009verification}
P.~Tabuada, \emph{Verification and control of hybrid systems: a symbolic
  approach}.\hskip 1em plus 0.5em minus 0.4em\relax Springer Science \&
  Business Media, 2009.

\bibitem{belta2017formal}
C.~Belta, B.~Yordanov, and E.~A. Gol, \emph{Formal methods for discrete-time
  dynamical systems}.\hskip 1em plus 0.5em minus 0.4em\relax Springer, 2017,
  vol.~15.

\bibitem{cao2021ral}
Z.~Cao, M.~Kwon, and D.~Sadigh, ``Transfer reinforcement learning across
  homotopy classes,'' in \emph{IEEE Robotics and Automation Letters}, 2021.

\bibitem{sun2022cdc}
X.~Sun, W.~Fatnassi, U.~{Santa Cruz}, and Y.~Shoukry, ``Provably safe
  model-based meta reinforcement learning: An abstraction-based approach,'' in
  \emph{60th IEEE Conference on Decision and Control}, 2021, pp. 2963--2968.

\bibitem{stone2020reward}
Y.~Jiang, S.~Bharadwaj, B.~Wu, R.~Shah, U.~Topcu, and P.~Stone,
  ``Temporal-logic-based reward shaping for continuing learning tasks,'' in
  \emph{Proceedings of the 33rd AAAI Conference on Artificial Intelligence},
  2020, pp. 7995--8003.

\bibitem{saunders2018trial}
W.~Saunders, G.~Sastry, A.~Stuhlmueller, and O.~Evans, ``Trial without error:
  Towards safe reinforcement learning via human intervention,'' in
  \emph{Proceedings of the 17th International Conference on Autonomous Agents
  and MultiAgent Systems}, 2018, pp. 2067--2069.

\bibitem{berkenkamp2016bayesian}
F.~Berkenkamp, A.~Krause, and A.~P. Schoellig, ``Bayesian optimization with
  safety constraints: safe and automatic parameter tuning in robotics,'' in
  \emph{Machine Learning}.\hskip 1em plus 0.5em minus 0.4em\relax Springer,
  2021.

\bibitem{liu2019robust}
A.~Liu, G.~Shi, S.-J. Chung, A.~Anandkumar, and Y.~Yue, ``Robust regression for
  safe exploration in control,'' in \emph{Proceedings of Machine Learning
  Research}, 2020.

\bibitem{pauli2020training}
P.~Pauli, A.~Koch, J.~Berberich, and F.~Allg{\"o}wer, ``Training robust neural
  networks using lipschitz bounds,'' \emph{IEEE Control Systems Letters}, 2020.

\bibitem{abbeel2017icml}
J.~Achiam, D.~Held, A.~Tamar, and P.~Abbeel, ``Constrained policy
  optimization,'' in \emph{Proceedings of the 34th International Conference on
  Machine Learning}, 2017, pp. 22--31.

\bibitem{turchetta2016safe}
M.~Turchetta, F.~Berkenkamp, and A.~Krause, ``Safe exploration in finite markov
  decision processes with gaussian processes,'' in \emph{Advances in Neural
  Information Processing Systems}, 2016, pp. 4312--4320.

\bibitem{wen2020safe}
L.~Wen, J.~Duan, S.~E. Li, S.~Xu, and H.~Peng, ``Safe reinforcement learning
  for autonomous vehicles through parallel constrained policy optimization,''
  in \emph{IEEE 23rd International Conference on Intelligent Transportation
  Systems}, 2020.

\bibitem{dutta2018output}
S.~Dutta, S.~Jha, S.~Sankaranarayanan, and A.~Tiwari, ``Output range analysis
  for deep feedforward neural networks,'' in \emph{NASA Formal Methods
  Symposium}.\hskip 1em plus 0.5em minus 0.4em\relax Springer, 2018.

\bibitem{liu2019algorithms}
C.~Liu, T.~Arnon, C.~Lazarus, C.~Barrett, and M.~J. Kochenderfer, ``Algorithms
  for verifying deep neural networks,'' \emph{arXiv preprint arXiv:1903.06758},
  2019.

\bibitem{sun2019formal}
X.~Sun, H.~Khedr, and Y.~Shoukry, ``Formal verification of neural network
  controlled autonomous systems,'' in \emph{Proceedings of the 22nd ACM
  International Conference on Hybrid Systems: Computation and Control}, 2019,
  pp. 147--156.

\bibitem{khedr2021peregrinn}
H.~Khedr, J.~Ferlez, and Y.~Shoukry, ``Peregrinn: Penalized-relaxation greedy
  neural network verifier,'' in \emph{International Conference on Computer
  Aided Verification}.\hskip 1em plus 0.5em minus 0.4em\relax Springer, 2021,
  pp. 287--300.

\bibitem{ferlez2022fast}
J.~Ferlez, H.~Khedr, and Y.~Shoukry, ``Fast batllnn: fast box analysis of
  two-level lattice neural networks,'' in \emph{25th ACM International
  Conference on Hybrid Systems: Computation and Control}, 2022, pp. 1--11.

\bibitem{santa2022nnlander}
U.~Santa~Cruz and Y.~Shoukry, ``Nnlander-verif: A neural network formal
  verification framework for vision-based autonomous aircraft landing,'' in
  \emph{NASA Formal Methods Symposium}.\hskip 1em plus 0.5em minus 0.4em\relax
  Springer, 2022, pp. 213--230.

\bibitem{fazlyab2019efficient}
M.~Fazlyab, A.~Robey, H.~Hassani, M.~Morari, and G.~Pappas, ``Efficient and
  accurate estimation of lipschitz constants for deep neural networks,'' in
  \emph{Advances in Neural Information Processing Systems}, 2019, pp.
  11\,423--11\,434.

\bibitem{ivanov2019verisig}
R.~Ivanov, J.~Weimer, R.~Alur, G.~J. Pappas, and I.~Lee, ``Verisig: verifying
  safety properties of hybrid systems with neural network controllers,'' in
  \emph{Proceedings of the 22nd ACM International Conference on Hybrid Systems:
  Computation and Control}, 2019, pp. 169--178.

\bibitem{xiang2019reachable}
W.~Xiang, D.~M. Lopez, P.~Musau, and T.~T. Johnson, ``Reachable set estimation
  and verification for neural network models of nonlinear dynamic systems,'' in
  \emph{Safe, Autonomous and Intelligent Vehicles}.\hskip 1em plus 0.5em minus
  0.4em\relax Springer, 2019, pp. 123--144.

\bibitem{fisac2018general}
J.~F. Fisac, A.~K. Akametalu, M.~N. Zeilinger, S.~Kaynama, J.~Gillula, and
  C.~J. Tomlin, ``A general safety framework for learning-based control in
  uncertain robotic systems,'' \emph{IEEE Transactions on Automatic Control},
  vol.~64, no.~7, pp. 2737--2752, 2018.

\bibitem{fisac2021rss}
K.-C. Hsu, V.~Rubies-Royo, C.~J. Tomlin, and J.~F. Fisac, ``Safety and liveness
  guarantees through reach-avoid reinforcement learning,'' in \emph{Robotics:
  Science and Systems}, 2021.

\bibitem{abate2021hscc}
A.~Abate, D.~Ahmed, A.~Edwards, M.~Giacobbe, and A.~Peruffo, ``Fossil: A
  software tool for the formal synthesis of lyapunov functions and barrier
  certificates using neural networks,'' in \emph{Proceedings of the 24th ACM
  International Conference on Hybrid Systems: Computation and Control}, 2021.

\bibitem{pappas2021hscc}
S.~Chen, M.~Fazlyab, M.~Morari, G.~J. Pappas, and V.~M. Preciado, ``Learning
  lyapunov functions for hybrid systems,'' in \emph{Proceedings of the 24th ACM
  International Conference on Hybrid Systems: Computation and Control}, 2021.

\bibitem{cheng2019end}
R.~Cheng, G.~Orosz, R.~M. Murray, and J.~W. Burdick, ``End-to-end safe
  reinforcement learning through barrier functions for safety-critical
  continuous control tasks,'' in \emph{Proceedings of the AAAI Conference on
  Artificial Intelligence}, vol.~33, 2019, pp. 3387--3395.

\bibitem{matni2020cdc}
A.~Robey, H.~Hu, L.~Lindemann, H.~Zhang, D.~V. Dimarogonas, S.~Tu, and
  N.~Matni, ``Learning control barrier functions from expert demonstrations,''
  in \emph{59th IEEE Conference on Decision and Control}, 2020.

\bibitem{taylor2020control}
A.~J. Taylor, A.~Singletary, Y.~Yue, and A.~D. Ames, ``A control barrier
  perspective on episodic learning via projection-to-state safety,'' in
  \emph{IEEE Control Systems Letters}, 2021, pp. 1019--1024.

\bibitem{wang2018safe}
L.~Wang, E.~A. Theodorou, and M.~Egerstedt, ``Safe learning of quadrotor
  dynamics using barrier certificates,'' in \emph{IEEE International Conference
  on Robotics and Automation}, 2018, pp. 2460--2465.

\bibitem{cassandras2020adaptivecbf}
W.~Xiao, C.~Belta, and C.~G. Cassandras, ``Adaptive control barrier
  functions,'' in \emph{IEEE Transactions on Automatic Control}, 2022.

\bibitem{bastani2021rss}
O.~Bastani, S.~Li, and A.~Xu, ``Safe reinforcement learning via statistical
  model predictive shielding,'' in \emph{Robotics: Science and Systems}, 2021.

\bibitem{zeilinger2018cdc}
K.~P. Wabersich and M.~N. Zeilinger, ``Linear model predictive safety
  certification for learning-based control,'' in \emph{57th IEEE Conference on
  Decision and Control}, 2018.

\bibitem{zeilinger2021filter}
------, ``A predictive safety filter for learning-based control of constrained
  nonlinear dynamical systems,'' in \emph{Automatica}, 2021.

\bibitem{hasanbeig2019cdc}
M.~Hasanbeig, Y.~Kantaros, A.~Abate, D.~Kroening, G.~J. Pappas, , and I.~Lee,
  ``Reinforcement learning for temporal logic control synthesis with
  probabilistic satisfaction guarantees,'' in \emph{58th IEEE Conference on
  Decision and Control}, 2019, pp. 5338--5343.

\bibitem{balakrishnan2019iros}
A.~Balakrishnan and J.~V. Deshmukh, ``Structured reward shaping using signal
  temporal logic specifications,'' in \emph{IEEE/RSJ International Conference
  on Intelligent Robots and Systems}, 2019, pp. 3481--3486.

\bibitem{alshiekh2018aaai}
M.~Alshiekh, R.~Bloem, R.~Ehlers, B.~Könighofer, S.~Niekum, and UfukTopcu,
  ``Safe reinforcement learning via shielding,'' in \emph{Proceedings of the
  32nd AAAI Conference on Artificial Intelligence}, 2018, pp. 2669--2678.

\bibitem{berkenkamp2017safe}
F.~Berkenkamp, M.~Turchetta, A.~Schoellig, and A.~Krause, ``Safe model-based
  reinforcement learning with stability guarantees,'' in \emph{Advances in
  neural information processing systems}, 2017.

\bibitem{chow2018lyapunov}
Y.~Chow, O.~Nachum, E.~Duenez-Guzman, and M.~Ghavamzadeh, ``A lyapunov-based
  approach to safe reinforcement learning,'' in \emph{Advances in neural
  information processing systems}, 2018, pp. 8092--8101.

\bibitem{chow2019lyapunov}
Y.~Chow, O.~Nachum, A.~Faust, E.~Duenez-Guzman, and M.~Ghavamzadeh,
  ``Lyapunov-based safe policy optimization for continuous control,'' in
  \emph{RL4RealLife Workshop in the 36th International Conference on Machine
  Learning}, 2019.

\bibitem{anderson2020neurosymbolic}
G.~Anderson, A.~Verma, I.~Dillig, and S.~Chaudhuri, ``Neurosymbolic
  reinforcement learning with formally verified exploration,'' in
  \emph{Advances in Neural Information Processing Systems}, 2020, pp.
  6172--6183.

\bibitem{verma2019imitation}
A.~Verma, H.~Le, Y.~Yue, and S.~Chaudhuri, ``Imitation-projected programmatic
  reinforcement learning,'' in \emph{Advances in Neural Information Processing
  Systems}, 2019, pp. 15\,752--15\,763.

\bibitem{bastani2018neurlps}
O.~Bastani, Y.~Pu, and A.~Solar-Lezama, ``Verifiable reinforcement learning via
  policy extraction,'' in \emph{Advances in Neural Information Processing
  Systems}, 2018, pp. 2499--2509.

\bibitem{weiss2018extracting}
G.~Weiss, Y.~Goldberg, and E.~Yahav, ``Extracting automata from recurrent
  neural networks using queries and counterexamples,'' in \emph{Proceedings of
  the 35th International Conference on Machine Learning}, 2018, pp. 5247--5256.

\bibitem{carr2020verifiable}
S.~Carr, N.~Jansen, and U.~Topcu, ``Verifiable rnn-based policies for pomdps
  under temporal logic constraints,'' in \emph{Proceedings of the 29th
  International Joint Conference on Artificial Intelligence}, 2020, p.
  4121–4127.

\bibitem{GP}
C.~E. Rasmussen and C.~Williams, ``Gaussian processes for machine learning,''
  \emph{the MIT Press}, 2006.

\bibitem{finn2016icml}
C.~Finn, S.~Levine, and P.~Abbeel, ``Guided cost learning: deep inverse optimal
  control via policy optimization,'' in \emph{Proceedings of the 33rd
  International Conference on Machine Learning}, 2016.

\bibitem{specGame}
F.~Rossi and N.~Mattei, ``Building ethically bounded \textsc{AI},'' in
  \emph{Proceedings of the 33rd AAAI Conference on Artificial Intelligence},
  2019, pp. 9785--9789.

\bibitem{montufar2014number}
G.~F. Montufar, R.~Pascanu, K.~Cho, and Y.~Bengio, ``On the number of linear
  regions of deep neural networks,'' in \emph{Advances in Neural Information
  Processing Systems}, 2014.

\bibitem{ferlez2021cdc}
J.~Ferlez and Y.~Shoukry, ``Bounding the complexity of formally verifying
  neural networks: A geometric approach,'' in \emph{60th IEEE Conference on
  Decision and Control}, 2021, pp. 5104--5109.

\bibitem{bak2020cav}
S.~Bak, H.-D. Tran, K.~Hobbs, and T.~T. Johnson, ``Improved geometric path
  enumeration for verifying relu neural networks,'' in \emph{Proceedings of the
  32nd International Conference on Computer Aided Verification}, 2020.

\bibitem{ppo}
J.~Schulman, F.~Wolski, P.~Dhariwal, A.~Radford, and O.~Klimov, ``Proximal
  policy optimization algorithms,'' \emph{arXiv:1707.06347}, 2017.

\bibitem{latvala2003spin}
T.~Latvala, ``Efficient model checking of safety properties,'' in \emph{Model
  Checking Software. 10th International SPIN Workshop}.\hskip 1em plus 0.5em
  minus 0.4em\relax Springer, 2003, pp. 74--88.

\bibitem{gerth1996}
R.~Gerth, D.~Peled, M.~Y. Vardi, and P.~Wolper, ``Simple on-the-fly automatic
  verification of linear temporal logic,'' in \emph{Proceedings of the 15th
  IFIP WG6.1 International Symposium on Protocol Specification, Testing and
  Verification XV}, 1996, pp. 3--18.

\bibitem{soudjani2013siam}
S.~{Esmaeil Zadeh Soudjani} and A.~Abate, ``Adaptive and sequential gridding
  procedures for the abstraction and verification of stochastic processes,''
  \emph{SIAM Journal on Applied Dynamical Systems}, vol.~12, no.~2, pp.
  921--956, 2013.

\bibitem{hsu2018hscc}
K.~Hsu, R.~Majumdar, K.~Mallik, and A.-K. Schmuck, ``Multi-layered
  abstraction-based controller synthesis for continuous-time systems,'' in
  \emph{Proceedings of the 21st International Conference on Hybrid Systems:
  Computation and Control}, 2018, p. 120–129.

\bibitem{chollet2015keras}
\BIBentryALTinterwordspacing
F.~Chollet \emph{et~al.} (2015) Keras. [Online]. Available:
  \url{https://github.com/fchollet/keras}
\BIBentrySTDinterwordspacing

\bibitem{rearwheel}
G.~Klančar, A.~Zdešar, S.~Blažič, and I.~Škrjanc, ``Wheeled mobile
  robotics,'' \emph{Elsevier}, 2017.

\bibitem{rakelly2019icml}
K.~Rakelly, A.~Zhou, D.~Quillen, C.~Finn, and S.~Levine, ``Efficient off-policy
  meta-reinforcement learning via probabilistic context variables,'' in
  \emph{Proceedings of the 36th International Conference on Machine Learning},
  2019.

\bibitem{kingma2014icml}
D.~P. Kingma and M.~Welling, ``Auto-encoding variational bayes,'' in
  \emph{Proceedings of the 31st International Conference on Machine Learning},
  2014.

\end{thebibliography}


 




\begin{IEEEbiography}[{\includegraphics[width=1in,height=1.25in,clip,keepaspectratio]{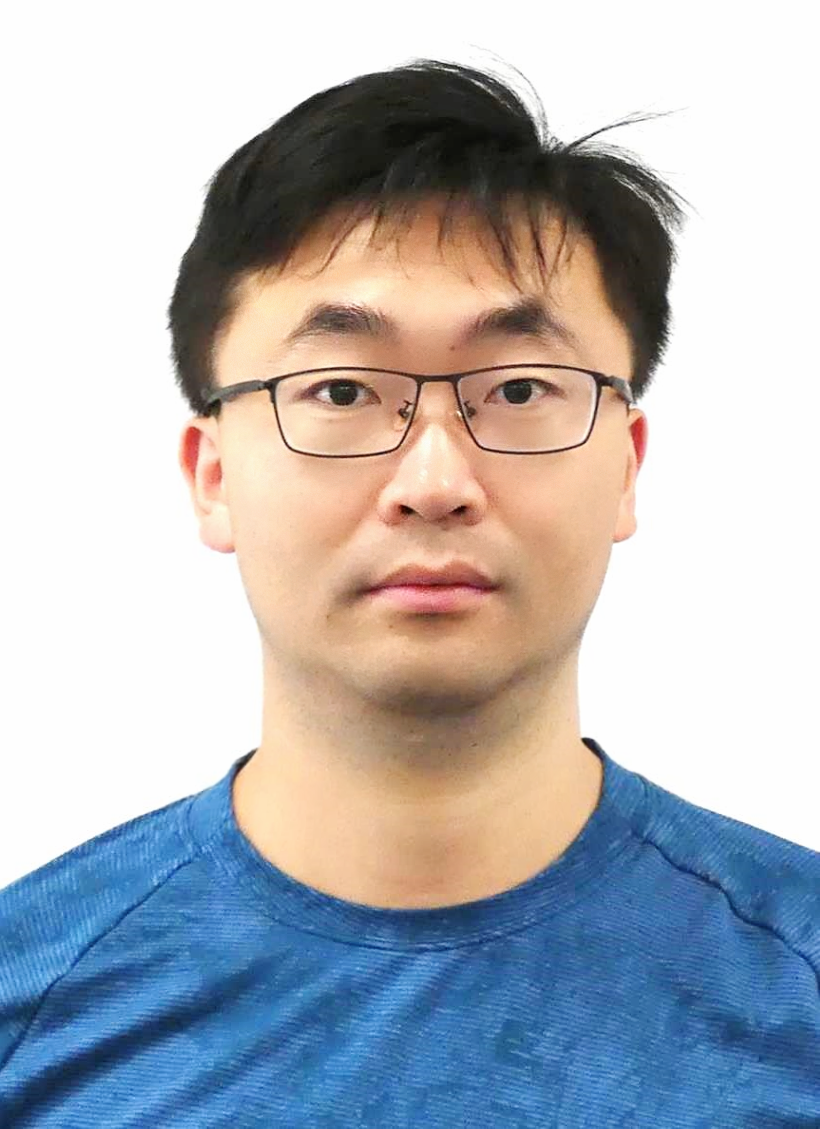}}]{Xiaowu Sun} is currently pursuing the Ph.D. degree with the Department of Electrical
Engineering and Computer Science at the University of California, Irvine. He received his M.Sc. degree in Electrical Engineering from the University of Maryland, College Park in 2018, and his B.Sc. degree in Physics from Nanjing University, China in 2013.  

His research interests include formal methods for control, neural networks, reinforcement learning and robotics. Xiaowu was the finalist in the ACM SIGBED SRC Student Competition at the Cyber-Physical Systems (CPS-IoT) Week 2021. His research on using formal verification to analyze nerual network controlled systems was nominated for consideration in the Communications of the ACM (CACM) Research Highlights.
\end{IEEEbiography}

\begin{IEEEbiography}[{\includegraphics[width=1in,height=1.25in,clip,keepaspectratio]{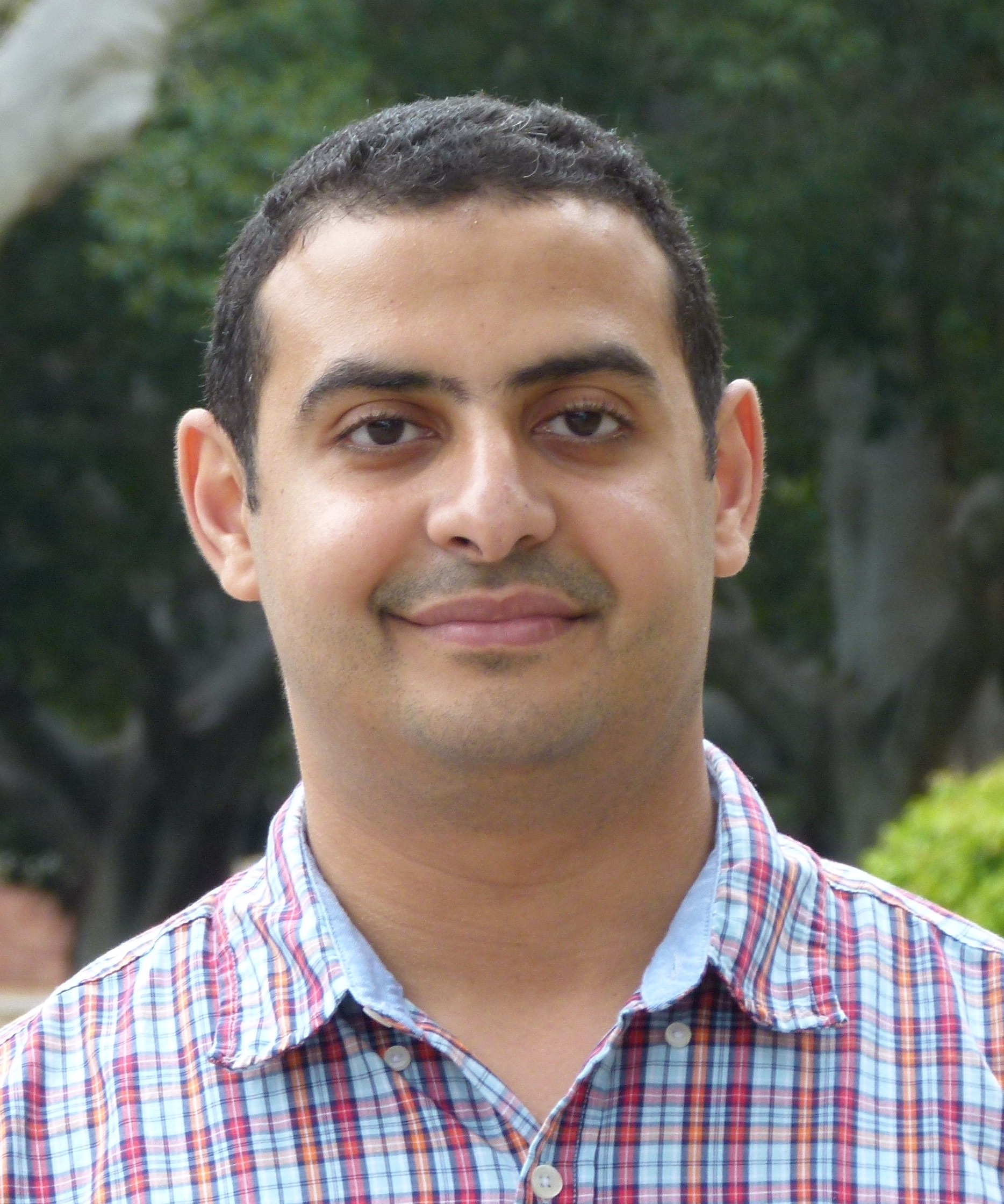}}]{Yasser Shoukry}
 is an Assistant Professor in the Department of Electrical Engineering and Computer Science at the University of California, Irvine where he leads the Resilient Cyber-Physical Systems Lab. Before joining UCI, he spent two years as an assistant professor at the University of Maryland, College Park. He received his Ph.D. in Electrical Engineering from the University of California, Los Angeles in 2015. Between September 2015 and July 2017, Yasser was a joint postdoctoral researcher at UC Berkeley, UCLA, and UPenn. His current research focuses on the design and implementation of resilient, AI-enabled, cyber-physical systems and IoT. His work in this domain was recognized by Early Career Award from the IEEE Technical Committee on Cyber-Physical Systems (TC-CPS) in 2021, the NSF CAREER Award in 2019, the Best Demo Award from the International Conference on Information Processing in Sensor Networks (IPSN) in 2017, the Best Paper Award from the International Conference on Cyber-Physical Systems (ICCPS) in 2016, and the Distinguished Dissertation Award from UCLA EE department in 2016. In 2015, he led the UCLA/Caltech/CMU team to win the NSF Early Career Investigators (NSF-ECI) research challenge. His team represented the NSF- ECI in the NIST Global Cities Technology Challenge, an initiative designed to advance the deployment of Internet of Things (IoT) technologies within a smart city. He is also the recipient of the 2019 George Corcoran Memorial Award from the University of Maryland for his contributions to teaching and educational leadership in the field of CPS and IoT.
\end{IEEEbiography}

\vfill

\end{document}